\documentclass[10pt,twocolumn,letterpaper]{article}

\usepackage[pagenumbers]{wacv} 

\usepackage{graphicx}
\usepackage{booktabs}
\usepackage{url}
\usepackage{amsfonts}       
\usepackage{nicefrac}       
\usepackage{microtype}      
\usepackage{xcolor}         
\usepackage{amsmath}
\usepackage{multirow,enumitem}
\usepackage{multicol}
\usepackage{subcaption}
\usepackage{colortbl} 
\usepackage{makecell}
\usepackage{algorithm}
\usepackage{algorithmic}
\usepackage{amssymb}
\usepackage{diagbox}

\definecolor{wacvblue}{rgb}{0.21,0.49,0.74}
\usepackage[pagebackref,breaklinks,colorlinks,allcolors=wacvblue]{hyperref}
\usepackage[capitalize]{cleveref} 

\def\wacvPaperID{1662} 
\def\confName{WACV}
\def\confYear{2027}

\title{Tracking the Discriminative Axis: \\Dual Prototypes for Test-Time OOD Detection Under Covariate Shift}

\author{
Wooseok Lee$^{*}$ \quad
Jin Mo Yang$^{*}$ \quad
Saewoong Bahk \quad
Hyung-Sin Kim%
\setcounter{footnote}{1}%
\thanks{Corresponding author}\\
Seoul National University\\
{\tt\small andylws@snu.ac.kr,
jmyang@netlab.snu.ac.kr,}\\
{\tt\small sbahk@snu.ac.kr,
hyungkim@snu.ac.kr}
}

\newcommand{\ours}{\textit{DART}}

\definecolor{mintblue}{RGB}{210,240,235}
\definecolor{skyblue}{RGB}{210,235,250}

\begin{document}
\maketitle

\begingroup
\makeatletter
\renewcommand{\thefootnote}{*}
\renewcommand{\@makefnmark}{\hbox{$^{*}$}}
\footnotetext[1]{Equal contribution.}
\makeatother
\endgroup
\setcounter{footnote}{0}

\begin{abstract}

For reliable deployment of deep-learning systems, out-of-distribution (OOD) detection is indispensable. 
In the real world, where test-time inputs often arrive as streaming mixtures of in-distribution (ID) and OOD samples under evolving covariate shifts, OOD samples are domain-constrained and bounded by the environment, and both ID and OOD are jointly affected by the same covariate factors. 
Existing methods typically assume a stationary ID distribution, but this assumption breaks down in such settings, leading to performance degradation. 
We empirically discover that, even under covariate shift, ID and OOD samples remain separable along a discriminative axis in feature space. 
Building on this observation, we propose \ours, a test-time, online OOD detection method that dynamically tracks dual prototypes---one for ID and the other for OOD---to recover the drifting discriminative axis, augmented with multi-layer fusion and flip correction for robustness. 
Extensive experiments on a wide range of challenging benchmarks, where all datasets are subjected to 15 corruption types, demonstrate that our method significantly improves performance, yielding 15.32 percentage points (pp) AUROC gain and 49.15 pp FPR95 reduction on ImageNet-C vs. Textures-C compared to established baselines. 
These results highlight the potential of the test-time discriminative axis tracking for dependable OOD detection in dynamically changing environments.
Code: \href{https://github.com/andylws/DART}{https://github.com/andylws/DART}
\end{abstract}
\section{Introduction}
\label{sec:introduction}

Deep neural networks (DNNs) achieve remarkable performance across applications such as image classification, object detection, medical imaging, and autonomous driving~\cite{alam2020survey}. 
These successes stem from large-scale datasets, high-performance hardware, and novel model architectures~\cite{deng2009imagenet, krizhevsky2012imagenet, he2016deep, vaswani2017attention}, motivating real-world deployment.

\begin{figure}[t]
    \centering
    \includegraphics[width=0.6\linewidth]{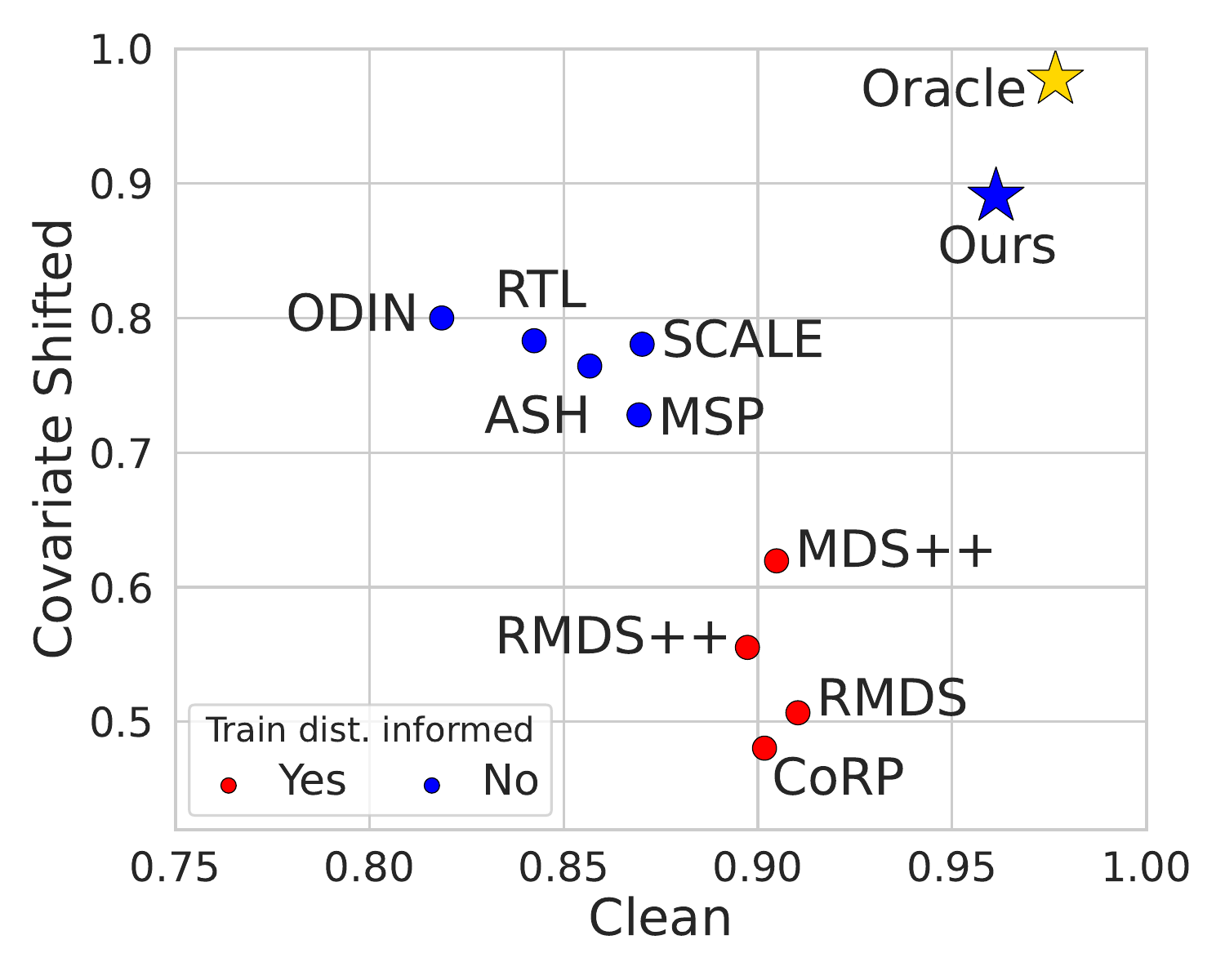}
    \vspace{-1em}
    \caption{AUROC comparison on both covariate shifted and clean ImageNet-based benchmark. Existing methods suffer under covariate shift, with train distribution–informed approaches dropping to around 0.5. In contrast, the oracle axis achieves consistently high performance regardless of shift, and our method effectively discovers this axis, attaining near-oracle results.}
    \label{fig:auroc_plot}
    \vspace{-1.2em}
\end{figure}

In practice, however, deployed models inevitably encounter test inputs that deviate from their training distributions. 
One form is \textbf{semantic shift}, where models face unknown semantics---commonly termed out-of-distribution (OOD) samples. 
Substantial progress has been made on OOD detection: existing methods typically 
assume either abstract characteristics~\cite{hendrycks2016baseline,liu2020energy,xu2023scaling} or data-specific characteristics~\cite{lee2018simple,sun2022out} of in-distribution (ID) data to distinguish ID from OOD. 
A second form is \textbf{covariate shift}, where data appears under new conditions such as changes in weather, illumination, or sensor noise ~\cite{moreno2012unifying, dockes2021preventing}. 
Most OOD methods implicitly assume stationary ID distributions as reference to separate ID and OOD; however, as shown in \cref{fig:auroc_plot}, in practice they struggle under covariate shifts~\cite{yang2024generalized, yang2021semantically, yang2023full} because shifting covariates alter the space geometry that their decision rules rely on.

\begin{figure}[t]
    \centering
    \includegraphics[width=0.95\linewidth]{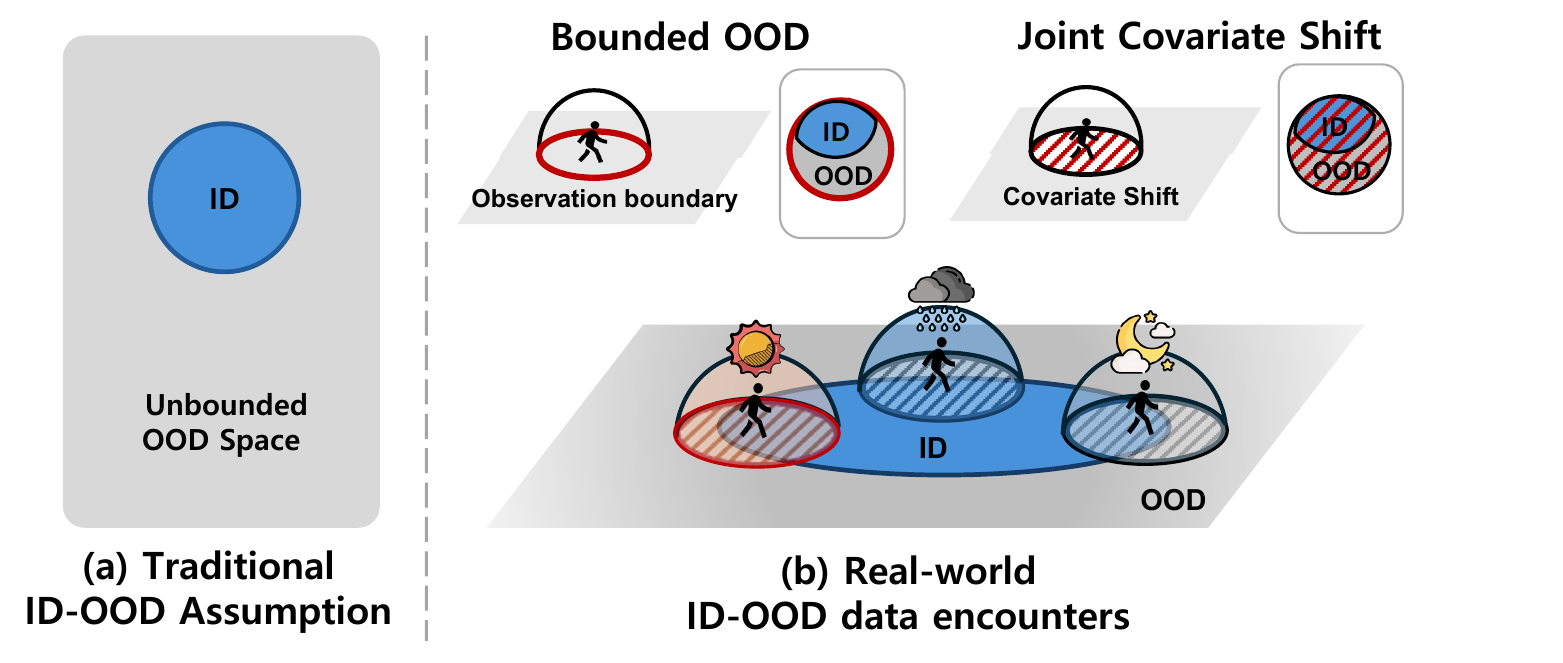}
    \vspace{-1.2em}
    \caption{Comparison of traditional and real-world ID-OOD assumptions. (a) Traditional OOD detection assumes ID data (blue circle) exists within an unbounded OOD space (gray background). (b) In real-world scenarios, OOD data is bounded by physical and environmental constraints (observation boundary, top-left inset), limiting the space where OOD samples can occur. Furthermore, covariate shifts such as weather conditions can simultaneously affect both ID and OOD distributions (dashed regions).
    }
    \label{fig:scenario}
    \vspace{-1em}
\end{figure}

We study \textbf{test-time OOD detection under covariate shift} in a realistic \emph{streaming mixture} setting: 
test-time inputs arrive in mini-batches as mixtures of ID and OOD samples, and both are simultaneously exposed to the \emph{same evolving covariate shifts} (\eg, a change in weather). We denote these as \textbf{covariate-shifted ID (csID)} and \textbf{covariate-shifted OOD (csOOD)}. 
Within each mini-batch---as illustrated in \cref{fig:scenario}---spatial and temporal coherence arises from the task environment, so OOD samples are \emph{domain-constrained} rather than arbitrary. For instance, in autonomous driving, encountering an unseen vehicle type is plausible OOD, whereas suddenly observing medical or satellite images is essentially impossible. Moreover, spatio-temporally correlated csID and csOOD undergo the \emph{same} covariate shift, so their distributions co-evolve during deployment.
In our setting, the test stream is unlabeled, the backbone is frozen, no training data are accessed at test time, and the algorithm maintains a small, bounded state.

In this practical scenario, we empirically observe a key insight that enables our approach. Across diverse datasets and shifts, we consistently observe: 
(i) \textit{local coherence}---within short windows, csOOD samples organize into coherent groupings in feature space; and 
(ii) a \textit{recoverable linear axis}---csID and csOOD remain approximately linearly separable along a dominant discriminative direction that \emph{drifts} as covariates evolve. 
As shown in \cref{fig:auroc_plot} with annotation ``Oracle'', computing our method’s OOD score with the optimal discriminative axis yields very high AUROC, demonstrating that separability in this direction can lead to strong detection performance.
These observations suggest focusing on \emph{tracking} the separation direction online rather than relying on a fixed, training-time score.

We propose \textbf{D}iscriminative \textbf{A}xis \textbf{R}eal-time \textbf{T}racker (\textbf{\ours}), a \textbf{test-time}, \textbf{online OOD detection} method that continuously tracks a \textit{discriminative axis} using a class-agnostic ID prototype and an OOD prototype \emph{per feature layer}. 
At each step, incoming test samples update the prototypes via lightweight, stable rules, yielding the vector connecting them as the discriminative axis. 
Each sample is then scored by its relative position to this axis, yielding a detector aligned with the evolving feature space. 
To address the fact that covariate shifts affect DNN layers differently~\cite{hendrycks2019benchmarking, yin2019fourier}, \ours~employs \textbf{multi-layer score fusion} to stabilize detection across heterogeneous, unpredictable shifts. 
The method requires only forward passes, no model weight update, no label, and no access to training data at test time, which suits privacy-sensitive or on-device deployments where retraining is infeasible. 
For robustness, prototypes are initialized conservatively and updated with safeguards to prevent collapse.

Across challenging benchmarks,
\ours~consistently delivers substantial gains over prior approaches. 
For example, on the ImageNet benchmark, \ours~achieves at least \textbf{9.04 percentage points (pp)} higher AUROC under covariate shift and \textbf{5.10 pp} AUROC improvement on the clean setting, as shown in \cref{fig:auroc_plot}, while attaining \textbf{40.15 pp} and \textbf{19.06 pp} FPR@95TPR reduction on the covariate shifted and clean datasets, respectively.
Notably, the performance of \ours~comes close to that of the Oracle, highlighting the effectiveness of our approach.

Our contributions can be summarized as below:
\begin{itemize}[leftmargin=*]
    \item We formalize \emph{test-time OOD detection under covariate shift} in a streaming mixture setting, distinguishing csID from csOOD and articulating realistic constraints (data stream, frozen backbone, small memory).
    \item We introduce \textbf{\ours}, which \emph{tracks dual-prototypes online} to recover the drifting discriminative axis and fuses \emph{multi-layer} scores for robustness to layer-wise covariate effects.
    \item We provide measurements and visualizations showing coherent csOOD groupings and approximately linear csID--csOOD separation within short  windows.
    \item We demonstrate consistent gains over strong post-hoc baselines on joint-shift suites, with large improvements in AUROC and FPR@95.
\end{itemize}

\section{Related Work}
\label{sec:related_work}

\subsection{OOD Detection: Training-Driven vs. Post-hoc}

Research on OOD detection can be broadly categorized into training-driven and post-hoc approaches.

\textbf{Training-driven approaches}
modify training to enhance OOD separability, \eg, Outlier Exposure (OE) with auxiliary outliers~\cite{hendrycks2018deep, zhang2023mixture, zhu2023diversified} and $N{+}1$ classifiers that add an ``unknown'' class~\cite{bendale2016towards, shu2017doc, chen2021adversarial}. While effective, they require additional data or altered objectives and may misalign with deployment OODs, with potential side effects on ID accuracy.
Some methods~\cite{katz2022training, yang2023auto} update model parameters at test time via backpropagation, which introduces latency and can compromise ID accuracy under non-stationary streams.

\begin{figure*}[t]
    \centering
    \includegraphics[width=0.7\textwidth]{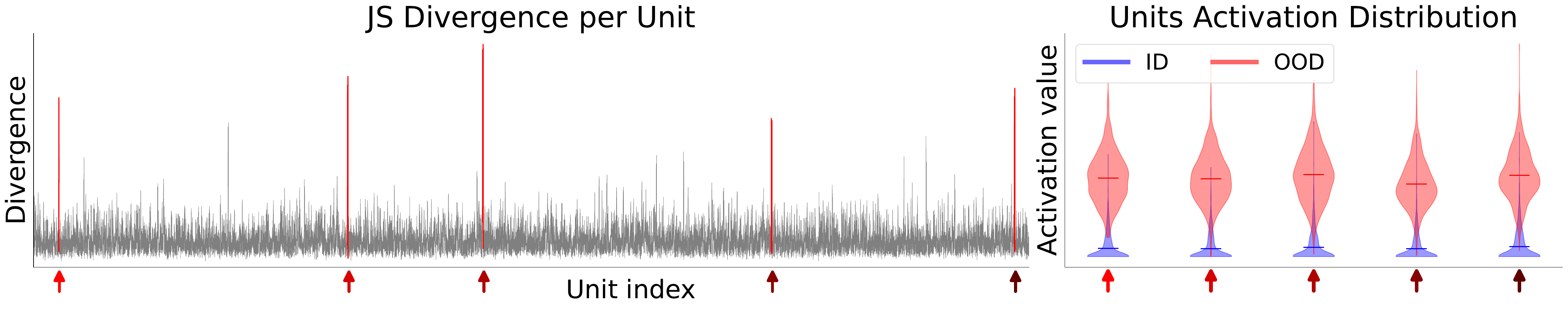}
    \vspace{-0.3em}
    \caption{Unit-wise activation analysis. The left panel shows the JSD between ID and OOD activations, with arrows marking units of large divergence. The right panel visualizes the activation distributions of these units, where ID (\textcolor{blue}{blue}) and OOD (\textcolor{red}{red}) are clearly separable.}
    \label{fig:unit_viz}
    \vspace{-0.7em}
\end{figure*}

\textbf{Post-hoc approaches}
operate on a frozen classifier without retraining and have gained widespread adoption due to their ease of use and compatibility with pretrained models.
%
Categories include: output-based~\cite{hendrycks2016baseline, hendrycks2019scaling, liu2020energy}, distance-based~\cite{lee2018simple, ren2021simple, mueller2025mahalanobis++, sun2022out, park2023nearest}, feature-based~\cite{liang2018enhancing, wang2022vim, sun2021react, djurisic2022extremely, xu2023scaling, zhang2022out,fang2024kernel}, and gradient-based~\cite{huang2021importance, behpour2023gradorth} methods. 
Furthermore, \emph{training distribution-informed} methods (\eg, Mahalanobis, ViM and KNN) \emph{assume access to training statistics} (feature means, covariances, principal subspaces)---assumptions that can become invalid under test-time covariate drift and are often infeasible when training data are unavailable. 
In contrast, our method is post-hoc and relies solely on the unlabeled test stream, without training statistics.

\textbf{Post-hoc, test-time adaptive approaches} are an emerging line of work which adjust OOD detection rule using test-time batches/streams \emph{without} weight update. 
RTL~\cite{fan2024test} uncovers a linear trend between OOD scores and features and fits a batch-level discriminator; OODD~\cite{yang2025oodd} maintains an online dictionary that updates only OOD features, keeping the ID side fixed from the training set.
Although these approaches are closer in spirit to our online setting, they do not explicitly address time-varying covariate shift and incur substantial test-time computation and memory overhead, potentially limiting practical deployment.
%
\ours~ maintains a persistent, memory-light state that adapts smoothly to test-time drift. 
Our goal is OOD detection under covariate drift in streaming mixtures; conventional test-time adaptation (TTA) methods that target closed-set robustness and/or adapt model weights~\cite{wang2020tent,niu2023towards} are orthogonal to this objective and not our focus.

\subsection{Covariate Shift and Joint-Shift Evaluation}

\textbf{Covariate corruptions.} 
Covariate shift---changes in input distributions with fixed labels---is commonly studied with corruption suites such as ImageNet-C~\cite{hendrycks2019benchmarking}, which introduce noise, blur, weather, and other factors. 
Recent work emphasizes natural shift from environment and sensor~\cite{baek2024unexplored,baekadaptive,kim2026imagenet}.

\noindent\textbf{Joint semantic and covariate shift.} 
Full-spectrum OOD~\cite{yang2023full} evaluates semantic OOD while allowing covariate variation; OpenOOD unifies large-scale OOD evaluation and includes \emph{joint-shift} settings~\cite{yang2022openood,zhang2023openood}. Dataset design such as NINCO~\cite{bitterwolf2023or} reduces ID contamination for clearer semantic separation. 
Our setting follows this trajectory but explicitly considers \emph{streaming mixtures} where spatio-temporally correlated csID and csOOD experience the \emph{same evolving covariates}. \ours~is designed to adapt online in such scenarios via dual prototypes and multi-layer score fusion, without requiring training data prior or weight updates.

\section{Method}
\label{sec:method}

In this section, we introduce our method \ours~for online test-time OOD detection under covariate shift.
We begin by revisiting a key motivation behind our approach: the empirical emergence of the discriminative axis in pre-trained feature space.
We then describe how the prototypes that define this axis are iteratively refined with incoming test batches.
Finally, we explain why multi-layer fusion is essential to maintain robustness to unpredictable covariate shifts.

\subsection{Formulation Setup}
We define \(\mathcal{D}_\mathrm{I}\) and \(\mathcal{D}_\mathrm{O}\) as the datasets for ID and OOD, observed by the OOD detection system. 
Then, let \(\mathcal{B}_t = \{{\mathbf{x}_{t,1}, \mathbf{x}_{t,2}, \dots, \mathbf{x}_{t,N}}\}\) denote an input batch received at test time, where each sample \(\mathbf{x}_{t,i}\) may belong to one of two categories under covariate shift: covariate-shifted in-distribution (csID) or covariate-shifted out-of-distribution (csOOD). 
We denote the subset of csID samples as $\mathcal{B}_t^{\mathrm{I}}$ and the csOOD samples as $\mathcal{B}_t^{\mathrm{O}}$, such that $\mathcal{B}_t = \mathcal{B}_t^{\mathrm{I}} \cup \mathcal{B}_t^{\mathrm{O}}$.

We extract intermediate features from several layers of the model. 
Let $f_{l}(\cdot)$ denote the feature mapping at layer $l \in \mathcal{L} = \{1, 2, \dots, L\}$. For a given input $\mathbf{x}$, we obtain multi-layer features $\{\mathbf{z}^{(l)}\}_{l \in \mathcal{L}}$, where $\mathbf{z}^{(l)} = f_{l}(\mathbf{x})$.
Thus, a csID sample $\mathbf{x}_t^{\mathrm{I}} \in \mathcal{B}_t^{\mathrm{I}}$ and a csOOD sample $\mathbf{x}_t^{\mathrm{O}} \in \mathcal{B}_t^{\mathrm{O}}$ yield feature sets $\mathbf{Z}_t^{\mathrm{I}} = \{\mathbf{z}_{t}^{(l),\mathrm{I}}\}_{l \in \mathcal{L}}$ and $\mathbf{Z}_t^{\mathrm{O}} = \{\mathbf{z}_{t}^{(l),\mathrm{O}}\}_{l \in \mathcal{L}}$.

\subsection{Discriminative Axis in Feature Space}
Prior works~\cite{sun2021react, xu2023scaling} have reported that ID and OOD samples exhibit distinct activation patterns in the feature space. 
In a similar spirit, we systematically examine unit-level activations from a distributional perspective. 
Specifically, we collect unit-wise activation distributions across ID and OOD samples and compare them by the Jensen--Shannon divergence (JSD) \cite{61115}~between the ID and OOD activation distributions $P_u$ and $Q_u$ of unit $u$:
\begin{equation}
\mathrm{JSD}(P_u \,\|\, Q_u) = \tfrac{1}{2} D_{\mathrm{KL}}(P_u \,\|\, M_u) + \tfrac{1}{2} D_{\mathrm{KL}}(Q_u \,\|\, M_u),
\label{eq:jsd}
\end{equation}
where $M_u = \tfrac{1}{2}(P_u + Q_u)$ and $D_{\mathrm{KL}}$ is the Kullback--Leibler divergence.
Our analysis reveals that there exist certain units where ID and OOD activations diverge substantially, as evidenced by a large JSD in \cref{fig:unit_viz}. 
Visualization via violin plots further demonstrates that ID and OOD activations can be sharply distinguished within those units. 

\begin{figure}[t]
    \centering
    \begin{subfigure}[b]{0.85\linewidth}
        \centering
        \includegraphics[width=\linewidth]{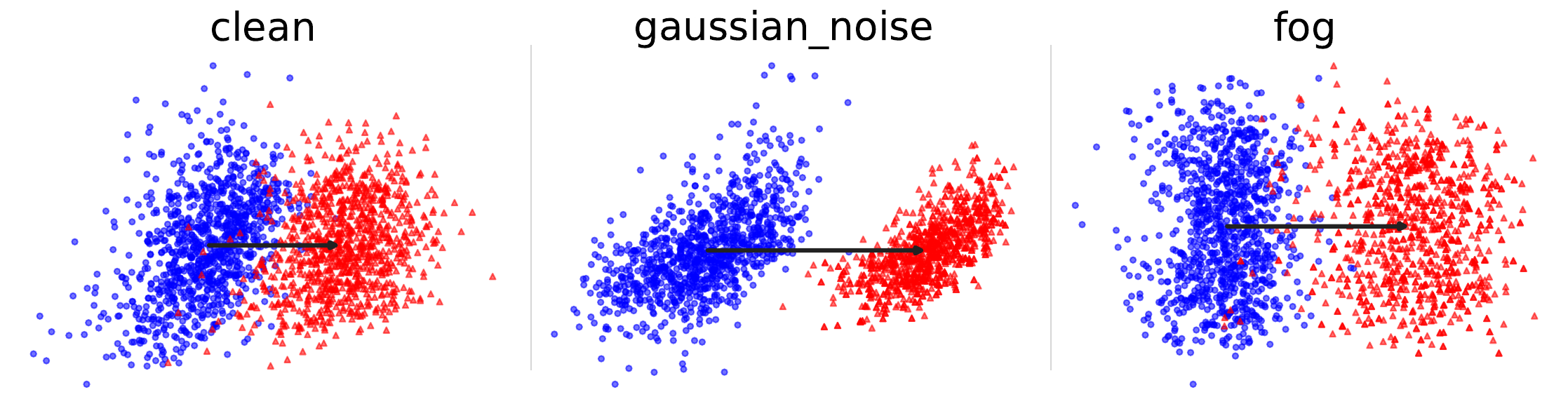}
        \caption{CIFAR-100 vs. LSUN}
        \label{fig:axis1}
    \end{subfigure}
    \hfill
    \begin{subfigure}[b]{0.85\linewidth}
        \centering
        \includegraphics[width=\linewidth]{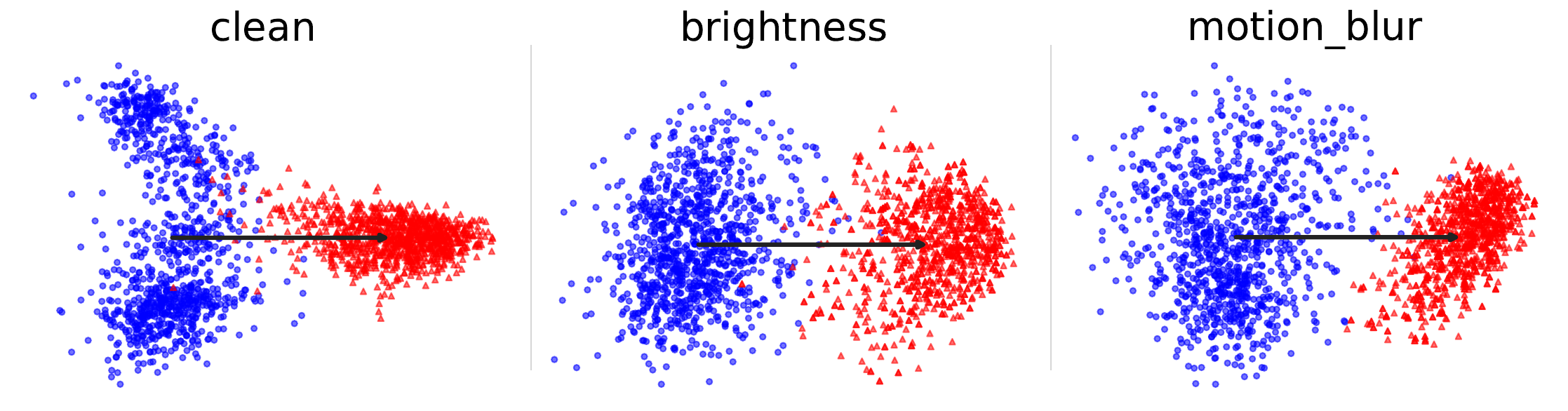}
        \caption{ImageNet vs. iNaturalist}
        \label{fig:axis2}
    \end{subfigure}
    \vspace{-0.5em}
    \caption{Distribution of ID (\textcolor{blue}{blue dots}) and OOD (\textcolor{red}{red dots}) samples in features space projected with the oracle discriminative axis as the horizontal axis.}
    \label{fig:discriminative_axis}
    \vspace{-1em}
\end{figure}

Building upon this insight, we leverage these distributional differences to construct a discriminative direction. 
We compute prototype representations by averaging features across all ID samples and all OOD samples respectively, yielding two representative points in the feature space\footnote{While separability analysis may be conducted at each feature layers, we omit the layer index from the notation in this section for clarity.}: 
\begin{equation}
\mathbf{p}_{\text{I}} = \frac{1}{|\mathcal{D}_{\text{I}}|} \sum_{\mathbf{x} \in \mathcal{D}_{\text{I}}} f(\mathbf{x}), \quad
\mathbf{p}_{\text{O}} = \frac{1}{|\mathcal{D}_{\text{O}}|} \sum_{\mathbf{x} \in \mathcal{D}_{\text{O}}} f(\mathbf{x}).
\end{equation}
Then, we define the vector connecting prototypes as the \textbf{discriminative axis}:
$
\mathbf{axis}^{oracle}_{\text{disc}} = \mathbf{p}_{\text{I}} - \mathbf{p}_{\text{O}}.
$
This averaging implicitly weights units by their ID–OOD mean gap, emphasizing discriminative units with large mean difference and divergence while suppressing non-discriminative units.
%
\Cref{fig:discriminative_axis} shows distinct ID/OOD clusters under projection onto this axis, regardless of the presence or type of covariate shift, supporting the existence of a discriminative axis.

Despite its existence, finding the discriminative axis at test time is another challenge. 
Since the nature of OOD is inherently unknown before test-time, it is not feasible to predefine such a discriminative axis in advance. 
This motivates the need to adaptively identify the optimal axis during test-time. 
To this end, we propose a method that \textit{progressively tracks two prototypes}---one for ID and the other for OOD---where the vector connecting them defines a discriminative axis that adapts to the evolving data stream.

\subsection{Batch-wise Prototype Refinement: Tracking the Discriminative Axis}

To identify the discriminative axis in an online manner, we refine prototypes through iterative pseudo-labeling and prototype updating. Refer to Appendix for the detailed algorithm.

Our method initializes and dynamically updates layer-specific prototypes for ID and OOD based on current test batch features. Since true ID/OOD labels are unavailable at test-time, we leverage pseudo-labeling to distinguish between ID and OOD. We employ the Otsu algorithm~\cite{otsu1975threshold} to automatically determine optimal thresholds by maximizing the between-class variance, providing a principled way to separate samples based on their score distributions.

\textbf{Dual-Prototype Initialization.} 
For the initial batch, we use a naive baseline score, Maximum Softmax Probability (MSP), as our reference score. We assign pseudo-labels using MSP with the Otsu-determined threshold, then compute initial prototypes as the mean feature vectors of their respective pseudo-labeled groups, \ie $\bar{\mathbf{p}}_{1}^{\mathrm{I}} = \frac{1}{|\mathcal{S}_{1}^{\mathrm{I}}|} \sum_{i \in \mathcal{S}_{1}^{\mathrm{I}}} f(\mathbf{x}_{1,i}), \bar{\mathbf{p}}_{1}^{\mathrm{O}} = \frac{1}{|\mathcal{S}_{1}^{\mathrm{O}}|} \sum_{i \in \mathcal{S}_{1}^{\mathrm{O}}} f(\mathbf{x}_{1,i})$.

\textbf{Dual-Prototype Tracking.} 
%
For subsequent batches, we carry the prototypes across the stream and repeat the following steps to refine their estimates. 
This state is twofold: the current-batch estimate $\hat{\mathbf{p}}_t$ reflects the present covariate shift but is subject to pseudo-label noise, whereas the maintained prototype $\bar{\mathbf{p}}_t$ averages this noise over batches.
First, we \textit{(i)} compute Euclidean distances between each current-batch sample and the dual prototypes from the previous timestep, then \textit{(ii)} calculate a Relative Distance Score (RDS) that reflects each sample's position relative to both prototypes:
\begin{equation}
\text{RDS}(\mathbf{x}_{t,i}) = 1 - \frac{\|\mathbf{z}_{t,i} - \bar{\mathbf{p}}_{t-1}^{\mathrm{I}}\|_2}{\|\mathbf{z}_{t,i} - \bar{\mathbf{p}}_{t-1}^{\mathrm{I}}\|_2 + \|\mathbf{z}_{t,i} - \bar{\mathbf{p}}_{t-1}^{\mathrm{O}}\|_2},
\end{equation}
where \(\mathbf{z}_{t,i}\) denotes the feature of \(i\)-th sample in the batch \(t\). 
Next, using the Otsu algorithm, we \textit{(iii)} determine a threshold to assign pseudo-labels based on this RDS score. 
We \textit{(iv)} apply Tukey's \textit{outlier filtering}~\cite{tukey1977exploratory} to exclude samples that lie too far from their assigned prototypes, keeping unreliable ones out of the mean, and then \textit{(v)} compute new $\hat{\mathbf{p}}_{t}^{\text{I}}$ and $\hat{\mathbf{p}}_{t}^{\text{O}}$ as the mean feature of the remaining pseudo-labeled groups.
Finally, we \textit{(vi)} refine prototypes using an \textit{exponential moving average} to maintain stability:
\begin{equation}
    \bar{\mathbf{p}}_t^{\text{I}} = \alpha\, \mathbf{\bar{p}}_{t-1}^{\text{I}} + (1 - \alpha)\, \hat{\mathbf{p}}_{t}^{\text{I}}, \quad \bar{\mathbf{p}}_t^{\text{O}} = \alpha\, \mathbf{\bar{p}}_{t-1}^{\text{O}} + (1 - \alpha)\, \hat{\mathbf{p}}_{t}^{\text{O}}.
\end{equation}
This iterative process enables prototypes to progressively converge toward the true underlying distributions and identify a near-optimal discriminative axis.

\textbf{\textbf{Flip Correction.}} \label{para:flip_correction}
Since DART relies on pseudo-labeling, unreliable pseudo-labels can drive the prototypes toward opposite sides of their desirable locations. 
Once displaced this way, the updates continue from where the prototypes now stand and sustain the reversed configuration. 
Flip correction supplies the missing degree of freedom: every $k$-th batch, we test whether the prototypes are \emph{``flipped''} and swap them if so.
The \emph{flip} test compares the maintained prototypes with an auxiliary MSP-based ID prototype, \ie $\mathbf{\hat{p}}_{t,\mathrm{M}}^{\mathrm{I}}$. A \emph{flip} occurs when the ID prototype is significantly farther from the MSP reference than the OOD prototype, while showing lower cosine similarity. Formally, we swap prototypes if
\begin{equation}
\begin{aligned}
\| \mathbf{\bar{p}}_t^{\mathrm{I}} - \mathbf{\hat{p}}_{t,\mathrm{M}}^{\mathrm{I}} \|_2 &> 2 \| \mathbf{\bar{p}}_t^{\mathrm{O}} - \mathbf{\hat{p}}_{t,\mathrm{M}}^{\mathrm{I}} \|_2
\\[3pt] \text{and} \quad
\cos\left(\mathbf{\bar{p}}_t^{\mathrm{I}}, \mathbf{\hat{p}}_{t,\mathrm{M}}^{\mathrm{I}}\right) &< \cos\left(\mathbf{\bar{p}}_t^{\mathrm{O}}, \mathbf{\hat{p}}_{t,\mathrm{M}}^{\mathrm{I}}\right).
\end{aligned}
\end{equation}
We use a weighted comparison (factor of 2) to impose a strict condition preventing unintended flip detections.

\begin{figure}[t]
    \centering
    
    \begin{subfigure}[b]{0.45\linewidth}
        \centering
        \includegraphics[width=\linewidth]{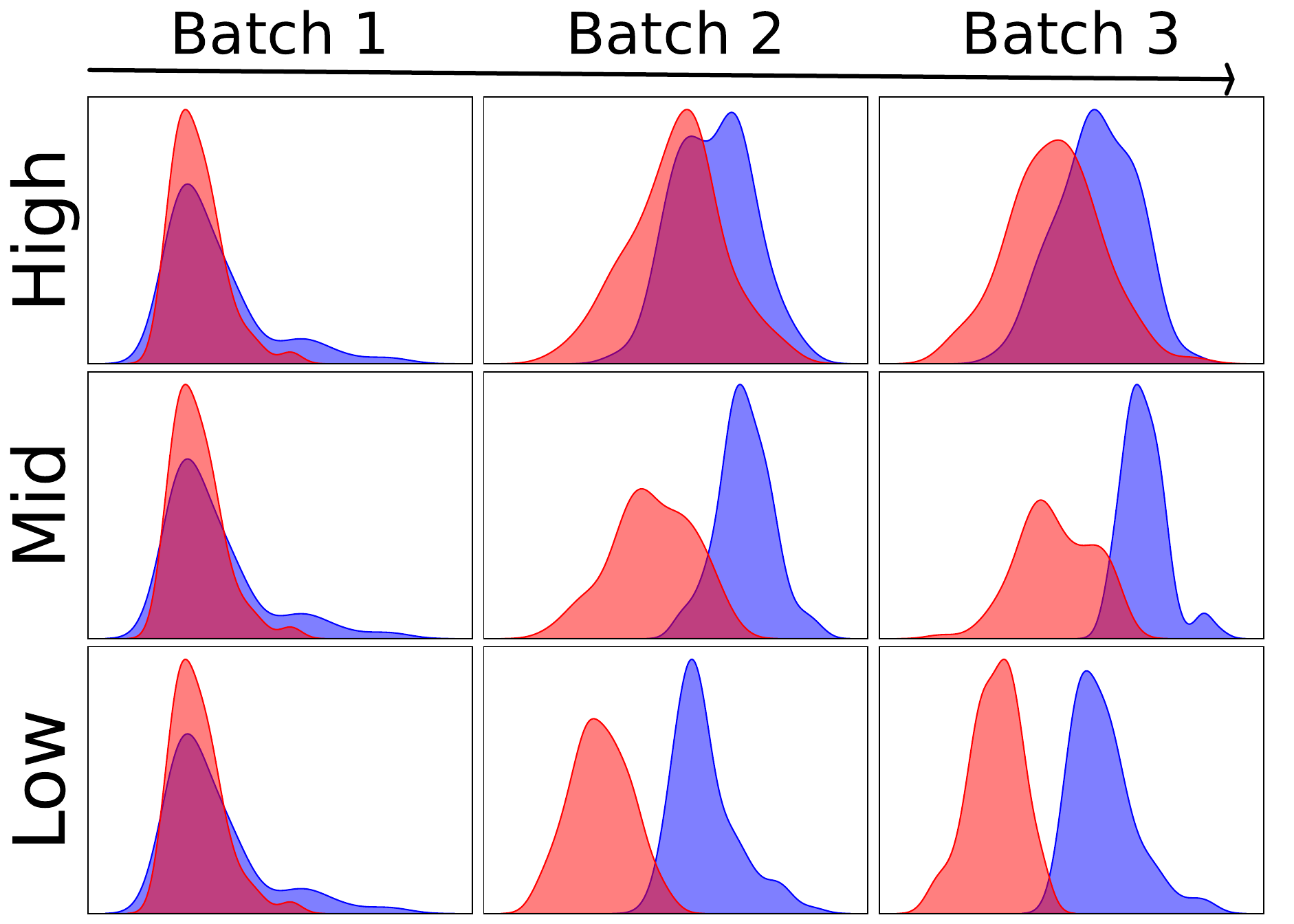}
        \caption{RDS under Gaussian noise}
        \label{fig:rds_distribution_gaus}
    \end{subfigure}
    \hfill
    \begin{subfigure}[b]{0.45\linewidth}
        \centering
        \includegraphics[width=\linewidth]{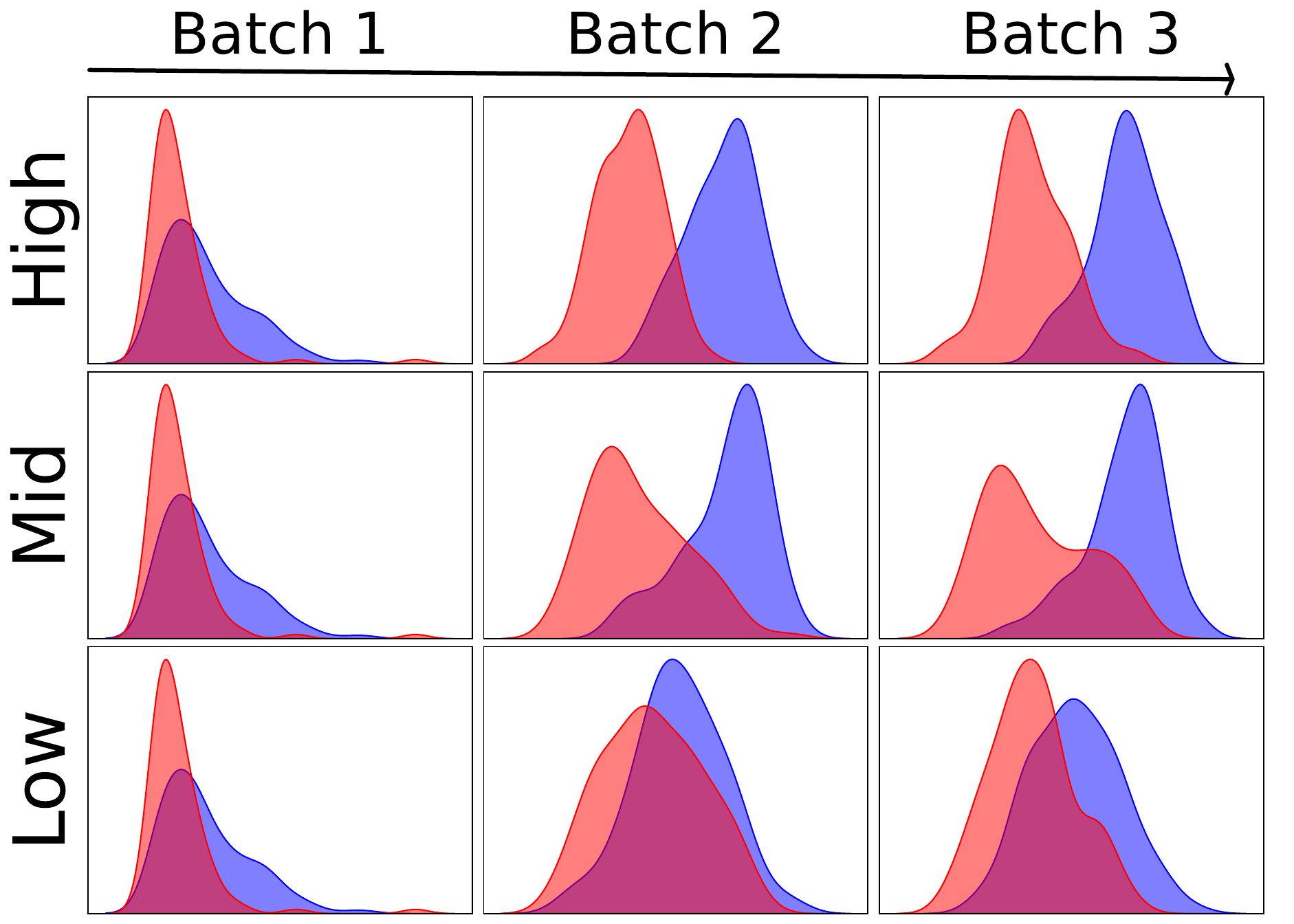}
        \caption{RDS under defocus blur}
        \label{fig:rds_distribution_defocus}
    \end{subfigure}
    \vspace{-0.5em}
    \caption{
    Layer-wise RDS distributions under two different covariate shift types. 
    Each plot shows the RDS distribution of \textbf{ID} (\textbf{\textcolor{blue}{blue curve}}, CIFAR-100) and \textbf{OOD} (\textbf{\textcolor{red}{red curve}}, LSUN) samples at different network depths (low, mid, high-level) through three sequential batches.
    The visualizations reveal how different corruption types affect feature separability at specific network layers; under Gaussian noise, separability degrades in high-level layers, whereas under defocus blur, it degrades in low-level layers.
    }
    \label{fig:multi_layer}
    \vspace{-1em}
\end{figure}

\subsection{Multi-Layer Score Fusion}
To enhance discriminative axis identification, we extend our approach to multi-layer features. Low-level features capture local patterns like textures and edges, while high-level features encode semantic concepts~\cite{guo2016deep}. However, covariate shifts can selectively disrupt different levels of visual information~\cite{hendrycks2019benchmarking,yin2019fourier}---blur corruptions primarily affect low-level features, while elastic transformations impair higher-level representations. As a result, different layers exhibit varying degrees of ID/OOD discriminability depending on the shift type, as illustrated in \cref{fig:multi_layer}. Since the nature of covariate shift is typically \textit{unknown beforehand}, leveraging information from all feature levels through multi-layer fusion is essential for robust OOD detection.

As the prototypes are updated for each batch, we compute the \(\text{RDS}_l\) for each selected layer \(l\).
To obtain the final OOD score for each sample \(\mathbf{x}_{t,i}\) in the batch, we fuse the RDS values from the selected layers \(\mathcal{L}\) by taking their \textit{average}, formally given by \(\text{RDS}_{\text{final}}(\mathbf{x}_{t,i}) = \frac{1}{|\mathcal{L}|} \sum_{l \in \mathcal{L}} \text{RDS}_l(\mathbf{x}_{t,i})\).
Using the fused OOD score \(\text{RDS}_{\text{final}}(\mathbf{x}_{t,i})\), we make the final OOD prediction.

\section{Experiments}
\label{sec:experiments}

\begin{table*}[t]
    \centering
    \caption{Main results of OOD detection on covariate-shifted data and clean data. 
    Results on covariate-shifted data are averaged over all 15 corruptions. Results on clean data are averaged over all 5 OOD datasets (Best: \textbf{bolded}, Second-best: \underline{underlined})}
    \vspace{-0.5em}
    \label{tab:results_cvs}
    \setlength{\tabcolsep}{3pt}
    \resizebox{0.95\linewidth}{!}{
    \begin{tabular}{l|c|c@{\hspace{2pt}}c|c@{\hspace{2pt}}c|c@{\hspace{2pt}}c|c@{\hspace{2pt}}c|c@{\hspace{2pt}}c||c@{\hspace{2pt}}c|||c@{\hspace{2pt}}c}
        \toprule
        \rowcolor{gray!15}
        \multicolumn{16}{c}{\textbf{ImageNet-C}} \\
        \midrule
        \multirow{3}{*}{Method} 
        & \multirow{3}{*}{\makecell[c]{Training set \\ prior}}
        & \multicolumn{12}{c|||}{Covariate shifted} & \multicolumn{2}{c}{Clean} \\
        \cmidrule(lr){3-16} 
        &
        & \multicolumn{2}{c|}{ImageNet-O-C} 
        & \multicolumn{2}{c|}{Places-C} 
        & \multicolumn{2}{c|}{SUN-C} 
        & \multicolumn{2}{c|}{iNaturalist-C} 
        & \multicolumn{2}{c||}{Textures-C} 
        & \multicolumn{2}{c|||}{Average} 
        & \multicolumn{2}{c}{Average} \\
        \cmidrule(lr){3-16} 
        &
        & FPR95 \(\downarrow\) & AUROC \(\uparrow\) 
        & FPR95 \(\downarrow\) & AUROC \(\uparrow\) 
        & FPR95 \(\downarrow\) & AUROC \(\uparrow\) 
        & FPR95 \(\downarrow\) & AUROC \(\uparrow\) 
        & FPR95 \(\downarrow\) & AUROC \(\uparrow\) 
        & FPR95 \(\downarrow\) & AUROC \(\uparrow\) 
        & FPR95 \(\downarrow\) & AUROC \(\uparrow\) \\
        \midrule
        MSP                         & \cellcolor{blue!10} NO 
        & 85.01 & 62.18 & 74.41 & 73.10 & 73.64 & 73.65 & 53.91 & 82.27 & 70.88 & 72.85 & 71.57 & 72.81 & 44.64 & 86.94 \\
        Energy                      & \cellcolor{blue!10} NO 
        & 82.68 & 63.65 & 70.13 & 77.66 & 69.70 & 78.79 & 59.45 & 82.48 & 62.58 & 78.16 & 68.91 & 76.15 & 38.25 & 85.52 \\
        Max logit                   & \cellcolor{blue!10} NO  
        & 83.69 & 63.51 & 71.00 & 76.64 & 70.11 & 77.59 & 53.19 & 83.76 & 65.85 & 76.85 & 68.77 & 75.67 & 37.05 & 86.43 \\
        GradNorm                    & \cellcolor{blue!10} NO 
        & 85.09 & 58.28 & 68.15 & 78.13 & 64.77 & 81.23 & 49.29 & 85.45 & \underline{51.70} & \underline{84.11} & 63.80 & 77.44 & 80.50 & 57.26 \\
        ODIN                        & \cellcolor{blue!10} NO 
        & 86.67 & 59.91 & \underline{57.03} & \underline{83.67} & \underline{54.88} & \underline{84.53} & 44.03 & \underline{88.03} & 53.95 & 83.93 & \underline{59.31} & \underline{80.01} & 60.84 & 81.86 \\
        SCALE                       & \cellcolor{blue!10} NO 
        & \underline{82.06} & 65.18 & 67.20 & 79.03 & 65.58 & 80.33 & 48.40 & 85.75 & 60.80 & 80.03 & 64.81 & 78.06 & \underline{35.01} & 87.02 \\
        ASH                         & \cellcolor{blue!10} NO 
        & 82.45 & 63.90 & 69.82 & 77.91 & 69.26 & 79.08 & 58.88 & 82.76 & 62.00 & 78.55 & 68.48 & 76.44 & 37.77 & 85.77 \\
        RTL                         & \cellcolor{blue!10} NO 
        & 83.70 & 65.26 & 65.44 & 79.96 & 64.27 & 80.32 & \underline{41.39} & 87.69 & 64.08 & 78.36 & 63.78 & 78.32 & 42.30 & 84.24 \\
        ViM                         & \cellcolor{red!10} YES 
        & 91.28 & 54.02 & 98.75 & 25.72 & 99.14 & 23.46 & 99.41 & 18.87 & 97.90 & 29.31 & 97.30 & 30.28 & 56.58 & 87.60 \\
        KNN                         & \cellcolor{red!10} YES 
        & 92.60 & 53.73 & 98.17 & 30.68 & 98.23 & 29.94 & 98.69 & 24.10 & 90.21 & 50.39 & 95.58 & 37.77 & 84.79 & 75.87 \\
        MDS\textsubscript{single}    & \cellcolor{red!10} YES 
        & 92.67 & 50.34 & 99.11 & 21.17 & 99.40 & 18.89 & 99.68 & 14.49 & 98.29 & 26.37 & 97.83 & 26.25 & 79.19 & 80.57 \\
        MDS\textsubscript{ensemble}  & \cellcolor{red!10} YES 
        & 85.89 & 58.76 & 98.08 & 27.51 & 98.23 & 26.49 & 98.31 & 21.40 & 81.15 & 47.73 & 92.33 & 36.38 & 55.30 & 86.86 \\
        ReAct                       & \cellcolor{red!10} YES 
        & 90.46 & 54.69 & 89.30 & 66.71 & 88.96 & 68.19 & 92.42 & 64.03 & 86.26 & 66.59 & 89.48 & 64.04 & 86.40 & 70.67 \\
        NNGuide                     & \cellcolor{red!10} YES 
        & 88.94 & 56.31 & 81.97 & 69.88 & 81.05 & 73.03 & 74.75 & 76.53 & 58.93 & 79.60 & 77.13 & 71.07 & 70.60 & 70.05 \\
        CoRP                        & \cellcolor{red!10} YES  
        & 93.02 & 52.30 & 95.18 & 46.32 & 95.22 & 46.33 & 95.99 & 43.54 & 93.05 & 51.72 & 94.49 & 48.04 & 49.03 & 90.17 \\
        MDS++                       & \cellcolor{red!10} YES  
        & 82.39 & 64.35 & 90.03 & 52.53 & 90.68 & 52.72 & 64.23 & 71.61 & 66.93 & 68.57 & 78.85 & 61.96 & 41.87 & 90.48 \\
        RMDS                        & \cellcolor{red!10} YES  
        & 92.91 & 58.76 & 96.74 & 48.73 & 97.50 & 47.40 & 95.35 & 53.60 & 96.64 & 44.86 & 95.83 & 50.69 & 46.28 & \underline{91.03} \\
        RMDS++                      & \cellcolor{red!10} YES  
        & 94.61 & 57.05 & 96.02 & 52.76 & 96.38 & 51.52 & 93.87 & 64.11 & 95.33 & 52.20 & 95.24 & 55.53 & 56.01 & 89.73 \\
        OODD                        & \cellcolor{red!10} YES  
        & 80.12 & \underline{65.83} & 82.45 & 66.56 & 77.69 & 70.18 & 45.45 & 85.79 & 52.42 & 80.17 & 67.63 & 73.71 & 74.31 & 86.93 \\
        \midrule
        \rowcolor{blue!5}
        \ours                       & \cellcolor{blue!10} NO  
        & \textbf{68.87} & \textbf{66.63} & \textbf{8.70}  & \textbf{92.96} & \textbf{8.08}  & \textbf{93.03} & \textbf{7.60}  & \textbf{93.18} & \textbf{2.55}  & \textbf{99.43} & \textbf{19.16} & \textbf{89.05} & \textbf{15.95} & \textbf{96.13} \\
        \midrule
        \rowcolor{gray!15}
        \multicolumn{16}{c}{\textbf{CIFAR-100-C}} \\
        \midrule
        \multirow{3}{*}{Method} 
        & \multirow{3}{*}{\makecell[c]{Training set \\ prior}}
        & \multicolumn{12}{c|||}{Covariate shifted} & \multicolumn{2}{c}{Clean} \\
        \cmidrule(lr){3-16} 
        &
        & \multicolumn{2}{c|}{SVHN-C} 
        & \multicolumn{2}{c|}{Places365-C} 
        & \multicolumn{2}{c|}{LSUN-C} 
        & \multicolumn{2}{c|}{iSUN-C} 
        & \multicolumn{2}{c||}{Textures-C} 
        & \multicolumn{2}{c|||}{Average}
        & \multicolumn{2}{c}{Average} \\
        \cmidrule(lr){3-16} 
        & & FPR95 \(\downarrow\) & AUROC \(\uparrow\) 
        & FPR95 \(\downarrow\) & AUROC \(\uparrow\) 
        & FPR95 \(\downarrow\) & AUROC \(\uparrow\) 
        & FPR95 \(\downarrow\) & AUROC \(\uparrow\) 
        & FPR95 \(\downarrow\) & AUROC \(\uparrow\) 
        & FPR95 \(\downarrow\) & AUROC \(\uparrow\) 
        & FPR95 \(\downarrow\) & AUROC \(\uparrow\) \\
        \midrule
        MSP                         & \cellcolor{blue!10} NO 
        & 91.48 & 58.40 & 87.30 & 64.23 & 86.85 & 64.49 & 88.40 & 62.74 & 91.91 & 58.24 & 89.19 & 61.62 & 80.37 & 75.29 \\
        Energy                      & \cellcolor{blue!10} NO 
        & 93.13 & 60.53 & 84.74 & 66.56 & 84.04 & 68.57 & 87.14 & 64.79 & 90.20 & 61.10 & 87.85 & 64.31 & 79.95 & 76.79 \\
        Max logit                   & \cellcolor{blue!10} NO  
        & 92.73 & 60.58 & 85.04 & 66.43 & 84.43 & 68.24 & 87.15 & 64.74 & 90.74 & 60.78 & 88.02 & 64.15 & 79.71 & 77.06 \\
        GradNorm                    & \cellcolor{blue!10} NO 
        & 96.07 & 47.75 & 92.56 & 52.99 & 94.89 & 45.14 & 94.79 & 45.32 & 84.59 & 62.52 & 92.58 & 50.74 & 94.92 & 43.62 \\
        ODIN                        & \cellcolor{blue!10} NO 
        & 79.07 & 70.05 & 88.23 & 63.54 & 89.96 & 59.13 & 90.48 & 59.18 & 77.69 & 70.79 & 85.09 & 64.54 & 81.11 & 69.50 \\
        SCALE                       & \cellcolor{blue!10} NO 
        & 88.88 & 66.46 & 85.55 & 66.91 & 86.30 & 65.89 & 88.06 & 64.21 & 81.14 & 70.69 & 85.99 & 66.83 & 76.60 & 77.53 \\
        ASH                         & \cellcolor{blue!10} NO 
        & 92.05 & 62.73 & 85.42 & 66.48 & 85.24 & 67.78 & 87.99 & 64.30 & 87.66 & 64.61 & 87.67 & 65.18 & 80.44 & 76.76 \\
        RTL                         & \cellcolor{blue!10} NO 
        & 89.27 & 58.60 & 84.59 & 64.62 & 83.65 & 66.99 & 86.90 & 63.60 & 89.30 & 57.65 & 86.74 & 62.29 & 64.11 & 80.76 \\
        ViM                         & \cellcolor{red!10} YES 
        & 77.24 & 72.64 & 90.23 & 59.74 & 86.41 & 64.41 & 87.65 & 62.40 & 93.15 & 55.11 & 86.94 & 62.86 & 71.94 & 75.61 \\
        KNN                         & \cellcolor{red!10} YES 
        & 88.45 & 66.38 & 85.63 & 63.99 & 83.87 & 70.22 & 87.06 & 65.25 & 90.83 & 60.93 & 87.17 & 65.35 & 71.48 & 81.16 \\
        MDS\textsubscript{single}   & \cellcolor{red!10} YES 
        & 89.04 & 61.64 & 91.30 & 57.42 & 84.94 & 65.60 & 86.68 & 62.82 & 95.16 & 48.58 & 89.42 & 59.21 & 80.01 & 69.98 \\
        MDS\textsubscript{ensemble} & \cellcolor{red!10} YES 
        & 63.57 & 79.33 & 92.49 & 48.49 & 73.78 & 60.77 & 73.95 & 58.87 & 75.16 & 57.51 & 75.79 & 60.99 & 45.94 & 86.36 \\
        ReAct                       & \cellcolor{red!10} YES 
        & 95.08 & 55.32 & 88.78 & 62.51 & 86.39 & 67.46 & 88.07 & 64.76 & 91.90 & 60.40 & 90.04 & 62.09 & 79.45 & 76.09 \\
        NNGuide                     & \cellcolor{red!10} YES 
        & 91.11 & 62.78 & 88.78 & 63.45 & 91.30 & 58.64 & 91.96 & 57.62 & 77.91 & 68.69 & 88.21 & 62.24 & 87.55 & 66.29 \\
        CoRP                        & \cellcolor{red!10} YES
        & 64.51 & \underline{80.52} & 89.98 & 60.28 & 92.56 & 59.80 & 91.30 & 59.91 & 55.25 & \underline{81.49} & 78.72 & 68.40 & 65.54 & 81.05 \\
        MDS++                       & \cellcolor{red!10} YES
        & 95.15 & 62.58 & 87.40 & 64.87 & 87.25 & 66.75 & 89.85 & 62.90 & 95.16 & 59.59 & 90.96 & 63.34 & 85.96 & 75.52 \\
        RMDS                        & \cellcolor{red!10} YES
        & 88.78 & 64.07 & 86.67 & 64.65 & 79.57 & 71.82 & 82.68 & 67.91 & 94.23 & 55.06 & 86.39 & 64.70 & 67.90 & 82.51 \\
        RMDS++                      & \cellcolor{red!10} YES
        & 91.52 & 65.51 & 85.70 & 65.59 & 82.73 & 70.26 & 86.08 & 66.38 & 93.78 & 58.41 & 87.96 & 65.23 & 73.91 & 81.57 \\
        OODD                        & \cellcolor{red!10} YES
        & \textbf{43.11} & \textbf{87.42} & \underline{73.13} & \textbf{76.16} & \underline{51.51} & \textbf{85.97} & \underline{56.87} & \textbf{82.84} & \textbf{47.54} & \textbf{85.40} & \underline{54.43} & \textbf{83.56} & \underline{33.19} & \textbf{91.93} \\
        \midrule
        \rowcolor{blue!5}
        \ours                       & \cellcolor{blue!10} NO  
        & \underline{48.60} & 79.82 & \textbf{68.66} & \underline{68.00} & \textbf{44.14} & \underline{80.29} & \textbf{50.76} & \underline{79.75} & \underline{51.48} & 80.60 & \textbf{52.73} & \underline{77.69} & \textbf{33.12} & \underline{89.78} \\
        \bottomrule
    \end{tabular}
    }
    \vspace{-0.5em}
\end{table*}

\subsection{Experimental Setup}

\subsubsection{Evaluation Details.} 
Although \ours~is designed to adaptively handle challenging covariate shifts, we evaluate on both clean and covariate-shifted datasets as the shift may be weak or absent in real-world scenarios.
We use CIFAR-100~\cite{krizhevsky2009learning} and ImageNet~\cite{deng2009imagenet} as ID datasets.  
For CIFAR-100, we use SVHN~\cite{netzer2011reading}, Places365~\cite{zhou2017places}, LSUN~\cite{yu2015lsun}, iSUN~\cite{xu2015turkergaze}, and Textures~\cite{cimpoi2014describing} as OOD sets.
For ImageNet, we use ImageNet-O~\cite{hendrycks2021natural}, Places~\cite{zhou2017places}, SUN~\cite{xiao2010sun}, iNaturalist~\cite{van2018inaturalist}, and Textures.
To simulate covariate shift, we apply 15 common corruption types~\cite{hendrycks2019benchmarking} at severity level 5 to both ID and OOD datasets, resulting in pairs such as CIFAR-100-C vs. SVHN-C.
In our main evaluations, each test-time batch contains an equal number of ID/OOD samples, and we report results under varying ID/OOD ratios in Appendix.
For CIFAR-100-based benchmarks, we use WideResNet-40-2~\cite{zagoruyko2016wide} pre-trained with AugMix~\cite{hendrycks2019augmix} on CIFAR-100, which is available from RobustBench~\cite{croce2020robustbench}. 
For ImageNet-based benchmarks, we use the pre-trained RegNetY-16GF~\cite{radosavovic2020designing} available from PyTorch.
In addition, we also evaluate with Transformer-based models on CIFAR-100, all of which are fine-tuned with CIFAR-100. 
Results with pre-trained ResNet-50~\cite{he2016deep} on ImageNet are also provided in Appendix.

\subsubsection{Baseline Methods.} 
%
MSP~\cite{hendrycks2016baseline}, Max Logit~\cite{hendrycks2019scaling}, Energy~\cite{liu2020energy}, ODIN~\cite{liang2018enhancing}, GradNorm~\cite{huang2021importance}, SCALE~\cite{xu2023scaling}, ASH~\cite{djurisic2022extremely}, and RTL~\cite{fan2024test}, like \ours, do not require any precomputed statistics or storage from the training data, whereas KNN~\cite{sun2022out}, ViM~\cite{wang2022vim}, ReAct~\cite{sun2021react}, NNGuide~\cite{park2023nearest}, CoRP~\cite{fang2024kernel}, OODD~\cite{yang2025oodd}, MDS~\cite{lee2018simple} and its variants~\cite{ren2021simple, mueller2025mahalanobis++} do.
As MDS has been implemented using either single layer or multi layers in prior work, we compare against both variants.

\subsection{Main OOD Detection Results}
In \cref{tab:results_cvs}, we compare \ours~with the other OOD detection methods on covariate-shifted data and clean data.  
As \ours~is designed to adapt to test-time covariate shift, it demonstrates its full potential in the covariate-shifted setting.
On the ImageNet-C benchmark, \ours~achieves the best performance on every OOD dataset and both evaluation metrics, with an average FPR@95TPR (FPR95) reduction of \textbf{40.15 pp} and average AUROC gain of \textbf{9.04 pp} compared to the second-best (\ie, ODIN). 
%
On CIFAR-100-C, \ours~again achieves the best FPR95 while requiring substantially less additional computation and test-time memory than OODD (Appendix \cref{tab:additional_computation_and_state}), the second-best method.
These results highlight the robustness and adaptability of \ours~in the face of test-time covariate shift.
An important observation is that methods using prior information from the training set tend to perform similarly to, or \textit{even worse} 
than, baselines that do not use such priors. This suggests that the prior information that is typically beneficial for OOD detection on clean datasets may become misaligned with test-time distributions under covariate shift, leading to degraded performance.

Although \ours~mainly targets covariate-shifted environments, it retains strong performance on clean data, achieving the best FPR95 on both benchmarks and the highest AUROC on ImageNet.
For example, as shown in the clean-data columns of \cref{tab:results_cvs}, \ours~reduces FPR95 by \textbf{19.06 pp} compared to SCALE on ImageNet.
%
These results suggest that ID and OOD samples are well-separated in the feature space and that \ours~finely adjusts the ID and OOD prototypes toward the optimal axis for discrimination. 
Notably, baselines that leverage prior information from the training set tend to outperform those not relying on such information. 
\ours~defies this trend, performing the best without any prior information.



\subsection{Generalization across Backbones}
We further show that \ours~generalizes beyond conventionally trained CNN backbones, with the discriminative axis emerging across model architectures and training objectives.

\subsubsection{Results with Transformer Architectures.}
We also evaluate with transformer-based architectures~\cite{vaswani2017attention}, specifically ViT-Tiny~\cite{vit,vit-tiny} and Swin-T~\cite{liu2021swin}, to verify robustness across different model architectures. 
As shown in \cref{tab:results_transformer}, \ours~outperforms all baselines, yielding an FPR95 drop of \textbf{17.45 pp} (covariate-shifted) and \textbf{8.6 pp} (clean) against the second-best method (\ie, MDS) with ViT-Tiny. Similarly, for Swin-T, \ours~reduces FPR95 by \textbf{37.11 pp} (covariate-shifted) and \textbf{3.42 pp} (clean) relative to MDS++.
This proves the superiority of \ours~and the emergence of the discriminative axis regardless of the model architecture.

\begin{table}[t]
    \centering
    \caption{OOD detection performance comparison with ViT-Tiny and Swin-T architectures. All results are reported as the mean over all five OOD datasets. For covariate shifted datasets, results are the average of all 15 corruptions with severity level 5. (Best: \textbf{bolded}, Second-best: \underline{underlined})}
    \vspace{-0.5em}
    \label{tab:results_transformer}
    \setlength{\tabcolsep}{2pt}
    \resizebox{\linewidth}{!}{
    \begin{tabular}{l|cc|cc|cc|cc}
        \toprule
        \multirow{3}{*}{Method} 
        & \multicolumn{4}{c|}{Covariate shifted} 
        & \multicolumn{4}{c}{Clean} \\
        \cmidrule(lr){2-9} 
        & \multicolumn{2}{c|}{ViT-Tiny} 
        & \multicolumn{2}{c|}{Swin-T} 
        & \multicolumn{2}{c|}{ViT-Tiny} 
        & \multicolumn{2}{c}{Swin-T} \\
        \cmidrule(lr){2-9} 
        & FPR95 \(\downarrow\) & AUROC \(\uparrow\) 
        & FPR95 \(\downarrow\) & AUROC \(\uparrow\) 
        & FPR95 \(\downarrow\) & AUROC \(\uparrow\) 
        & FPR95 \(\downarrow\) & AUROC \(\uparrow\) \\
        \midrule
        MSP                         & 89.56 & 57.03 & 85.42 & 63.39 & 70.36 & 79.79 & 60.16 & 84.31 \\
        Energy                      & 87.01 & 61.20 & 80.30 & 68.35 & 58.77 & 84.83 & 41.06 & 89.80 \\
        Max Logit                   & 87.58 & 60.65 & 81.90 & 67.73 & 60.27 & 84.54 & 42.11 & 89.52 \\
        GradNorm                    & 89.54 & 58.42 & 78.77 & 71.44 & 78.33 & 75.24 & 71.63 & 74.53 \\
        ODIN                        & 87.53 & 60.52 & 84.88 & 63.63 & 80.25 & 71.22 & 91.20 & 63.65 \\
        SCALE                       & 88.84 & 58.99 & 76.68 & 71.57 & 61.75 & 83.66 & 44.84 & 88.50 \\
        ASH                         & 92.00 & 56.00 & 76.95 & 71.39 & 81.17 & 74.00 & 50.03 & 86.97 \\
        RTL                         & 84.42 & 58.69 & 83.71 & 63.62 & 40.20 & 87.75 & 50.90 & 84.23 \\
        ViM                         & 89.49 & 58.75 & 96.94 & 45.76 & 58.27 & 85.06 & 93.00 & 66.26 \\
        KNN                         & 90.15 & 56.81 & 84.25 & 64.91 & 67.35 & 79.58 & 45.93 & 88.51 \\
        MDS\textsubscript{single}   & 97.22 & 32.40 & 98.76 & 33.09 & 96.38 & 35.57 & 98.67 & 41.11 \\
        MDS\textsubscript{ensemble} & \underline{61.73} & 65.95 & 98.60 & 33.75 & \underline{29.18} & 88.40 & 98.42 & 45.53 \\
        ReAct                       & 85.94 & 62.01 & 76.76 & 71.14 & 56.86 & 81.79 & 41.78 & 89.51 \\
        NNGuide                     & 86.49 & 61.17 & 76.53 & \underline{73.38} & 60.30 & 83.43 & 38.04 & \underline{91.06} \\
        CoRP                        & 88.31 & 59.85 & 86.17 & 63.65 & 67.28 & 82.43 & 50.37 & 88.53 \\
        MDS++                       & 78.15 & 66.25 & \underline{76.01} & 69.80 & 46.70 & 87.12 & \underline{35.28} & \textbf{91.54} \\
        RMDS                        & 82.63 & 63.56 & 91.33 & 58.14 & 47.98 & 86.93 & 54.00 & 87.65 \\
        RMDS++                      & 81.30 & 64.05 & 88.13 & 61.81 & 48.11 & 86.71 & 47.01 & 88.90 \\
        OODD                        & 65.58 & \textbf{72.43} & 86.69 & 53.79 & 31.50 & \underline{91.09} & 78.32 & 79.18 \\
        \midrule
        \rowcolor{blue!5}
        \ours                       & \textbf{44.28} & \underline{68.60} & \textbf{38.90} & \textbf{81.28} & \textbf{20.58} & \textbf{94.31} & \textbf{31.86} & 88.03 \\
        \bottomrule
    \end{tabular}
    }
\end{table}

\begin{table}[t]
    \centering
    \caption{OOD detection on a vision-language backbone (CLIP-ViT-B/16, ImageNet ID). All results are the mean over all five OOD datasets; covariate-shifted results average all 15 corruptions at severity level 5. (Best: \textbf{bolded}, Second-best: \underline{underlined})}
    \vspace{-0.5em}
    \label{tab:results_vlm}
    \resizebox{0.75\linewidth}{!}{
    \begin{tabular}{l|cc|cc}
        \toprule
        \multirow{2}{*}{Method}
        & \multicolumn{2}{c|}{Covariate shifted}
        & \multicolumn{2}{c}{Clean} \\
        \cmidrule(lr){2-3} \cmidrule(lr){4-5}
        & FPR95 \(\downarrow\) & AUROC \(\uparrow\)
        & FPR95 \(\downarrow\) & AUROC \(\uparrow\) \\
        \midrule
        MCM     & 63.08 & 81.43 & 14.40 & 96.39 \\
        GL-MCM  & \underline{57.45} & 82.85 & 22.28 & 93.75 \\
        LoCoOp  & 58.02 & \underline{84.18} & 10.73 & \underline{97.14} \\
        AdaNeg  & 66.56 & 77.14 & \underline{10.54} & 95.64 \\
        \midrule
        \rowcolor{blue!5}
        \ours   & \textbf{8.76} & \textbf{96.30} & \textbf{1.40} & \textbf{99.65} \\
        \bottomrule
    \end{tabular}
    }
    \vspace{-0.5em}
\end{table}

\subsubsection{Results on Vision-Language Backbones.}
To examine whether the discriminative axis emerges beyond conventionally trained vision models, we evaluate \ours~on a frozen CLIP-ViT-B/16 backbone against recent vision-language OOD detectors (MCM~\cite{ming2022mcm}, GL-MCM~\cite{miyai2025glmcm}, LoCoOp~\cite{miyai2023locoop}, and AdaNeg~\cite{zhang2024adaneg}).
Since CLIP exposes no softmax classifier, we replace MSP with MCM for first-batch pseudo-labeling.
As shown in \cref{tab:results_vlm}, \ours~improves over the best baseline by \textbf{+12.12 pp} AUROC under covariate shift and by \textbf{+2.51 pp} on clean data, indicating that the discriminative axis is consistent not only across model architectures but also across training objectives.

\subsection{Cross-domain Evaluation Results}
In the previous experiments, each test stream contains a single OOD dataset and a single type of covariate shift.
We further evaluate \ours~in more challenging regimes with multiple OOD datasets and multiple covariate shifts.
All cross-domain evaluations use ImageNet as the ID and ResNet-50 as the backbone.

\subsubsection{Mixed OOD.}

We believe our bounded OOD setting reflects realistic deployment scenarios where OOD inputs tend to concentrate within a limited semantic space due to observation boundaries. 
However, to further show robustness \textit{beyond this assumption}, we conduct additional experiments where two different OOD sources are \textit{mixed}. 
At each batch, OOD samples are drawn from two diverse sources rather than a single dataset.
In~\cref{tab:mixed_ood}, \ours~remains best even under mixed OOD, indicating that our OOD prototype successfully finds the discriminative axis even when OOD samples span multiple semantic categories. 
We observe performance degradation only in an extreme case where all five OOD sets are mixed (68.08\% AUROC). 
However, we believe such extreme mixing rarely occurs in practice, and our assumption holds within realistic deployment boundaries.

\subsubsection{Mixed Covariate Shift.}

We further test streams where each batch contains data from multiple covariate shifts, considering mixtures of two shifts and mixture of all 15 shifts. As shown in~\cref{tab:mixed_cvs}, \ours~consistently surpasses all baselines under mixed shifts, remaining best even in the extreme mixture-of-15 setting.

\begin{table}[t]
    \centering
    \caption{Mixed OOD evaluation}
    \vspace{-0.5em}
    \resizebox{\linewidth}{!}{
    \begin{tabular}{l|cc|cc|cc}
        \toprule
        \multirow{2}{*}{Method} & \multicolumn{2}{c|}{ImageNet-O + Places} & \multicolumn{2}{c|}{Places + SUN} & \multicolumn{2}{c}{SUN + Textures} \\
        \cmidrule(lr){2-3} \cmidrule(lr){4-5} \cmidrule(lr){6-7}
        & FPR95 \(\downarrow\) & AUROC \(\uparrow\) & FPR95 \(\downarrow\) & AUROC \(\uparrow\) & FPR95 \(\downarrow\) & AUROC \(\uparrow\) \\
        \midrule
        MSP      & 48.43 & 83.70 & 48.65 & 84.26 & 43.65 & 85.52 \\
        Energy   & 51.56 & 78.42 & 44.24 & 82.57 & 38.31 & 85.65 \\
        Max Logit& 48.14 & 80.20 & 42.85 & 83.49 & 37.29 & 86.24 \\
        GradNorm & 86.49 & 43.25 & 77.26 & 57.44 & 67.54 & 66.80 \\
        ODIN     & 62.28 & 79.44 & 66.10 & 76.78 & 60.37 & 80.13 \\
        SCALE    & 47.62 & 80.09 & 40.36 & 84.39 & 33.92 & 87.81 \\
        RTL      & 46.82 & 81.46 & 44.77 & 82.94 & 40.64 & 83.98 \\
        \midrule
        \rowcolor{blue!5}
        DART     & \textbf{19.23} & \textbf{94.45} & \textbf{25.96} & \textbf{91.15} & \textbf{18.65} & \textbf{94.80} \\
        \bottomrule
    \end{tabular}
    }
    \label{tab:mixed_ood}
    \vspace{-0.5em}
\end{table}

\begin{table}[t]
    \centering
    \caption{Mixed covariate shift evaluation}
    \vspace{-0.5em}
    \setlength{\tabcolsep}{2pt}
    \resizebox{\linewidth}{!}{
    \begin{tabular}{l|cc|cc|cc|cc}
        \toprule
        \multirow{2}{*}{Method} & \multicolumn{2}{c|}{Clean + Gauss.} & \multicolumn{2}{c|}{Gauss. + Snow} & \multicolumn{2}{c|}{Snow + JPEG} & \multicolumn{2}{c}{All 15 corruptions} \\
        \cmidrule(lr){2-3} \cmidrule(lr){4-5} \cmidrule(lr){6-7} \cmidrule(lr){8-9}
        & FPR95 \(\downarrow\) & AUROC \(\uparrow\) & FPR95 \(\downarrow\) & AUROC \(\uparrow\) & FPR95 \(\downarrow\) & AUROC \(\uparrow\) & FPR95 \(\downarrow\) & AUROC \(\uparrow\) \\
        \midrule
        MSP       & 82.63 & 67.43 & 85.01 & 62.49 & 83.07 & 68.22 & 86.64 & 62.71 \\
        Energy    & 72.01 & 76.77 & 83.91 & 69.82 & 89.72 & 66.83 & 86.27 & 63.02 \\
        Max Logit & 77.19 & 73.34 & 85.40 & 67.10 & 85.73 & 68.66 & 86.46 & 63.67 \\
        GradNorm  & 54.72 & 83.51 & 65.33 & 81.94 & 72.16 & 79.49 & 69.55 & 77.45 \\
        ODIN      & 66.85 & 77.72 & 45.20 & 83.46 & 12.08 & 97.39 & 25.45 & 92.16 \\
        SCALE     & 66.99 & 78.21 & 80.27 & 71.68 & 79.15 & 73.94 & 78.20 & 71.19 \\
        RTL       & 79.84 & 67.73 & 81.92 & 65.32 & 79.86 & 71.65 & 81.27 & 65.63 \\
        \midrule
        \rowcolor{blue!5}
        DART      & \textbf{22.04} & \textbf{93.93} & \textbf{0.85} & \textbf{99.62} & \textbf{0.79} & \textbf{99.72} & \textbf{19.68} & \textbf{92.70} \\
        \bottomrule
    \end{tabular}
    }
    \label{tab:mixed_cvs}
    \vspace{-0.5em}
\end{table}


\subsubsection{Continual OOD.}

We also consider a continual-OOD stream where the OOD source \textit{changes} abruptly across temporal segments while the ID stream remains fixed.
Since this setting is designed to evaluate online adaptation capability, we only include methods that adapt online.
Results in \cref{fig:continual_ood} show that \ours~remains stable across segments, with only modest fluctuations at switch points, indicating that our prototype-based tracking can re-estimate the relevant axis over time.

\subsubsection{Continual Covariate Shift.}

\begin{figure}[t]
    \centering
    \hspace{0.2em}
    \begin{minipage}[t]{0.45\linewidth}
        \centering

        \begin{subfigure}[t]{\linewidth}
            \centering
            \caption*{IN-O $\rightarrow$ Places}
            \includegraphics[width=\linewidth]{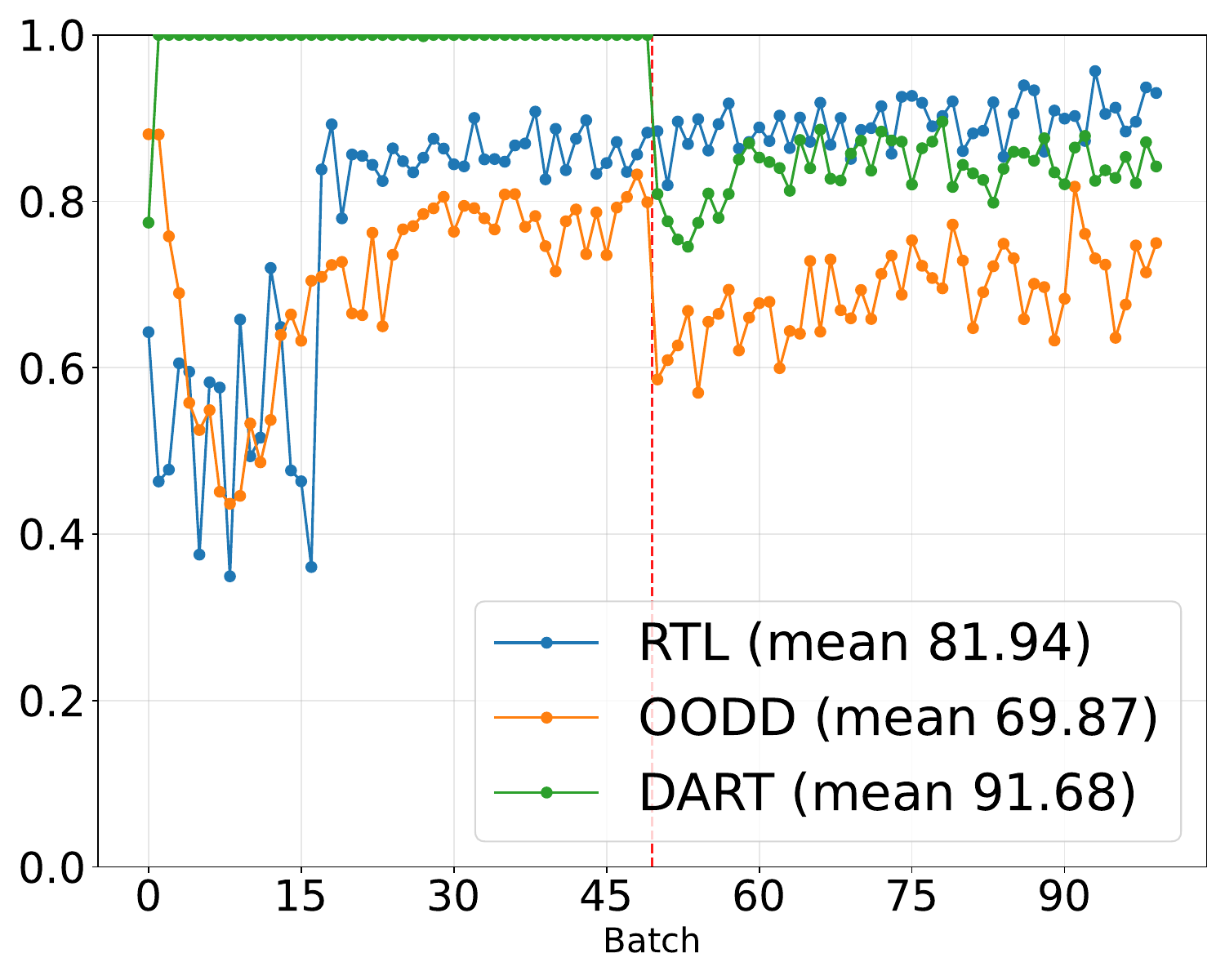}
            \label{fig:cont_ino_places}
        \end{subfigure}

        \vspace{-1.2em}

        \begin{subfigure}[t]{\linewidth}
            \centering
            \caption*{Places $\rightarrow$ SUN}
            \includegraphics[width=\linewidth]{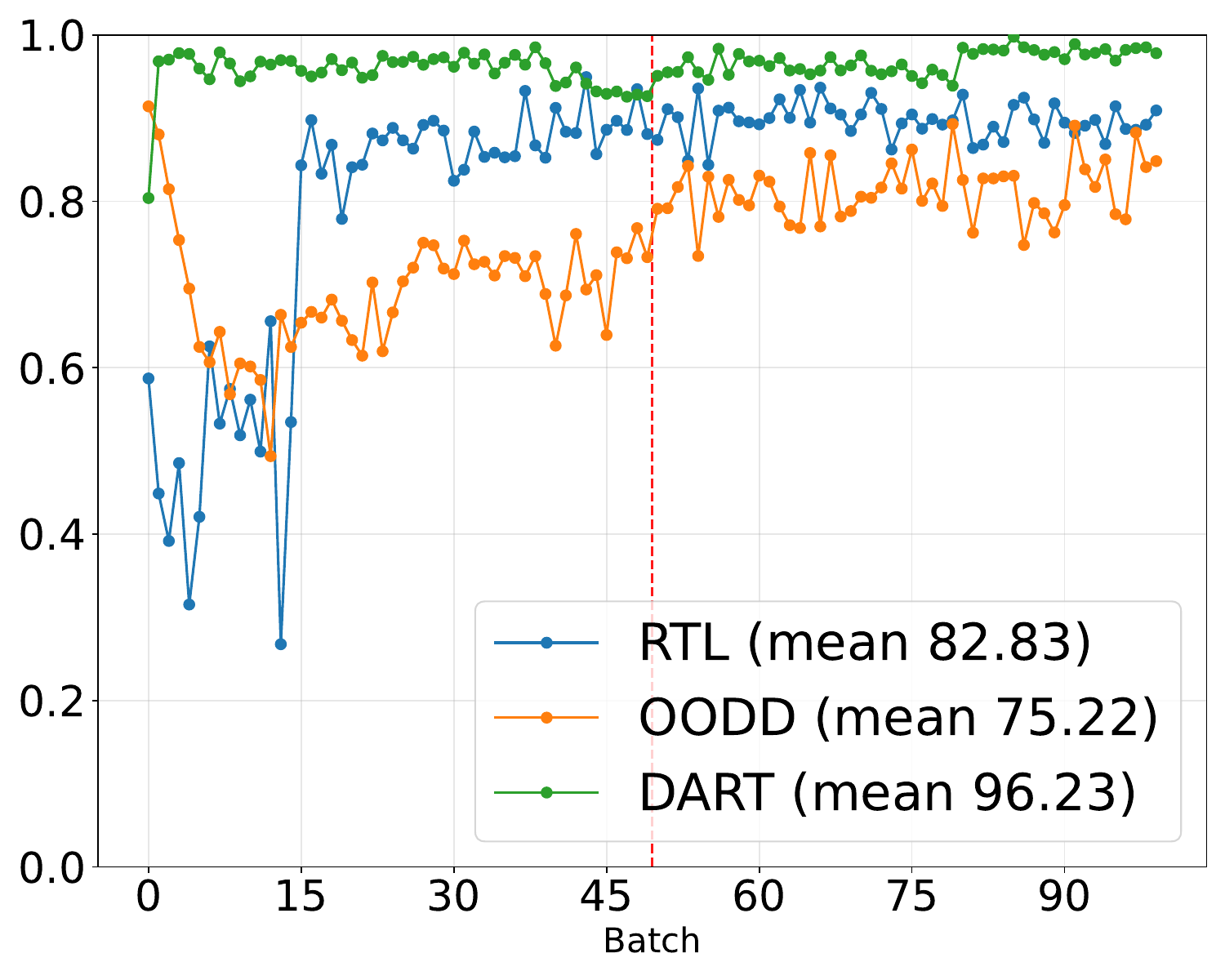}
            \label{fig:cont_places_sun}
        \end{subfigure}

        \vspace{-1.2em}

        \begin{subfigure}[t]{\linewidth}
            \centering
            \caption*{SUN $\rightarrow$ Textures}
            \includegraphics[width=\linewidth]{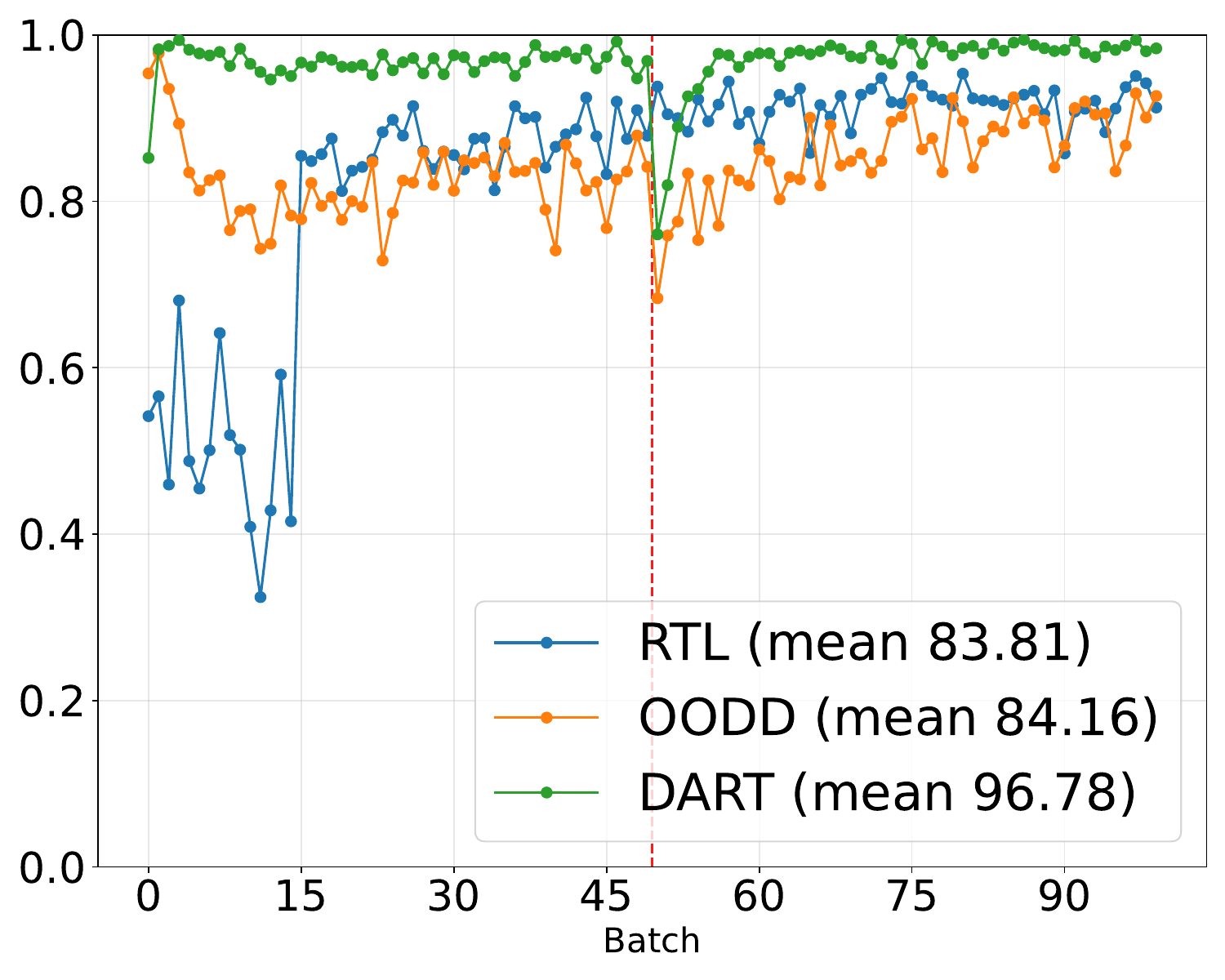}
            \label{fig:cont_sun_textures}
        \end{subfigure}
        \vspace{-1.2em}
        \subcaption{Continual OOD}
        \label{fig:continual_ood}
    \end{minipage}
    \hfill
    \begin{minipage}[t]{0.01\textwidth}
        \centering
        \vspace{1em}
        \hspace{-0.7em}
        \rule{0.1pt}{60\textwidth}
    \end{minipage}%
    \begin{minipage}[t]{0.45\linewidth}
        \centering

        \begin{subfigure}[t]{\linewidth}
            \centering
            \caption*{Clean $\rightarrow$ Gauss.}
            \includegraphics[width=\linewidth]{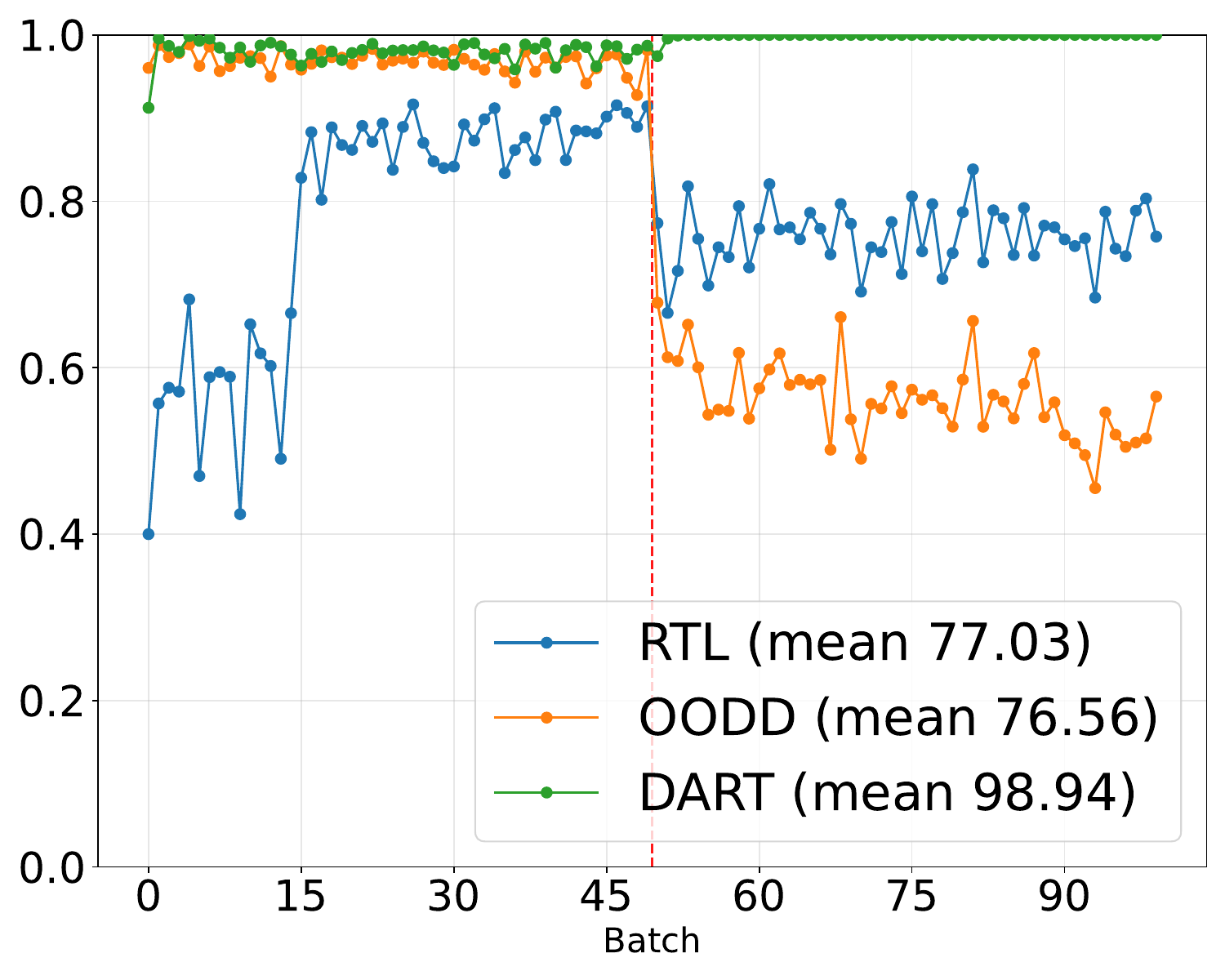}
            \label{fig:cont_clean_gaussian}
        \end{subfigure}

        \vspace{-1.2em}

        \begin{subfigure}[t]{\linewidth}
            \centering
            \caption*{Gauss. $\rightarrow$ Snow}
            \includegraphics[width=\linewidth]{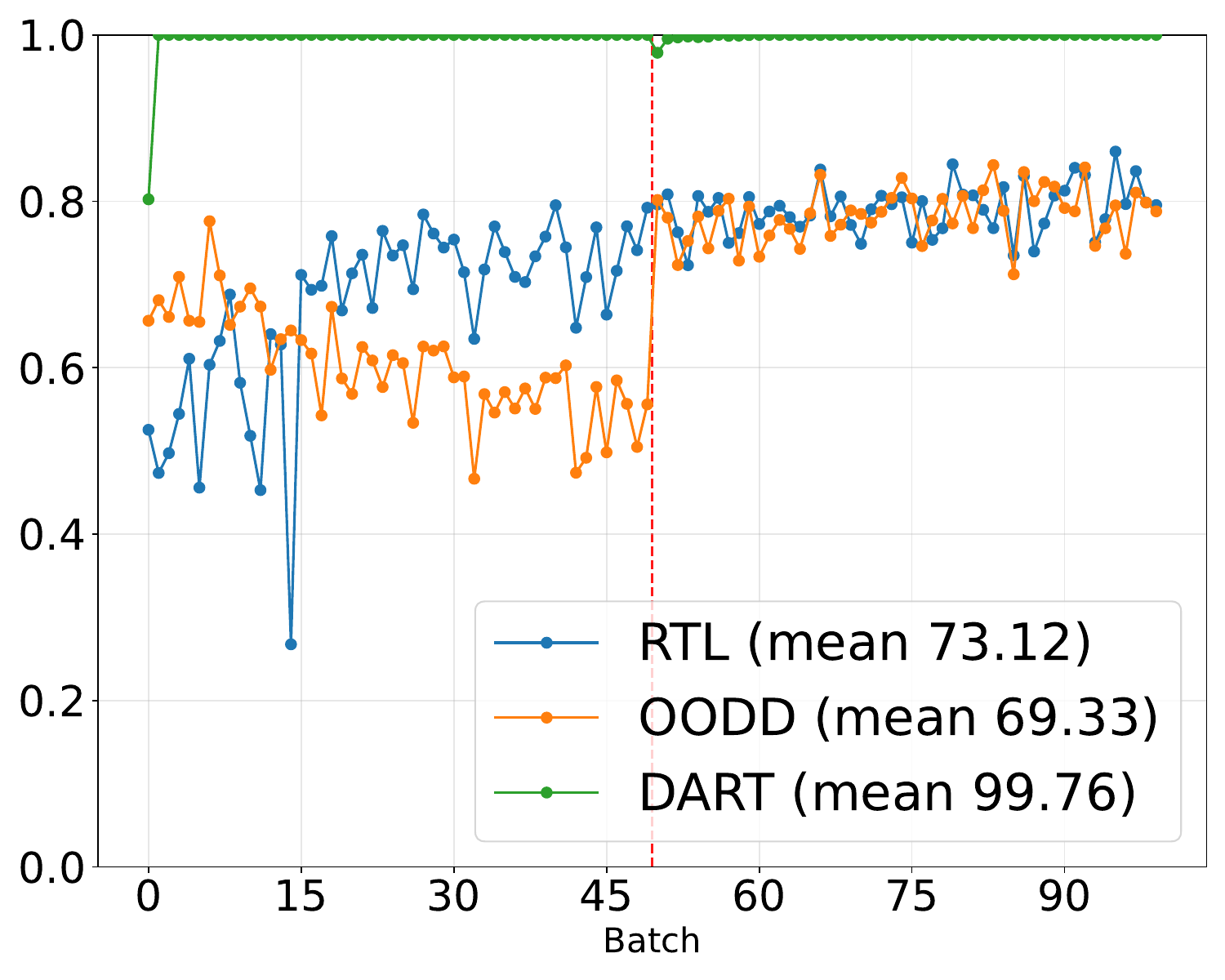}
            \label{fig:cont_gaussian_snow}
        \end{subfigure}

        \vspace{-1.2em}

        \begin{subfigure}[t]{\linewidth}
            \centering
            \caption*{Snow $\rightarrow$ JPEG}
            \includegraphics[width=\linewidth]{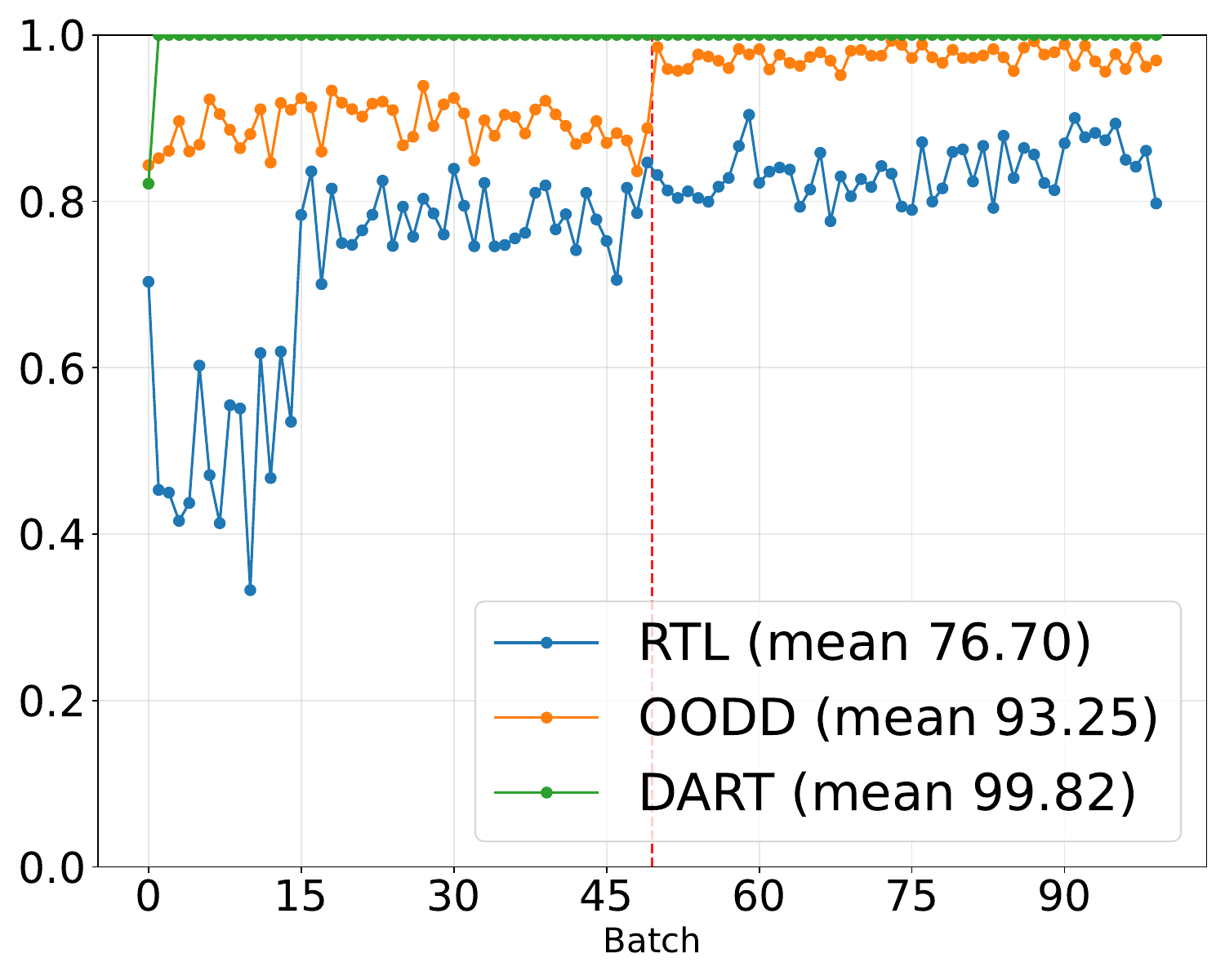}
            \label{fig:cont_snow_jpeg}
        \end{subfigure}
        \vspace{-1.2em}
        \subcaption{Continual covariate shift}
        \label{fig:continual_cvs}
    \end{minipage}
    \hfill
    \vspace{-0.5em}
    \caption{Continual evaluation results. Left column: continual OOD evaluation. Right column: continual covariate shift evaluation. AUROC vs. batch.}
    \label{fig:continual_combined}
    \vspace{-0.5em}
\end{figure}

Finally, we evaluate continual covariate shift where the corruption type evolves over time while samples within each segment remain temporally correlated.
As in the continual OOD setting, we only include online methods. 
In \cref{fig:continual_cvs}, \ours~maintains strong performance after the shift transition.

\subsection{Ablation Studies}

We ablate \ours's two core components, multi-layer fusion and flip correction, both of which prove critical to robust detection. We also provide ablations in Appendix on \ours's remaining design choices, such as the initialization score and the EMA coefficient $\alpha$, where performance remains robust to these variations.



\subsubsection{Impact of Multi-layer Fusion.} 
\begin{figure}[t]
    \centering
    \begin{minipage}{0.43\linewidth}
        \centering
        \caption*{Defocus Blur}
        \vspace{-1em}
        \includegraphics[width=\linewidth]{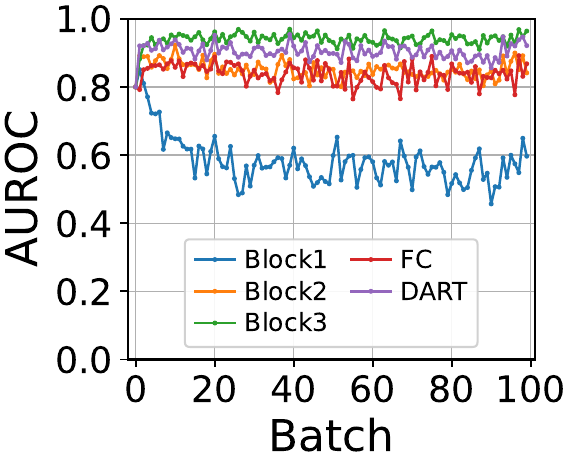}
    \end{minipage}
    \begin{minipage}{0.39\linewidth}
        \centering
        \caption*{Gaussian Noise}
        \vspace{-1em}
        \includegraphics[width=\linewidth]{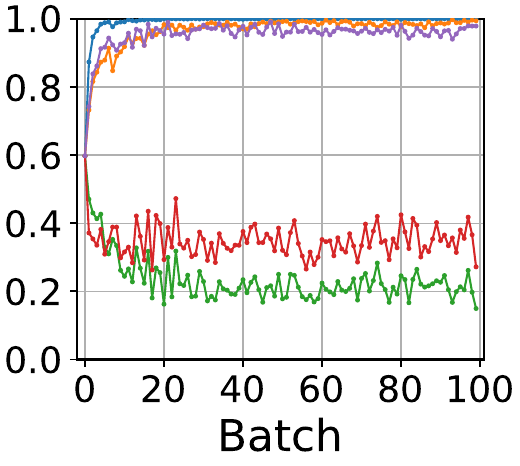}
    \end{minipage}
    \vspace{-0.7em}
    \caption{Impact of Multi-layer Feature Utilization}
    \label{fig:impact_multi_layer} 
    \vspace{-0.5em}
\end{figure}

\Cref{fig:impact_multi_layer} compares  full \ours~against its single-layer variants on CIFAR-100 vs. LSUN, demonstrating the impact of multi-layer fusion. 
While \ours~achieves the highest average performance, the key advantage lies in \textit{stability}. 
Single-layer variants occasionally outperform \ours~in specific settings (\eg, Block3 under Gaussian noise) but exhibit catastrophic failures under other corruptions due to varying covariate-shift impacts across layers. 
Therefore, \ours's multi-layer ensemble provides robustness under diverse covariate shifts.
We provide quantitative results in Appendix.

\subsubsection{Impact of Flip Correction.} 
\Cref{fig:effect_flip_correction} demonstrates the critical role of \ours's flip correction. 
We compare \ours~with \ours-NoFlip (without flip correction) on CIFAR-100 vs. iSUN for original and corrupted datasets. 
While both perform identically on original data, shot noise reveals a drastic difference. 
Standard \ours~recovers via prototype flip and performs robustly, while \ours-NoFlip suffers catastrophic degradation of performance due to reversed prototypes under shot noise. 
The flip correction detects and rectifies these inversions by ensuring consistent prototype proximity. 
Similar effects with JPEG compression further validate its effectiveness. 
We quantify the performance gain from flip correction and report the results in Appendix.

\begin{figure}[t]
    \centering
    \begin{minipage}{0.34\linewidth}
        \centering
        \caption*{Original}
        \vspace{-1em}
        \includegraphics[width=\linewidth]{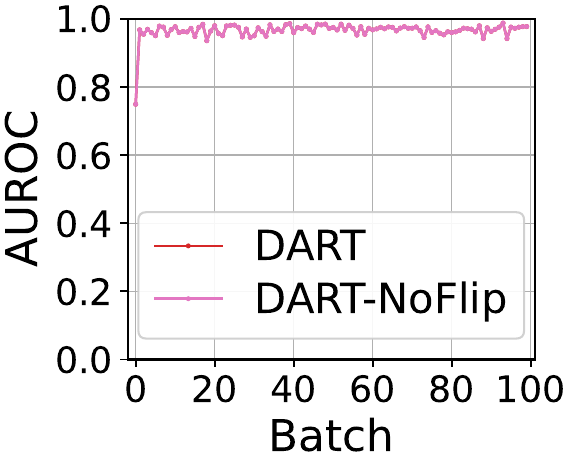}
    \end{minipage}
    \hfill
    \begin{minipage}{0.31\linewidth}
        \centering
        \caption*{Shot noise}
        \vspace{-1em}
        \includegraphics[width=\linewidth]{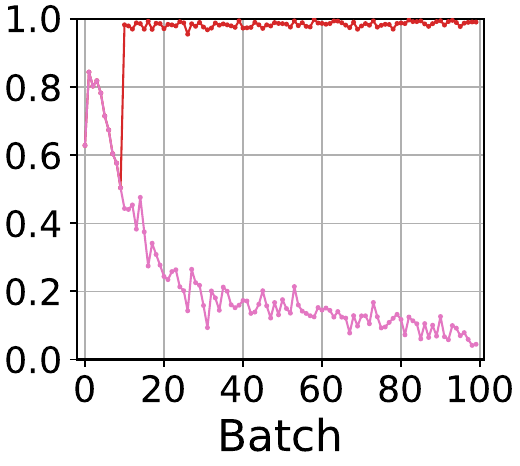}
    \end{minipage}
    \hfill
    \begin{minipage}{0.31\linewidth}
        \centering
        \caption*{JPEG comp.}
        \vspace{-1em}
        \includegraphics[width=\linewidth]{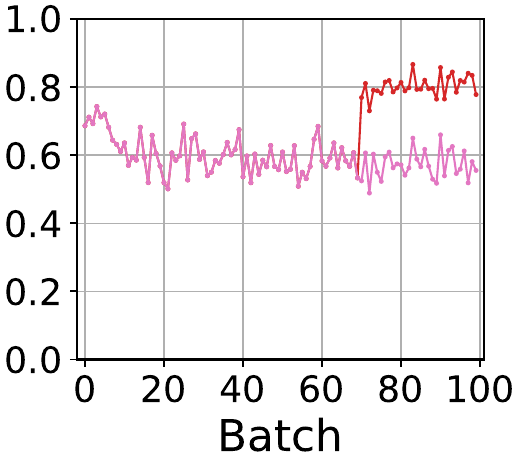}
    \end{minipage}
    \vspace{-0.7em}
    \caption{Impact of Flip Correction.}
    \label{fig:effect_flip_correction}
    \vspace{-1em}
\end{figure}
\section{Conclusion}
\label{sec:conclusion}

In this work, we addressed the realistic challenge of OOD detection under test-time covariate shift, a scenario where existing methods often collapse. 
Our analysis revealed the consistent existence of a discriminative axis along which covariate-shifted ID and OOD samples remain separable. Building on this insight, we proposed \ours, which dynamically tracks prototypes to recover the evolving discriminative axis with multi-layer fusion. Extensive experiments across diverse datasets and architectures confirmed its superiority over strong baselines, underscoring the promise of prototype-based axis tracking as a practical solution for reliable OOD detection in real-world environments.

\noindent\textbf{Limitations and Future Works.}  
While \ours~demonstrates strong performance, several avenues remain for improvement. 
First, since \ours~recovers the discriminative axis from each batch's ID/OOD composition, severely imbalanced batches can destabilize prototype estimation. 
Yet once the axis is stabilized early via a simple initialization strategy detailed in the Appendix, tracking remains reliable even at an OOD ratio as low as 0.01, suggesting that stable initialization is important under extreme imbalance; more principled stabilization schemes are a promising direction.
Additionally, extending \ours~beyond vision tasks to other modalities presents an opportunity to validate the universality of discriminative axis tracking across different data representations.

\section*{Acknowledgments}

This work was supported in part by the Institute of Information \& communications Technology Planning \& Evaluation (IITP) grant funded by the Korea government (MSIT) (No. RS-2026-25507282), and in part by the National Research Foundation (NRF) of Korea grant funded by the Korea government (MSIT) (No. RS-2023-00222663).

{
    \small
    \bibliographystyle{ieeenat_fullname}
    \bibliography{main}

@String(IJCV  = {Int. J. Comput. Vis.})

@String(NeurIPS = {Adv. Neural Inform. Process. Syst.})

@String(IJCV  = {IJCV})

@String(NeurIPS = {NeurIPS})

@article{alam2020survey,
  title={Survey on deep neural networks in speech and vision systems},
  author={Alam, Mahbubul and Samad, Manar D and Vidyaratne, Lasitha and Glandon, Alexander and Iftekharuddin, Khan M},
  journal={Neurocomputing},
  volume={417},
  pages={302--321},
  year={2020},
  publisher={Elsevier}
}

@inproceedings{kim2026imagenet,
  title={ImageNet-sES: A First Systematic Study of Sensor-Environment Simulation Anchored by Real Recaptures},
  author={Kim, Ji-yoon and Baek, Eunsu and Kim, Hyung-Sin},
  booktitle={Proceedings of the IEEE/CVF Winter Conference on Applications of Computer Vision},
  pages={1117--1126},
  year={2026}
}

@article{krizhevsky2012imagenet,
  title={Imagenet classification with deep convolutional neural networks},
  author={Krizhevsky, Alex and Sutskever, Ilya and Hinton, Geoffrey E},
  journal={Advances in neural information processing systems},
  volume={25},
  year={2012}
}

@article{zhang2023openood,
  title={Openood v1. 5: Enhanced benchmark for out-of-distribution detection},
  author={Zhang, Jingyang and Yang, Jingkang and Wang, Pengyun and Wang, Haoqi and Lin, Yueqian and Zhang, Haoran and Sun, Yiyou and Du, Xuefeng and Li, Yixuan and Liu, Ziwei and others},
  journal={arXiv preprint arXiv:2306.09301},
  year={2023}
}

@article{yang2022openood,
  title={Openood: Benchmarking generalized out-of-distribution detection},
  author={Yang, Jingkang and Wang, Pengyun and Zou, Dejian and Zhou, Zitang and Ding, Kunyuan and Peng, Wenxuan and Wang, Haoqi and Chen, Guangyao and Li, Bo and Sun, Yiyou and others},
  journal={Advances in Neural Information Processing Systems},
  volume={35},
  pages={32598--32611},
  year={2022}
}

@inproceedings{baekadaptive,
  title={Adaptive Camera Sensor for Vision Models},
  author={Baek, Eunsu and Han, Sung-hwan and Gong, Taesik and Kim, Hyung-Sin},
  booktitle={The Thirteenth International Conference on Learning Representations},
  year={2025}
}

@inproceedings{baek2024unexplored,
  title={Unexplored faces of robustness and out-of-distribution: Covariate shifts in environment and sensor domains},
  author={Baek, Eunsu and Park, Keondo and Kim, Jiyoon and Kim, Hyung-Sin},
  booktitle={Proceedings of the IEEE/CVF Conference on Computer Vision and Pattern Recognition},
  pages={22294--22303},
  year={2024}
}

@inproceedings{bitterwolf2023or,
  title={In or Out? Fixing ImageNet Out-of-Distribution Detection Evaluation},
  author={Bitterwolf, Julian and M{\"u}ller, Maximilian and Hein, Matthias},
  booktitle={International Conference on Machine Learning},
  pages={2471--2506},
  year={2023},
  organization={PMLR}
}

@article{dockes2021preventing,
  title={Preventing dataset shift from breaking machine-learning biomarkers},
  author={Dock{\`e}s, J{\'e}r{\^o}me and Varoquaux, Ga{\"e}l and Poline, Jean-Baptiste},
  journal={GigaScience},
  volume={10},
  number={9},
  pages={giab055},
  year={2021},
  publisher={Oxford University Press}
}

@article{moreno2012unifying,
  title={A unifying view on dataset shift in classification},
  author={Moreno-Torres, Jose G and Raeder, Troy and Alaiz-Rodr{\'\i}guez, Roc{\'\i}o and Chawla, Nitesh V and Herrera, Francisco},
  journal={Pattern recognition},
  volume={45},
  number={1},
  pages={521--530},
  year={2012},
  publisher={Elsevier}
}

@article{yang2024generalized,
  title={Generalized out-of-distribution detection: A survey},
  author={Yang, Jingkang and Zhou, Kaiyang and Li, Yixuan and Liu, Ziwei},
  journal={International Journal of Computer Vision},
  volume={132},
  number={12},
  pages={5635--5662},
  year={2024},
  publisher={Springer}
}

@inproceedings{yang2021semantically,
  title={Semantically coherent out-of-distribution detection},
  author={Yang, Jingkang and Wang, Haoqi and Feng, Litong and Yan, Xiaopeng and Zheng, Huabin and Zhang, Wayne and Liu, Ziwei},
  booktitle={Proceedings of the IEEE/CVF International Conference on Computer Vision},
  pages={8301--8309},
  year={2021}
}

@article{yang2023full,
  title={Full-spectrum out-of-distribution detection},
  author={Yang, Jingkang and Zhou, Kaiyang and Liu, Ziwei},
  journal={International Journal of Computer Vision},
  volume={131},
  number={10},
  pages={2607--2622},
  year={2023},
  publisher={Springer}
}

@article{wang2020tent,
  title={Tent: Fully test-time adaptation by entropy minimization},
  author={Wang, Dequan and Shelhamer, Evan and Liu, Shaoteng and Olshausen, Bruno and Darrell, Trevor},
  journal={arXiv preprint arXiv:2006.10726},
  year={2020}
}

@article{niu2023towards,
  title={Towards stable test-time adaptation in dynamic wild world},
  author={Niu, Shuaicheng and Wu, Jiaxiang and Zhang, Yifan and Wen, Zhiquan and Chen, Yaofo and Zhao, Peilin and Tan, Mingkui},
  journal={arXiv preprint arXiv:2302.12400},
  year={2023}
}

@inproceedings{katz2022training,
  title={Training ood detectors in their natural habitats},
  author={Katz-Samuels, Julian and Nakhleh, Julia B and Nowak, Robert and Li, Yixuan},
  booktitle={International Conference on Machine Learning},
  pages={10848--10865},
  year={2022},
  organization={PMLR}
}

@inproceedings{huang2021mos,
  title={Mos: Towards scaling out-of-distribution detection for large semantic space},
  author={Huang, Rui and Li, Yixuan},
  booktitle={Proceedings of the IEEE/CVF Conference on Computer Vision and Pattern Recognition},
  pages={8710--8719},
  year={2021}
}

@article{hendrycks2018deep,
  title={Deep anomaly detection with outlier exposure},
  author={Hendrycks, Dan and Mazeika, Mantas and Dietterich, Thomas},
  journal={arXiv preprint arXiv:1812.04606},
  year={2018}
}

@article{zhu2023diversified,
  title={Diversified outlier exposure for out-of-distribution detection via informative extrapolation},
  author={Zhu, Jianing and Geng, Yu and Yao, Jiangchao and Liu, Tongliang and Niu, Gang and Sugiyama, Masashi and Han, Bo},
  journal={Advances in Neural Information Processing Systems},
  volume={36},
  pages={22702--22734},
  year={2023}
}

@inproceedings{zhang2023mixture,
  title={Mixture outlier exposure: Towards out-of-distribution detection in fine-grained environments},
  author={Zhang, Jingyang and Inkawhich, Nathan and Linderman, Randolph and Chen, Yiran and Li, Hai},
  booktitle={Proceedings of the IEEE/CVF Winter Conference on Applications of Computer Vision},
  pages={5531--5540},
  year={2023}
}

@inproceedings{bendale2016towards,
  title={Towards open set deep networks},
  author={Bendale, Abhijit and Boult, Terrance E},
  booktitle={Proceedings of the IEEE conference on computer vision and pattern recognition},
  pages={1563--1572},
  year={2016}
}

@article{shu2017doc,
  title={Doc: Deep open classification of text documents},
  author={Shu, Lei and Xu, Hu and Liu, Bing},
  journal={arXiv preprint arXiv:1709.08716},
  year={2017}
}

@article{chen2021adversarial,
  title={Adversarial reciprocal points learning for open set recognition},
  author={Chen, Guangyao and Peng, Peixi and Wang, Xiangqian and Tian, Yonghong},
  journal={IEEE Transactions on Pattern Analysis and Machine Intelligence},
  volume={44},
  number={11},
  pages={8065--8081},
  year={2021},
  publisher={IEEE}
}

@article{hendrycks2016baseline,
  title={A baseline for detecting misclassified and out-of-distribution examples in neural networks},
  author={Hendrycks, Dan and Gimpel, Kevin},
  journal={arXiv preprint arXiv:1610.02136},
  year={2016}
}

@article{hendrycks2019scaling,
  title={Scaling out-of-distribution detection for real-world settings},
  author={Hendrycks, Dan and Basart, Steven and Mazeika, Mantas and Zou, Andy and Kwon, Joe and Mostajabi, Mohammadreza and Steinhardt, Jacob and Song, Dawn},
  journal={arXiv preprint arXiv:1911.11132},
  year={2019}
}

@article{liu2020energy,
  title={Energy-based out-of-distribution detection},
  author={Liu, Weitang and Wang, Xiaoyun and Owens, John and Li, Yixuan},
  journal={Advances in neural information processing systems},
  volume={33},
  pages={21464--21475},
  year={2020}
}

@article{huang2021importance,
  title={On the importance of gradients for detecting distributional shifts in the wild},
  author={Huang, Rui and Geng, Andrew and Li, Yixuan},
  journal={Advances in Neural Information Processing Systems},
  volume={34},
  pages={677--689},
  year={2021}
}

@article{behpour2023gradorth,
  title={Gradorth: A simple yet efficient out-of-distribution detection with orthogonal projection of gradients},
  author={Behpour, Sima and Doan, Thang Long and Li, Xin and He, Wenbin and Gou, Liang and Ren, Liu},
  journal={Advances in Neural Information Processing Systems},
  volume={36},
  pages={38206--38230},
  year={2023}
}

@article{lee2018simple,
  title={A simple unified framework for detecting out-of-distribution samples and adversarial attacks},
  author={Lee, Kimin and Lee, Kibok and Lee, Honglak and Shin, Jinwoo},
  journal={Advances in neural information processing systems},
  volume={31},
  year={2018}
}

@article{ren2021simple,
  title={A simple fix to mahalanobis distance for improving near-ood detection},
  author={Ren, Jie and Fort, Stanislav and Liu, Jeremiah and Roy, Abhijit Guha and Padhy, Shreyas and Lakshminarayanan, Balaji},
  journal={arXiv preprint arXiv:2106.09022},
  year={2021}
}

@article{mueller2025mahalanobis++,
  title={Mahalanobis++: Improving OOD Detection via Feature Normalization},
  author={Mueller, Maximilian and Hein, Matthias},
  journal={arXiv preprint arXiv:2505.18032},
  year={2025}
}

@inproceedings{sun2022out,
  title={Out-of-distribution detection with deep nearest neighbors},
  author={Sun, Yiyou and Ming, Yifei and Zhu, Xiaojin and Li, Yixuan},
  booktitle={International Conference on Machine Learning},
  pages={20827--20840},
  year={2022},
  organization={PMLR}
}

@inproceedings{park2023nearest,
  title={Nearest neighbor guidance for out-of-distribution detection},
  author={Park, Jaewoo and Jung, Yoon Gyo and Teoh, Andrew Beng Jin},
  booktitle={Proceedings of the IEEE/CVF international conference on computer vision},
  pages={1686--1695},
  year={2023}
}

@inproceedings{wang2022vim,
  title={Vim: Out-of-distribution with virtual-logit matching},
  author={Wang, Haoqi and Li, Zhizhong and Feng, Litong and Zhang, Wayne},
  booktitle={Proceedings of the IEEE/CVF conference on computer vision and pattern recognition},
  pages={4921--4930},
  year={2022}
}

@article{djurisic2022extremely,
  title={Extremely simple activation shaping for out-of-distribution detection},
  author={Djurisic, Andrija and Bozanic, Nebojsa and Ashok, Arjun and Liu, Rosanne},
  journal={arXiv preprint arXiv:2209.09858},
  year={2022}
}

@article{xu2023scaling,
  title={Scaling for training time and post-hoc out-of-distribution detection enhancement},
  author={Xu, Kai and Chen, Rongyu and Franchi, Gianni and Yao, Angela},
  journal={arXiv preprint arXiv:2310.00227},
  year={2023}
}

@article{sun2021react,
  title={React: Out-of-distribution detection with rectified activations},
  author={Sun, Yiyou and Guo, Chuan and Li, Yixuan},
  journal={Advances in neural information processing systems},
  volume={34},
  pages={144--157},
  year={2021}
}

@article{fang2024kernel,
  title={Kernel PCA for out-of-distribution detection},
  author={Fang, Kun and Tao, Qinghua and Lv, Kexin and He, Mingzhen and Huang, Xiaolin and Yang, Jie},
  journal={Advances in Neural Information Processing Systems},
  volume={37},
  pages={134317--134344},
  year={2024}
}

@inproceedings{zhang2022out,
  title={Out-of-distribution detection based on in-distribution data patterns memorization with modern hopfield energy},
  author={Zhang, Jinsong and Fu, Qiang and Chen, Xu and Du, Lun and Li, Zelin and Wang, Gang and Han, Shi and Zhang, Dongmei and others},
  booktitle={The Eleventh International Conference on Learning Representations},
  year={2022}
}

@inproceedings{fan2024test,
  title={Test-time linear out-of-distribution detection},
  author={Fan, Ke and Liu, Tong and Qiu, Xingyu and Wang, Yikai and Huai, Lian and Shangguan, Zeyu and Gou, Shuang and Liu, Fengjian and Fu, Yuqian and Fu, Yanwei and others},
  booktitle={Proceedings of the IEEE/CVF Conference on Computer Vision and Pattern Recognition},
  pages={23752--23761},
  year={2024}
}

@inproceedings{yang2025oodd,
  title={OODD: Test-time Out-of-Distribution Detection with Dynamic Dictionary},
  author={Yang, Yifeng and Zhu, Lin and Sun, Zewen and Liu, Hengyu and Gu, Qinying and Ye, Nanyang},
  booktitle={Proceedings of the Computer Vision and Pattern Recognition Conference},
  pages={30630--30639},
  year={2025}
}

@article{yang2023auto,
  title={Auto: Adaptive outlier optimization for online test-time ood detection},
  author={Yang, Puning and Liang, Jian and Cao, Jie and He, Ran},
  journal={arXiv preprint arXiv:2303.12267},
  year={2023}
}

@article{guo2016deep,
  title={Deep learning for visual understanding: A review},
  author={Guo, Yanming and Liu, Yu and Oerlemans, Ard and Lao, Songyang and Wu, Song and Lew, Michael S},
  journal={Neurocomputing},
  volume={187},
  pages={27--48},
  year={2016},
  publisher={Elsevier}
}

@article{yin2019fourier,
  title={A fourier perspective on model robustness in computer vision},
  author={Yin, Dong and Gontijo Lopes, Raphael and Shlens, Jon and Cubuk, Ekin Dogus and Gilmer, Justin},
  journal={Advances in Neural Information Processing Systems},
  volume={32},
  year={2019}
}

@article{otsu1975threshold,
  title={A threshold selection method from gray-level histograms},
  author={Otsu, Nobuyuki and others},
  journal={Automatica},
  volume={11},
  number={285-296},
  pages={23--27},
  year={1975}
}

@book{tukey1977exploratory,
  title={Exploratory data analysis},
  author={Tukey, John Wilder and others},
  volume={2},
  year={1977},
  publisher={Springer}
}

@article{krizhevsky2009learning,
  title={Learning multiple layers of features from tiny images},
  author={Krizhevsky, Alex and Hinton, Geoffrey and others},
  year={2009},
  publisher={Toronto, ON, Canada}
}

@inproceedings{netzer2011reading,
  title={Reading digits in natural images with unsupervised feature learning},
  author={Netzer, Yuval and Wang, Tao and Coates, Adam and Bissacco, Alessandro and Wu, Baolin and Ng, Andrew Y and others},
  booktitle={NIPS workshop on deep learning and unsupervised feature learning},
  volume={2011},
  number={2},
  pages={4},
  year={2011},
  organization={Granada}
}

@article{zhou2017places,
  title={Places: A 10 million image database for scene recognition},
  author={Zhou, Bolei and Lapedriza, Agata and Khosla, Aditya and Oliva, Aude and Torralba, Antonio},
  journal={IEEE transactions on pattern analysis and machine intelligence},
  volume={40},
  number={6},
  pages={1452--1464},
  year={2017},
  publisher={IEEE}
}

@article{yu2015lsun,
  title={Lsun: Construction of a large-scale image dataset using deep learning with humans in the loop},
  author={Yu, Fisher and Seff, Ari and Zhang, Yinda and Song, Shuran and Funkhouser, Thomas and Xiao, Jianxiong},
  journal={arXiv preprint arXiv:1506.03365},
  year={2015}
}

@article{xu2015turkergaze,
  title={Turkergaze: Crowdsourcing saliency with webcam based eye tracking},
  author={Xu, Pingmei and Ehinger, Krista A and Zhang, Yinda and Finkelstein, Adam and Kulkarni, Sanjeev R and Xiao, Jianxiong},
  journal={arXiv preprint arXiv:1504.06755},
  year={2015}
}

@inproceedings{cimpoi2014describing,
  title={Describing textures in the wild},
  author={Cimpoi, Mircea and Maji, Subhransu and Kokkinos, Iasonas and Mohamed, Sammy and Vedaldi, Andrea},
  booktitle={Proceedings of the IEEE conference on computer vision and pattern recognition},
  pages={3606--3613},
  year={2014}
}

@inproceedings{deng2009imagenet,
  title={Imagenet: A large-scale hierarchical image database},
  author={Deng, Jia and Dong, Wei and Socher, Richard and Li, Li-Jia and Li, Kai and Fei-Fei, Li},
  booktitle={2009 IEEE conference on computer vision and pattern recognition},
  pages={248--255},
  year={2009},
  organization={Ieee}
}

@inproceedings{hendrycks2021natural,
  title={Natural adversarial examples},
  author={Hendrycks, Dan and Zhao, Kevin and Basart, Steven and Steinhardt, Jacob and Song, Dawn},
  booktitle={Proceedings of the IEEE/CVF conference on computer vision and pattern recognition},
  pages={15262--15271},
  year={2021}
}

@inproceedings{xiao2010sun,
  title={Sun database: Large-scale scene recognition from abbey to zoo},
  author={Xiao, Jianxiong and Hays, James and Ehinger, Krista A and Oliva, Aude and Torralba, Antonio},
  booktitle={2010 IEEE computer society conference on computer vision and pattern recognition},
  pages={3485--3492},
  year={2010},
  organization={IEEE}
}

@inproceedings{van2018inaturalist,
  title={The inaturalist species classification and detection dataset},
  author={Van Horn, Grant and Mac Aodha, Oisin and Song, Yang and Cui, Yin and Sun, Chen and Shepard, Alex and Adam, Hartwig and Perona, Pietro and Belongie, Serge},
  booktitle={Proceedings of the IEEE conference on computer vision and pattern recognition},
  pages={8769--8778},
  year={2018}
}

@article{hendrycks2019benchmarking,
  title={Benchmarking neural network robustness to common corruptions and perturbations},
  author={Hendrycks, Dan and Dietterich, Thomas},
  journal={arXiv preprint arXiv:1903.12261},
  year={2019}
}

@inproceedings{he2016deep,
  title={Deep residual learning for image recognition},
  author={He, Kaiming and Zhang, Xiangyu and Ren, Shaoqing and Sun, Jian},
  booktitle={Proceedings of the IEEE conference on computer vision and pattern recognition},
  pages={770--778},
  year={2016}
}

@article{zagoruyko2016wide,
  title={Wide residual networks},
  author={Zagoruyko, Sergey and Komodakis, Nikos},
  journal={arXiv preprint arXiv:1605.07146},
  year={2016}
}

@inproceedings{radosavovic2020designing,
  title={Designing network design spaces},
  author={Radosavovic, Ilija and Kosaraju, Raj Prateek and Girshick, Ross and He, Kaiming and Doll{\'a}r, Piotr},
  booktitle={Proceedings of the IEEE/CVF conference on computer vision and pattern recognition},
  pages={10428--10436},
  year={2020}
}

@inproceedings{vit,
  author    = {Alexey Dosovitskiy and Lucas Beyer and Alexander Kolesnikov and Dirk Weissenborn and Xiaohua Zhai and Thomas Unterthiner and Mostafa Dehghani and Matthias Minderer and G. Heigold and S. Gelly and Jakob Uszkoreit and N. Houlsby},
  year      = {2021},
  title     = {An Image is Worth 16x16 Words: Transformers for Image Recognition at Scale},
  booktitle = {International Conference on Learning Representations},
}

@misc{vit-tiny,
  author = {Winkawaks},
  title = {vit-tiny-patch16-224},
  year = {2023},
  publisher = {HuggingFace},
  journal = {HuggingFace  Model Hub},
  howpublished = {\url{https://huggingface.co/WinKawaks/vit-tiny-patch16-224}}
}

@inproceedings{liu2021swin,
  title={Swin transformer: Hierarchical vision transformer using shifted windows},
  author={Liu, Ze and Lin, Yutong and Cao, Yue and Hu, Han and Wei, Yixuan and Zhang, Zheng and Lin, Stephen and Guo, Baining},
  booktitle={Proceedings of the IEEE/CVF international conference on computer vision},
  pages={10012--10022},
  year={2021}
}

@article{vaswani2017attention,
  title={Attention is all you need},
  author={Vaswani, Ashish and Shazeer, Noam and Parmar, Niki and Uszkoreit, Jakob and Jones, Llion and Gomez, Aidan N and Kaiser, {\L}ukasz and Polosukhin, Illia},
  journal={Advances in neural information processing systems},
  volume={30},
  year={2017}
}

@article{hendrycks2019augmix,
  title={Augmix: A simple data processing method to improve robustness and uncertainty},
  author={Hendrycks, Dan and Mu, Norman and Cubuk, Ekin D and Zoph, Barret and Gilmer, Justin and Lakshminarayanan, Balaji},
  journal={arXiv preprint arXiv:1912.02781},
  year={2019}
}

@article{croce2020robustbench,
  title={Robustbench: a standardized adversarial robustness benchmark},
  author={Croce, Francesco and Andriushchenko, Maksym and Sehwag, Vikash and Debenedetti, Edoardo and Flammarion, Nicolas and Chiang, Mung and Mittal, Prateek and Hein, Matthias},
  journal={arXiv preprint arXiv:2010.09670},
  year={2020}
}

@inproceedings{
liang2018enhancing,
title={Enhancing The Reliability of Out-of-distribution Image Detection in Neural Networks},
author={Shiyu Liang and Yixuan Li and R. Srikant},
booktitle={International Conference on Learning Representations},
year={2018},
url={https://openreview.net/forum?id=H1VGkIxRZ},
}

@article{paszke2019pytorch,
  title={Pytorch: An imperative style, high-performance deep learning library},
  author={Paszke, Adam and Gross, Sam and Massa, Francisco and Lerer, Adam and Bradbury, James and Chanan, Gregory and Killeen, Trevor and Lin, Zeming and Gimelshein, Natalia and Antiga, Luca and others},
  journal={Advances in neural information processing systems},
  volume={32},
  year={2019}
}

@inproceedings{ming2022mcm,
  title={Delving into Out-of-Distribution Detection with Vision-Language Representations},
  author={Ming, Yifei and Cai, Ziyang and Gu, Jiuxiang and Sun, Yiyou and Li, Wei and Li, Yixuan},
  booktitle={Advances in Neural Information Processing Systems (NeurIPS)},
  year={2022}
}

@inproceedings{miyai2023locoop,
  title={LoCoOp: Few-Shot Out-of-Distribution Detection via Prompt Learning},
  author={Miyai, Atsuyuki and Yu, Qing and Irie, Go and Aizawa, Kiyoharu},
  booktitle={Advances in Neural Information Processing Systems (NeurIPS)},
  year={2023}
}

@article{miyai2025glmcm,
  title={GL-MCM: Global and Local Maximum Concept Matching for Zero-Shot Out-of-Distribution Detection},
  author={Miyai, Atsuyuki and Yu, Qing and Irie, Go and Aizawa, Kiyoharu},
  journal={International Journal of Computer Vision (IJCV)},
  year={2025}
}

@inproceedings{zhang2024adaneg,
  title={AdaNeg: Adaptive Negative Proxy Guided OOD Detection with Vision-Language Models},
  author={Zhang, Yabin and Zhang, Lei},
  booktitle={Advances in Neural Information Processing Systems (NeurIPS)},
  year={2024}
}

@ARTICLE{61115,
  author={Lin, J.},
  journal={IEEE Transactions on Information Theory}, 
  title={Divergence measures based on the Shannon entropy}, 
  year={1991},
  volume={37},
  number={1},
  pages={145-151},
  doi={10.1109/18.61115}}
}

\newpage

\appendix
\section*{\Huge Appendix}

\section{Experimental details}
 
\subsection{Datasets details}

\subsubsection{ID: CIFAR-100}

CIFAR-100~\cite{krizhevsky2009learning} consists of 60,000 color images of size $32 \times 32$ across 100 object classes, with 600 images per class. The dataset is divided into 50,000 training and 10,000 test samples. It includes diverse categories such as animals, vehicles, and everyday objects, and is commonly used for evaluating fine-grained image classification and representation learning. In our experiments, we use CIFAR-100 as the in-distribution dataset.

\subsubsection{ID: ImageNet}

ImageNet-1K dataset~\cite{deng2009imagenet} contains 1.28M training images and 50K validation images across 1,000 object categories.

\subsubsection{OOD: SVHN}

The Street View House Numbers (SVHN) dataset~\cite{netzer2011reading} contains real-world digit images collected from Google Street View. It consists of over 600,000 images, each containing a single digit cropped from house number signs, with a resolution of $32 \times 32$. The dataset includes 10 classes (digits 0–9) and is known for its relatively low intra-class variability and high image quality. We use SVHN as an out-of-distribution dataset in our evaluation.

\subsubsection{OOD: Places365}

Places365~\cite{zhou2017places} is the standard subset of the Places database, comprising 365 scene categories such as kitchen, mountain, and library, with approximately 1.8 million training and 36,500 validation images. 
In OOD detection benchmarks, a 10,000-image subset drawn from the Places365 test split is commonly used by resizing images to $32 \times 32$ resolution to match CIFAR-style inputs. 
In our experiments, we use the resized Places365 images as out-of-distribution samples.

\subsubsection{OOD: LSUN}

The Large-scale Scene UNderstanding (LSUN) dataset~\cite{yu2015lsun} contains millions of high-resolution images across various indoor and outdoor scene categories such as classroom, church, and bridge. In OOD detection benchmarks, a subset of LSUN is often used by resizing images to $32 \times 32$ resolution to match CIFAR-style inputs. In our experiments, we use the resized LSUN images as out-of-distribution samples.

\subsubsection{OOD: iSUN}

The iSUN dataset~\cite{xu2015turkergaze} consists of natural scene images collected for saliency prediction, containing various indoor and outdoor environments. It includes around 6,000 images, which are typically resized to $32 \times 32$ for compatibility with CIFAR-based architectures. Due to its scene-centric content, iSUN is commonly used as an out-of-distribution dataset in image classification tasks. We follow prior works and use the resized version of iSUN for OOD evaluation.

\subsubsection{OOD: Textures}

The Textures dataset~\cite{cimpoi2014describing}, also known as the Describable Textures Dataset (DTD), contains 5,640 texture images spanning 47 categories such as striped, dotted, and cracked. The images are collected "in the wild" and exhibit a wide range of fine-grained, low-level patterns. Its low semantic content and high texture diversity make it a challenging out-of-distribution benchmark.

\subsubsection{OOD: ImageNet-O}

ImageNet-O~\cite{hendrycks2021natural} is a curated out-of-distribution dataset containing 2,000 natural images that are semantically distinct from the 1,000 classes in ImageNet-1k. The images were collected to naturally lie outside the ImageNet taxonomy while maintaining comparable visual complexity. This dataset serves as a challenging benchmark for evaluating semantic OOD detection.

\subsubsection{OOD: SUN}

The SUN dataset~\cite{xiao2010sun} is a large-scale scene understanding benchmark containing over 130,000 images across a wide variety of indoor and outdoor environments. It covers hundreds of semantic scene categories such as kitchen, mountain, and library. The diversity and scene-centric nature of SUN make it a strong candidate for out-of-distribution evaluation.

\subsubsection{OOD: iNaturalist}

The iNaturalist dataset~\cite{van2018inaturalist} contains high-resolution images of fine-grained natural categories such as plants, insects, birds, and mammals, collected from citizen science platforms. Due to its distinct domain and taxonomic diversity, iNaturalist is widely used as an out-of-distribution benchmark in vision tasks. Its semantic gap from object-centric datasets makes it a challenging OOD evaluation setting.

\subsubsection{OOD: Places}

The Places dataset~\cite{zhou2017places} is a large-scale scene-centric database containing approximately 10 million images across 434 scene categories such as bedroom, forest path, and amphitheater. 
For out-of-distribution evaluation with ImageNet-1k as the in-distribution dataset, we follow the curated benchmark of Huang and Li~\cite{huang2021mos}, in which 10,000 images are sampled from 50 scene categories. 
Its scene-centric content provides a strong semantic contrast to object-centric in-distribution data, making it a widely adopted out-of-distribution benchmark.

\subsubsection{Common corruptions}

To evaluate robustness under covariate shift, we use a set of common image corruptions introduced by Hendrycks and Dietterich~\cite{hendrycks2019benchmarking}. This benchmark includes 15 corruption types, grouped into noise (e.g., Gaussian noise, shot noise), blur (e.g., defocus, motion blur), weather (e.g., snow, fog), and digital distortions (e.g., JPEG compression, pixelation). We apply these corruptions to both in-distribution and out-of-distribution test samples to simulate realistic distribution shifts. Each corruption is applied at severity level 5, following the standard protocol used in prior works.

\subsection{Baselines details}

\subsubsection{MSP} Maximum Softmax Probability \cite{hendrycks2016baseline} uses the highest softmax output value as the confidence score, assuming in-distribution samples yield higher confidence. We extract this directly from the classifier's final layer.

\subsubsection{Energy} Energy-based detection \cite{liu2020energy} computes $E(x) = -\log \sum_{i} \exp(f_i(x))$ from network logits, with lower values indicating in-distribution samples. $T=1.0$ is used for temperature scaling.

\subsubsection{Max logit} This method \cite{hendrycks2019scaling} uses the maximum pre-softmax logit value as the score, avoiding the normalization effect of softmax that may mask useful signals in relative logit magnitudes.


\subsubsection{GradNorm} GradNorm \cite{huang2021importance} measures the gradient magnitude of the loss with respect to the penultimate layer features. OOD samples tend to produce larger gradient norms. We use a temperature of 1.0 for all experiments.

\subsubsection{ViM} Virtual logit Matching \cite{wang2022vim} projects features into a null space and creates a virtual logit to enhance separation between ID and OOD samples. We set the dimension of the null space to 1000 for feature dimensions $\geq$ 1500, to 512 for feature dimensions $\geq$ 768, and to half the size of the feature dimensions otherwise.

\subsubsection{KNN} K-nearest neighbors \cite{sun2022out} measures the distance to k-nearest neighbors in feature space, with OOD samples typically farther from ID samples. We use L2 normalization for features and set k=50 for CIFAR-based experiments and k=200 for ImageNet-based experiments.

\subsubsection{Mahalanobis distance} We implement both single-layer and ensemble versions of this method \cite{lee2018simple}. The single-layer version models class-conditional feature distributions using Gaussian distributions and measures the distance to the nearest class distribution, using the penultimate layer features. The ensemble version combines layer-wise scores from multiple network layers using pre-computed weights. These weights are learned by utilizing FGSM-perturbed inputs (magnitude 0.001) as synthetic OOD data and applying logistic regression (regularization strength C=1.0, max iterations=1000) to determine the contribution of each layer's feature. We extract features for each layer or block depending on the model architecture.

\subsubsection{ODIN} ODIN \cite{liang2018enhancing} enhances OOD detection by applying input perturbations with temperature scaling to create a larger gap between ID and OOD confidence scores. FGSM epsilon values are set as 0.002 for both CIFAR-100 and ImageNet.

\subsubsection{ReAct} ReAct \cite{sun2021react} truncates abnormally high hidden activations at test time, reducing model overconfidence on OOD data while preserving ID performance, thereby improving the separability between ID and OOD sample.

\subsubsection{SCALE} SCALE \cite{xu2023scaling} is a post-hoc OOD detection method that applies activation scaling to penultimate features, thereby enlarging the separation between ID and OOD energy scores.

\subsubsection{ASH} ASH \cite{djurisic2022extremely} prunes a large portion of late-layer activations (e.g., by top-K percentile) and either leaves the remaining values(ASH-P), binarizes them(ASH-B), or rescales them(ASH-S), then propagates the simplified representation through the network for scoring. We use ASH-P for performance comparison.

\subsubsection{RTL} RTL \cite{fan2024test} fits a linear regression between OOD scores and network features at test time, calibrating base detector outputs to improve detection performance through test-time adaptation.

\subsubsection{NNGuide} NNGuide \cite{park2023nearest} leverages k-nearest neighbor distances in the feature space, scaled by the model's confidence scores, to guide OOD detection by measuring how similar a test sample is to training samples while accounting for prediction confidence.

\subsubsection{CoRP} CoRP \cite{fang2024kernel} applies cosine normalization followed by Random Fourier Features approximation of a Gaussian kernel, then computes PCA reconstruction errors for OOD detection.

\subsubsection{MDS++} MDS++ \cite{mueller2025mahalanobis++} enhances the standard Mahalanobis Distance Score by applying L2 normalization to feature representations before computing class-conditional statistics, thereby improving the geometric separation between ID and
OOD samples in the normalized feature space.

\subsubsection{RMDS} RMDS (Relative Mahalanobis Distance Score) \cite{ren2021simple} computes relative Mahalanobis distances by comparing class-conditional scores against global background scores, effectively measuring how much a sample deviates from both
class-specific and overall data distributions.

\subsubsection{RMDS++} RMDS++ \cite{mueller2025mahalanobis++} extends RMDS by incorporating L2 feature normalization before computing relative Mahalanobis distances, combining the benefits of normalized feature spaces with relative distance measurements to achieve more robust OOD detection.

\subsubsection{OODD} OODD \cite{yang2025oodd} performs test-time OOD detection by maintaining an online dynamic dictionary of representative ID and OOD features. The OOD dictionary is continuously updated with incoming test samples, while the ID dictionary is fixed from the training set.

\subsection{Evaluation model details}

For CIFAR-100-based benchmark, we use the pre-trained WideResNet~\cite{zagoruyko2016wide} with 40 layers and widen factor of 2 pretrained with AugMix~\cite{hendrycks2019augmix} on clean CIFAR-100. The pretrained weights for this model is available from RobustBench~\cite{croce2020robustbench}. 

For ImageNet-based benchmark, we use the pre-trained RegNetY-16GF~\cite{radosavovic2020designing} and ResNet-50~\cite{he2016deep} with the PyTorch checkpoint \cite{paszke2019pytorch}, which are trained on ImageNet and widely used for the OOD detection task.

For evaluation on transformer-based architectures, we train two models: ViT-Tiny and Swin-Tiny. Both models are initialized with ImageNet-pretrained weights provided by HuggingFace model hub. We then fine-tune the model weights and classifier on the CIFAR-100 dataset. Training continues until each model reaches its target accuracy threshold (80\% for ViT-Tiny and 85\% for Swin-Tiny), after which early stopping is applied.



\subsection{Evaluation details}

For evaluation, we construct each test-time batch to contain 100 in-distribution (ID) samples and 100 out-of-distribution (OOD) samples, resulting in a fixed batch size of 200. We sample a total of 100 such test batches for each experimental setting. For each batch, we compute the AUROC and FPR@95TPR metrics, and report the final performance by averaging the values across all batches.

\subsection{Compute resources}

All experiments were conducted using NVIDIA RTX 3090 and RTX 4090 GPUs.

\newpage

\section{Method details}


\subsection{Algorithm of \ours} \label{appendix: algorithm}

\begin{algorithm}
\caption{\ours}
\label{alg:dart}
\resizebox{0.94\linewidth}{!}{%
\begin{minipage}{\linewidth}
\begin{algorithmic}[1]
\REQUIRE Pre-trained model $f$, layers $L = \{1, \ldots, L\}$, EMA coefficient $\alpha$

\vspace{1ex}

\STATE \textbf{Initialization using the first batch $\mathcal{B}_1$:}
    \STATE Let $o = (o_1, \dots, o_C)$ be the logit vector
    \STATE Compute MSP: $\text{MSP}_{1,i} = \max_c \frac{\exp(o_c)}{\sum_{j=1}^C \exp(o_j)}$
    \STATE Apply Otsu threshold $\tau_1$ on MSP scores for pseudo-labeling:
    
        $\;\hat{y}_i = \begin{cases}
        \text{ID} & \text{if } \text{MSP}_{1,i} \geq \tau_1 \\
        \text{OOD} & \text{otherwise}
        \end{cases}$

    \STATE Partition features into $\mathcal{B}_{1}^{ID}$ and $\mathcal{B}_{1}^{OOD}$ based on $\hat{y}_i$
    
\FOR{each layer $l \in L$}
    \STATE Initialize prototypes: 
    

    $
    \bar{\mathbf{p}}_{1}^{(l),\mathrm{ID}} 
    = \frac{1}{|\mathcal{B}_{1}^{\mathrm{ID}}|} 
    \sum_{\mathbf{x}_{1,i} \in \mathcal{B}_{1}^{\mathrm{ID}}} f_l(\mathbf{x}_{1,i}),
    \quad
    \bar{\mathbf{p}}_{1}^{(l),\mathrm{OOD}} 
    = \frac{1}{|\mathcal{B}_{1}^{\mathrm{OOD}}|} 
    \sum_{\mathbf{x}_{1,i} \in \mathcal{B}_{1}^{\mathrm{OOD}}} f_l(\mathbf{x}_{1,i})
    $
\ENDFOR


\FOR{\textbf{each batch $\mathcal{B}_t$}}
\FOR{each layer $l \in L$}
    \IF{$t \bmod k = 0$}
        \STATE Apply flip correction if prototypes are misaligned. Refer to Section \ref{para:flip_correction} in main paper for details.
        $k=10$ as default.
    \ENDIF
    \STATE Extract features: $\mathbf{z}_{t,i}^{(l)} = f_l(\mathbf{x}_{t,i}), \; \mathbf{x}_{t,i} \in \mathcal{B}_t$
    
    \STATE Compute RDS: 
    
    $\text{RDS}_i^{(l)} = 1 - \frac{\|\mathbf{z}_i^{(l)} - \bar{\mathbf{p}}_{t-1}^{ID,(l)}\|}{\|\mathbf{z}_i^{(l)} - \bar{\mathbf{p}}_{t-1}^{ID,(l)}\| + \|\mathbf{z}_i^{(l)} - \bar{\mathbf{p}}_{t-1}^{OOD,(l)}\|}$
    
    \STATE Apply Otsu threshold $\tau_t$ on RDS scores for pseudo-labeling:

        $\;\hat{y}_i = \begin{cases}
        \text{ID} & \text{if } \text{RDS}_i^{(l)} \geq \tau_t \\
        \text{OOD} & \text{otherwise}
        \end{cases}$

    \STATE Partition features into $\mathcal{B}_{t}^{ID}$ and $\mathcal{B}_{t}^{OOD}$ based on $\hat{y}_i$
    
    \STATE Apply Tukey's method for outlier filtering: 
    
    Let $s_i = \|z_{t,i}^{(l)} - p^{\mathrm{proto}} \|_2$, 
    where $p^{\mathrm{proto}}$ is the corresponding prototype
    
    Filter out $z_{t,i}^{(l)}$ if $s_i > Q_3 + \gamma \cdot \mathrm{IQR}$
    
    \STATE Compute new centers: 
    
    $
    \hat{\mathbf{p}}_{t}^{(l),\mathrm{ID}} 
    = \frac{1}{|\mathcal{B}_{t}^{\mathrm{ID}}|} 
    \sum_{\mathbf{x}_{t,i} \in \mathcal{B}_{t}^{\mathrm{ID}}} f_l(\mathbf{x}_{t,i}),
    \quad
    \hat{\mathbf{p}}_{t}^{(l),\mathrm{OOD}} 
    = \frac{1}{|\mathcal{B}_{t}^{\mathrm{OOD}}|} 
    \sum_{\mathbf{x}_{t,i} \in \mathcal{B}_{t}^{\mathrm{OOD}}} f_l(\mathbf{x}_{t,i})
    $
    
    \STATE Update prototypes with EMA: 
    
    $\bar{\mathbf{p}}_t^{ID,(l)} = \alpha \cdot \bar{\mathbf{p}}_{t-1}^{ID,(l)} + (1-\alpha) \cdot \hat{\mathbf{p}}_{t}^{ID,(l)},
    \bar{\mathbf{p}}_t^{OOD,(l)} = \alpha \cdot \bar{\mathbf{p}}_{t-1}^{OOD,(l)} + (1-\alpha) \cdot \hat{\mathbf{p}}_{t}^{OOD,(l)}$
\ENDFOR
\STATE $\text{RDS}_{\text{multi}}(x_i) = \frac{1}{L} \sum_{l=1}^{L} \text{RDS}_i^{(l)}$
\ENDFOR


\end{algorithmic}
\end{minipage}
}
\end{algorithm}

\vspace{-1em}

\subsection{Otsu algorithm}

To automatically determine a threshold that separates two distributions (e.g., ID and OOD) based on their scalar scores, we adopt Otsu algorithm~\cite{otsu1975threshold}. Originally proposed for image binarization, Otsu algorithm selects the threshold that minimizes the intra-class variance (or equivalently maximizes the inter-class variance) when partitioning a set of scalar values into two groups.

Given a histogram of score values, the algorithm exhaustively searches for the threshold $\tau$ that minimizes the weighted sum of within-class variances:
\begin{equation}
\sigma^2_{\text{within}}(\tau) = \omega_0(\tau) \sigma_0^2(\tau) + \omega_1(\tau) \sigma_1^2(\tau),
\end{equation}
where $\omega_0(\tau)$ and $\omega_1(\tau)$ are the probabilities of the two classes separated by threshold $\tau$, and $\sigma_0^2(\tau)$, $\sigma_1^2(\tau)$ are the corresponding class variances. This approach allows for an adaptive and data-driven determination of the decision threshold, without requiring access to ground-truth labels or distributional assumptions.

In our method, Otsu algorithm is applied to the distribution of OOD scores computed over each test-time batch. This enables unsupervised, on-the-fly threshold selection for distinguishing ID and OOD samples, and plays a critical role in decision-making process during inference.

\subsection{Tukey's method}

To ensure robust prototype estimation, we apply outlier filtering prior to aggregating the feature representations of test samples. Specifically, we adopt Tukey's method, a non-parametric technique for identifying outliers based on the interquartile range (IQR)~\cite{tukey1977exploratory}.

Given a set of distance values (e.g., Euclidean distances between features and their assigned prototype), we first compute the lower quartile ($Q_1$) and upper quartile ($Q_3$). The interquartile range is then defined as:
\begin{equation}
\text{IQR} = Q_3 - Q_1.
\end{equation}
A sample is identified as a potential outlier if its score $x$ satisfies:
\begin{equation}
\quad x > Q_3 + 1.5 \cdot \text{IQR}.
\end{equation}

We use Tukey’s method with an IQR factor of 1.5 throughout our experiments.

This filtering step is applied independently to the distance scores within each test-time batch, effectively removing extreme values that may otherwise distort the prototype update.

\subsection{Layer selection for \ours}

While it is possible to utilize the output of all intermediate layers for multi-layer aggregation, doing so incurs additional computational overhead. To reduce this overhead while still capturing hierarchical representations, we select a subset of representative layers at coarse block granularity, as specified in Table~\ref{tab:ablation_layers}. 

\begin{table}[h]
    \centering
    \caption{Included layers list for \ours}
    \label{tab:ablation_layers}
    \resizebox{\linewidth}{!}{
    \begin{tabular}{c|c}
        \toprule
        Model architecture & Included layers list \\
        \midrule
        \midrule
        WideResNet-40-2 & \texttt{block1, block2, block3, fc} \\
        \midrule
        RegNetY-16GF & \makecell[c]{
        \texttt{stem, trunk output \{block1.block1-0, block1.block1-1,} \\ 
        \texttt{block2.block2-0 - block2.block2-3, } \\
        \texttt{block3.block3-0 - block3.block3-10, } \\
        \texttt{block4.block4-0\}, fc}
        } \\
        \midrule
        ViT-Tiny & \makecell[c]{
        \texttt{vit.encoder.layer\{0 - 11\}, classifier} \\ 
        } \\
        \midrule
        Swin-T & \makecell[c]{
        \texttt{swin.encoder.layers\{0.blocks.0, 0.blocks.1, } \\ \texttt{1.blocks.0, 1.blocks.1} \\ 
        \texttt{2.blocks.0 - 2.blocks.5, } \\ 
        \texttt{3.blocks.0, 3.blocks.1\}, classifier} \\
        } \\
        \midrule
        ResNet-50 & \texttt{layer1, layer2, layer3, layer4, fc} \\
        \bottomrule
    \end{tabular}
    }
\end{table}

\begin{table*}[h]
    \centering
    \caption{Performance comparison between \ours-NoFlip and \ours}
    \label{tab:ablation_flip_performance}
    \resizebox{0.9\linewidth}{!}{
    \begin{tabular}{l|c@{\hspace{2pt}}c|c@{\hspace{2pt}}c|c@{\hspace{2pt}}c|c@{\hspace{2pt}}c|c@{\hspace{2pt}}c||c@{\hspace{2pt}}c}
        \toprule
        \multirow{2}{*}{Method}
        & \multicolumn{2}{c|}{SVHN-C}
        & \multicolumn{2}{c|}{Places365-C}
        & \multicolumn{2}{c|}{LSUN-C}
        & \multicolumn{2}{c|}{iSUN-C}
        & \multicolumn{2}{c||}{Textures-C}
        & \multicolumn{2}{c}{Average} \\
        \cmidrule(lr){2-13}
        & FPR95 \(\downarrow\) & AUROC \(\uparrow\)
        & FPR95 \(\downarrow\) & AUROC \(\uparrow\)
        & FPR95 \(\downarrow\) & AUROC \(\uparrow\)
        & FPR95 \(\downarrow\) & AUROC \(\uparrow\)
        & FPR95 \(\downarrow\) & AUROC \(\uparrow\)
        & FPR95 \(\downarrow\) & AUROC \(\uparrow\) \\
        \midrule
        \ours-NoFlip & 51.14 & 71.84 & 70.14 & 66.08 & 44.17 & 80.15 & 52.30 & 78.58 & 55.37 & 75.01 & 54.62 & 74.33 \\
        \rowcolor{blue!5}
        \ours        & 48.60 & 79.82 & 68.66 & 68.00 & 44.14 & 80.29 & 50.76 & 79.75 & 51.48 & 80.60 & 52.73 & 77.69 \\
        \bottomrule
    \end{tabular}
    }
\end{table*}

\begin{figure*}[t]
    \centering
    \includegraphics[width=0.9\linewidth]{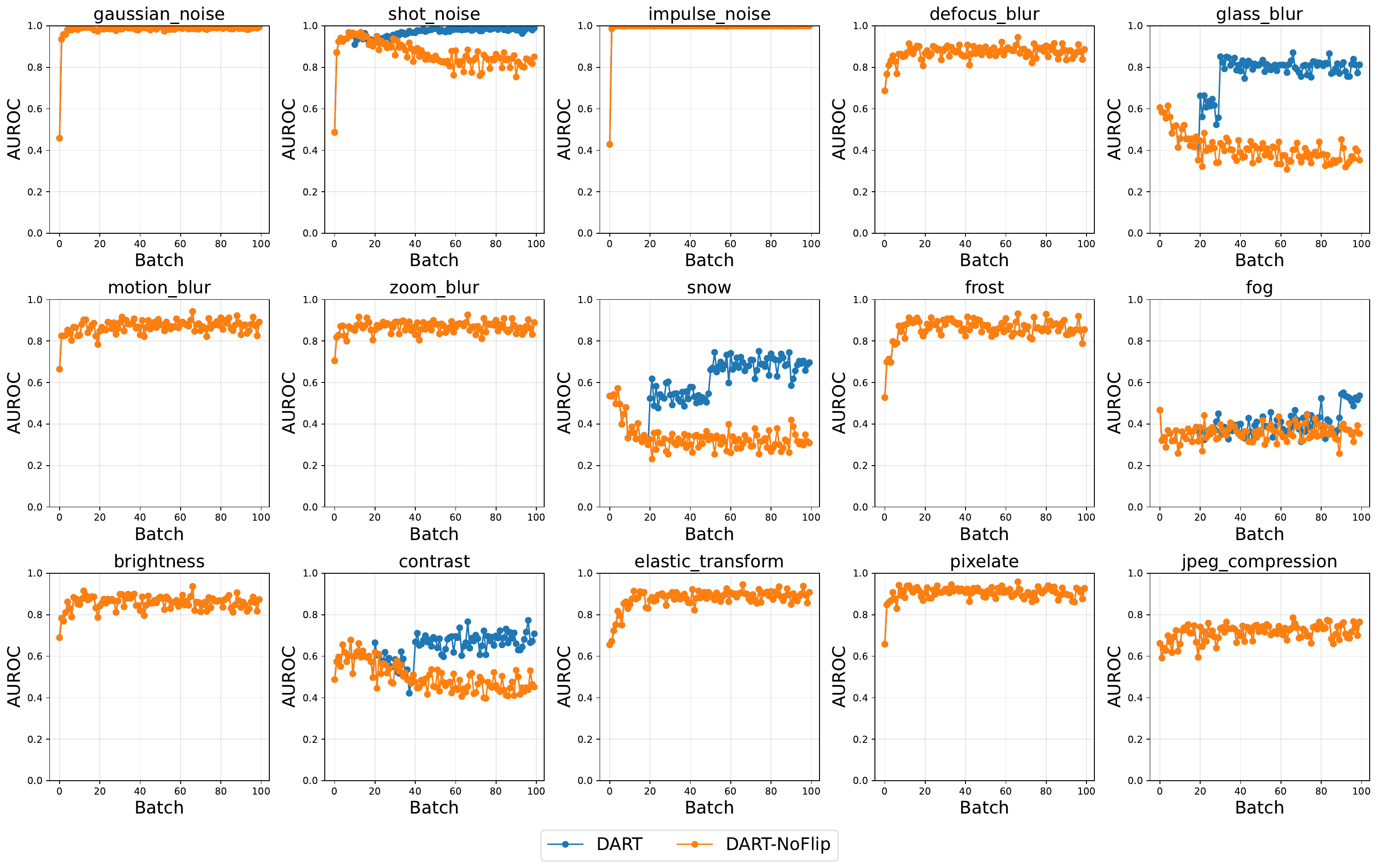}
    \caption{Impact of Flip Correction over time (CIFAR-100-C vs. Textures-C)}
    \label{fig:ablation_flip_viz}
    \vspace{-1em}
\end{figure*}


\section{Theoretical Analysis}
\label{sec:theoretical_analysis}

We analyze why the discriminative axis $\mathbf{d} = \boldsymbol{\mu}_{\mathrm{ID}} - \boldsymbol{\mu}_{\mathrm{OOD}}$ is an effective OOD detector and why its norm is large.
Throughout, $\hat{\mathbf{z}}(x) \in \mathbb{R}^K$ denotes the post-BN feature vector with $K$ channels, and BN normalizes each channel $k$ by frozen ID statistics:
$\hat{Z}_k(x) = \gamma_k \frac{({Z_k(x) - \mu_{\mathrm{ID},k}})}{{\sqrt{\sigma_{\mathrm{ID},k}^2 + \epsilon}}} + \beta_k$.
Since $\epsilon \approx 10^{-5}$, the post-BN ID variance satisfies $v_{\mathrm{ID},k}^2 \triangleq \mathrm{Var}_{x \sim P_{\mathrm{ID}}}[\hat{Z}_k(x)] = \gamma_k^2 \frac{\sigma_{\mathrm{ID},k}^2}{\sigma_{\mathrm{ID},k}^2 + \epsilon} \approx \gamma_k^2$.

\subsection{Separation Bound for the Discriminative Axis}
\label{sec:separation_bound}

Consider the centered projection score $\tilde{s}(x) = \mathbf{d}^\top (\hat{\mathbf{z}}(x) - \boldsymbol{\mu}_{\mathrm{OOD}})$.
The projected class means are
\begin{equation}
    \mathbb{E}[\tilde{s} \mid \mathrm{ID}] = \|\mathbf{d}\|^2,
    \qquad
    \mathbb{E}[\tilde{s} \mid \mathrm{OOD}] = 0,
\end{equation}
giving an exact mean gap of $\|\mathbf{d}\|^2$.
The projected variance is $\mathrm{Var}[\tilde{s} \mid C] = \mathbf{d}^\top \boldsymbol{\Sigma}_{C} \, \mathbf{d}$.
We assume that the within-class variance remains controlled along the discriminative direction:

\paragraph{Assumption A1 (bounded directional variance).} There exists a constant $\kappa > 0$ such that
\begin{equation}
    \mathbf{d}^\top \boldsymbol{\Sigma}_{C} \, \mathbf{d}
    \le \kappa \|\mathbf{d}\|^2,
    \qquad C \in \{\mathrm{ID}, \mathrm{OOD}\}.
    \label{eq:directional_variance_appendix}
\end{equation}
This condition constrains only the variance along the single direction $\mathbf{d}$, without imposing structure on the full covariance matrix.
A sufficient condition is bounded operator norm: $\|\boldsymbol{\Sigma}_C\|_{\mathrm{op}} \le \kappa$ implies A1 immediately.

Using the midpoint threshold $\tau = \|\mathbf{d}\|^2/2$, Chebyshev's inequality yields
\begin{equation}
    P(\tilde{s} < \tau \mid \mathrm{ID})
    \le \frac{4\,\kappa}{\|\mathbf{d}\|^2},
    \label{eq:chebyshev_bound_id_appendix}
\end{equation}
because the ID mean is $\|\mathbf{d}\|^2$, and similarly
\begin{equation}
    P(\tilde{s} \ge \tau \mid \mathrm{OOD})
    \le \frac{4\,\kappa}{\|\mathbf{d}\|^2},
    \label{eq:chebyshev_bound_ood_appendix}
\end{equation}
because the OOD mean is $0$.
Therefore, the overlap decreases as $\mathcal{O}(1/\|\mathbf{d}\|^2)$: larger norm of the discriminative axis directly improves separation between ID and OOD along the projected score.

\paragraph{Fisher optimality of the discriminative axis.}
Let $\mathbf{S}_W = \boldsymbol{\Sigma}_{\mathrm{ID}} + \boldsymbol{\Sigma}_{\mathrm{OOD}}$ be the within-class scatter matrix.
The SNR of the projection along $\mathbf{d}$ equals the Fisher criterion: $\mathrm{SNR} = \|\mathbf{d}\|^4 / (\mathbf{d}^\top \mathbf{S}_W \mathbf{d}) = J(\mathbf{d})$.
In general, the Fisher-optimal direction is $\mathbf{w}^* \propto \mathbf{S}_W^{-1}\mathbf{d}$, which need not align with $\mathbf{d}$.
However, when the within-class scatter is approximately diagonal with comparable entries --- i.e.\ $\mathbf{S}_W \approx 2\gamma^2 \mathbf{I}$ --- the inverse acts as a uniform rescaling and $\mathbf{S}_W^{-1}\mathbf{d} \propto \mathbf{d}$, so the discriminative axis coincides with the Fisher-optimal direction.

\subsection{Why the Discriminative Axis Norm Is Large}
\label{sec:bn_boosts_axis}

The $k$-th component of $\mathbf{d}$ in post-BN space decomposes as
\begin{equation}
    d_k
    = \frac{\gamma_k}{\sqrt{\sigma_{\mathrm{ID},k}^2 + \epsilon}}
      \underbrace{\bigl(\mathbb{E}[Z_k \mid \mathrm{ID}] - \mathbb{E}[Z_k \mid \mathrm{OOD}]\bigr)}_{\Delta_k}
    = \lambda_k \, \Delta_k,
\end{equation}
where $\lambda_k = \gamma_k / \sqrt{\sigma_{\mathrm{ID},k}^2 + \epsilon}$ is the BN scaling factor and $\Delta_k$ is the pre-BN mean shift.
The axis norm is $\|\mathbf{d}\|^2 = \sum_{k=1}^K \lambda_k^2 \, \Delta_k^2$.

Any $\lambda_k > 1$ means BN amplifies the pre-BN mean shift in channel~$k$, and the key observation is that this amplification \emph{compounds across $K$ channels}: even $\bar{\lambda}$ modestly above~$1$ can yield $\|\mathbf{d}\|^2 \gg 4\kappa$ when $K$ is large.

In the final BN layer of a pretrained ResNet-50 ($K{=}2048$), the measured scaling factors have mean $\bar{\lambda} \approx 23.3$ with a coefficient of variation of $5.7\%$, placing the bound deep in the non-trivial regime: moderate per-channel shifts accumulate across $2048$ channels into a large $\|\mathbf{d}\|^2$, directly tightening the bound in \cref{eq:chebyshev_bound_id_appendix}.

\begin{figure*}[t]
    \centering
    \includegraphics[width=\linewidth]{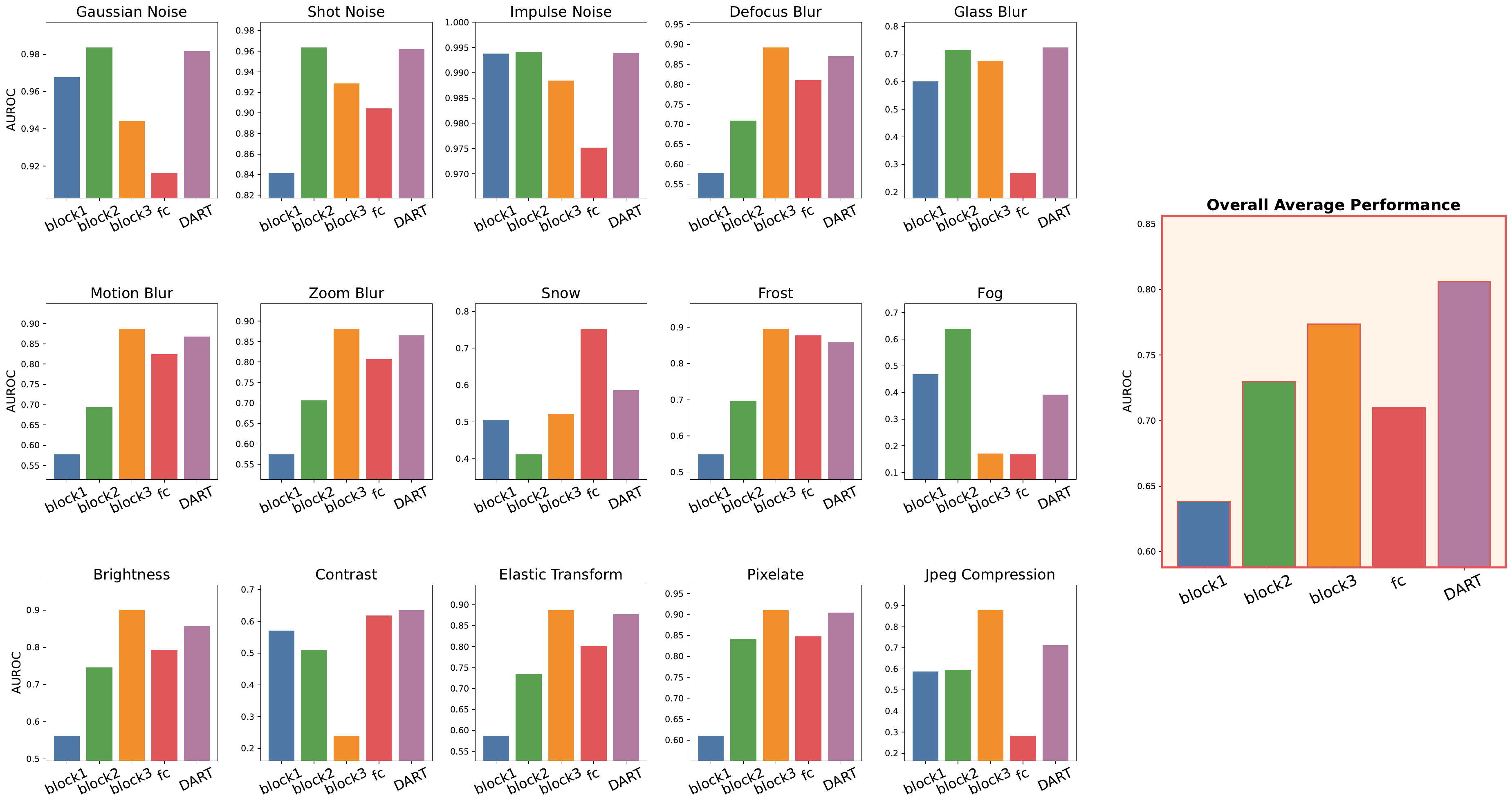}
    \caption{Impact of Multi-layer Fusion (CIFAR-100-C vs. Textures-C)}
    \label{fig:ablation_multi}
    \vspace{-1em}
\end{figure*}

\section{Impact of Flip Correction}

To quantitatively assess the impact of flip correction, we compare with a variant that does not perform flip correction, \ours-NoFlip, using CIFAR-100-C as the csID dataset. As shown in Table~\ref{tab:ablation_flip_performance}, flip correction improves the performance of our method.

To further illustrate the effect on performance over time, we additionally analyze CIFAR-100-C vs Textures-C on a per-corruption basis. Figure~\ref{fig:ablation_flip_viz} visualizes the detection performance over time (per-batch AUROC) before and after applying flip correction. For several corruptions—glass blur, snow, fog, and contrast—we observe that once flip correction is triggered, the dual prototype axis is realigned toward the oracle discriminative direction, leading to an abrupt jump and sustained improvement in performance. For shot noise, flip correction occurs early in the stream at the 20-th batch, after which the subsequent batches exhibit much more stable and higher performance compared to \ours-NoFlip.

\section{Impact of Multi-layer Fusion}

We report the AUROC of each single-layer variant (Block1, Block2, Block3, FC) and compare them against full \ours~with multi-layer fusion for all 15 corruption types, using CIFAR-100-C as csID and Textures-C as csOOD. See Figure~\ref{fig:ablation_multi} for all results.

This extended analysis reveals that the best-performing layer is highly shift-dependent: under noise-type corruptions such as Gaussian or impulse noise, deeper layers suffer larger degradation, whereas under corruptions such as motion blur and brightness, earlier layers are more severely affected and later layers remain relatively more informative. As a result, relying on any single fixed layer for OOD detection is brittle when the covariate shift type is unknown a priori. In contrast, the fused \ours~score achieves both the highest mean performance and the most stable behavior: averaged over all corruption types, \ours~not only outperforms every single-layer variant, but also exhibits substantially smaller variation than the strongest single-layer baseline (Block3), with standard deviation 0.1661 versus 0.2582 for Block3. These results quantitatively support our claim that multi-layer fusion is crucial for robust OOD detection under unpredictable covariate shift.

\section{Effect of Design Choices}

In \cref{tab:heuristics}, we quantify \ours's design sensitivity: MSP/Energy/ODIN/ASH initialization gives covariate-shifted AUROC 88.04/84.39/93.81/88.72 and clean 95.41/94.96/95.51/95.86. Stronger ODIN can improve the first axis estimate, while MSP remains an effective lightweight default. Tukey factors 1.0/1.5/2.0/no-filter give shifted AUROC 88.97/88.04/87.02/86.13, confirming filtering helps under shift. Flip weights 0/0.5/1/2/5 stay within 87.06–88.04 shifted and 95.34–95.99 clean; EMA std is $\le$0.29 pp. Otsu is deterministic, label-free between-class-variance thresholding, not a tuned threshold.

\newlength{\wFive}
\newlength{\wFour}
\setlength{\wFive}{%
  \dimexpr(\linewidth - 10\tabcolsep - 6\arrayrulewidth)/5\relax
}
\setlength{\wFour}{%
  \dimexpr(\linewidth - 8\tabcolsep - 5\arrayrulewidth)/4\relax
}

\newcommand{\FiveFirst}[1]{%
  \multicolumn{4}{>{\centering}p{\wFive}}{#1}%
}
\newcommand{\FiveCell}[1]{%
  \multicolumn{4}{>{\centering}p{\wFive}}{#1}%
}

\newcommand{\FourFirst}[1]{%
  \multicolumn{5}{>{\centering}p{\wFour}}{#1}%
}
\newcommand{\FourCell}[1]{%
  \multicolumn{5}{>{\centering}p{\wFour}}{#1}%
}

\newcommand{\graycell}{\cellcolor{gray!15}}

\begin{table}[h]
\centering
\setlength{\tabcolsep}{3pt}
\captionof{table}{\small AUROC across design choices. (Gray-shaded: default)}
\vspace{-0.5em}
\label{tab:heuristics}
\resizebox{\linewidth}{!}{
\begin{tabular}{l|l| *{21}{c|}}
\toprule
\multirow{9}{*}{\rotatebox{90}{\makecell[c]{\textbf{ResNet-50}. ImageNet as ID. \\ Avg. performance over 5 OODs.}}}
& \multirow{3}{*}{\makecell[c]{\textbf{Init.} \\ \textbf{method}}} & \textbf{init. method} & \FourFirst{\graycell\textbf{MSP}} & \FourCell{\textbf{Energy}} & \FourCell{\textbf{ODIN}} & \FourCell{\textbf{ASH}} \\
\cmidrule(lr){3-23}
& & Cov. shift & \FourFirst{\graycell88.04} & \FourCell{84.39} & \FourCell{93.81} & \FourCell{88.72} \\
& & Clean & \FourFirst{\graycell95.41} & \FourCell{94.96} & \FourCell{95.51} & \FourCell{95.86} \\
\cmidrule(lr){2-23}
& \multirow{3}{*}{\makecell[c]{\textbf{Tukey} \\ \textbf{filtering}}} & \textbf{IQR factor} & \FourFirst{\textbf{1.0}} & \FourCell{\graycell\textbf{1.5}} & \FourCell{\textbf{2.0}} & \FourCell{\textbf{100.0}} \\
\cmidrule(lr){3-23}
& & Cov. shift & \FourFirst{88.97} & \FourCell{\graycell88.04} & \FourCell{87.02} & \FourCell{86.13} \\
& & Clean & \FourFirst{94.65} & \FourCell{\graycell95.41} & \FourCell{94.87} & \FourCell{94.90} \\
\cmidrule(lr){2-23}
& \multirow{3}{*}{\makecell[c]{\textbf{Flip} \\ \textbf{weight}}} & \textbf{weight factor} & \FiveFirst{\textbf{0.0}} & \FiveCell{\textbf{0.5}} & \FiveCell{\textbf{1.0}} & \FiveCell{\graycell\textbf{2.0}} & \FiveCell{\textbf{5.0}} \\
\cmidrule(lr){3-23}
& & Cov. shift & \FiveFirst{87.33} & \FiveCell{87.31} & \FiveCell{87.21} & \FiveCell{\graycell88.04} & \FiveCell{87.06} \\
& & Clean & \FiveFirst{95.48} & \FiveCell{95.48} & \FiveCell{95.34} & \FiveCell{\graycell95.41} & \FiveCell{95.99} \\
\bottomrule
\end{tabular}
}
\vspace{-1em}
\end{table}

\section{Effect of EMA $\alpha$}

We employ an exponential moving average (EMA) to update the prototype, where the smoothing coefficient $\alpha$ is treated as the only hyperparameter of \ours. To examine the sensitivity of our method to this hyperparameter, we conduct an ablation study on the CIFAR-100-C vs. SVHN-C and ImageNet-C vs. iNaturalist-C benchmark, and report the results as AUROC(\%) in Table~\ref{tab:ablation_ema_alpha}. Results in Table~\ref{tab:ablation_ema_alpha} demonstrate that our method is robust to the choice of the EMA coefficient $\alpha$. We use $\alpha=0.8$ as the default.

\begin{table}[h]
    \centering
    \caption{The effect of EMA \(\alpha\) on the performance of \ours}
    \vspace{-0.5em}
    \label{tab:ablation_ema_alpha}
    \setlength{\tabcolsep}{6pt}
    \resizebox{\linewidth}{!}{
    \begin{tabular}{c|ccccccc|c}
        \toprule
        \(\alpha\) & 0.5 & 0.6 & 0.7 & 0.8 & 0.9 & std. \\
        \midrule
        \midrule
        CIFAR-100-C vs. SVHN-C & 78.79 & 78.68 & 78.84 & 78.58 & 78.11 & \(\pm\) 0.29 \\
        \midrule
        ImageNet-C vs. iNaturalist-C & 93.18 & 93.18 & 93.18 & 93.18 & 93.19 & \(\pm\) 0.00 \\
        \bottomrule
    \end{tabular}
    }
    \vspace{-1em}
\end{table}

\section{Sensitivity to the ID/OOD Ratio}

\Cref{tab:ablation_id_ood_ratio} shows that \ours~remains robust across a broad range of batch compositions, achieving consistently strong performance when the ID proportion is between 0.2 and 0.8. 
We use ResNet-50 with ImageNet as ID and iNaturalist as OOD. 
This highlights an important strength of our method: it does not rely on a narrowly tuned ID/OOD mixture and generalizes well under moderate batch imbalance. 
Nevertheless, we observe a performance drop in more extreme cases, such as when the ID proportion reaches 0.9. 
Because \ours~leverages the internal geometry of each batch, a sufficient number of both ID and OOD samples is generally needed to stably initialize and preserve the discriminative axis. 
When OOD samples become overly scarce, the estimated axis can become less reliable and more sensitive.

\begin{table}[h]
    \centering
    \caption{The sensitivity to the ID proportion}
    \vspace{-0.5em}
    \label{tab:ablation_id_ood_ratio}
    \setlength{\tabcolsep}{6pt}
    \resizebox{\linewidth}{!}{
    \begin{tabular}{c|ccccccccc}
        \toprule
        $| \mathcal{B}_t^{\mathrm{I}} | / | \mathcal{B}_t |$ & 0.1 & 0.2 & 0.3 & 0.4 & 0.5 & 0.6 & 0.7 & 0.8 & 0.9 \\
        \midrule
        \midrule
        FPR95 & 32.31 & 20.15 & 16.12 & 15.50 & 14.29 & 12.14 & 12.45 & 23.02 & 68.57 \\
        \midrule 
        AUROC & 84.67 & 94.81 & 96.33 & 96.05 & 96.46 & 97.04 & 96.80 & 93.88 & 70.31 \\
        \bottomrule
    \end{tabular}
    }
\end{table}

Importantly, this issue can be mitigated through an axis stabilization strategy which combines balanced initialization with adaptive and conservative axis updates. 
To verify this, we conduct an additional experiment in \cref{tab:ablation_id_ood_ratio_stable}, where the discriminative axis is initialized using five balanced batches before the stream transitions to the target highly imbalanced ID/OOD ratio, and updated with a large EMA coefficient $\alpha=0.95$. 
Under this setting, the performance degradation is substantially reduced even in highly imbalanced regimes that can occur in practice when OOD samples are particularly scarce, including ID proportions of 0.9, 0.95, and 0.99. 
These results suggest that the stabilized initialization is important and that the performance drop of \ours~under extreme imbalance is largely associated with instability in axis estimation, which can be effectively controlled with proper initialization and conservative updates. 
A promising direction for future work is to adapt the EMA coefficient to the batch composition, rather than using a fixed value, in order to further improve axis stability under severe imbalance.

\begin{table}[h]
    \centering
    \caption{The sensitivity to the ID proportion with axis stabilization strategy}
    \vspace{-0.5em}
    \label{tab:ablation_id_ood_ratio_stable}
    \setlength{\tabcolsep}{6pt}
    \resizebox{0.5\linewidth}{!}{
    \begin{tabular}{c|ccc}
        \toprule
        $\| \mathcal{B}_t^{\mathrm{I}} \| / \| \mathcal{B}_t \|$ & 0.9 & 0.95 & 0.99 \\
        \midrule
        \midrule
        FPR95 & 14.00 & 16.96 & 30.18 \\
        \midrule 
        AUROC & 96.21 & 95.34 & 89.20 \\
        \bottomrule
    \end{tabular}
    }
    \vspace{-1em}
\end{table}

\section{Single Sample Batch}

Batch-size-1 (BS=1) streaming is a practical setting; however, under \ours’s current dual-prototype initialization, a one-sample cold-start is not well-defined because one sample cannot form two pseudo-groups. 
We therefore initialize with a small unlabeled warm-up buffer (10/20/200 samples) and then process a single-sample batch stream. 
The warm-up buffer is class-balanced, containing an equal number of ID and OOD samples, and is used only to initialize the dual prototypes; no labels are used at any point. 
Once the stream begins, each batch carries a single sample, so the batch-wise pseudo-labeling of Sec. 3.3 no longer applies. 
We therefore assign the sample to the nearer of the two prototypes under the Euclidean distance and update only that prototype by EMA, leaving the opposite prototype unchanged for that batch.
Since a single sample admits no batch-level ranking, we cannot compute AUROC per batch as in the main experiments. 
We instead accumulate the scores of all samples over the entire stream and report a single AUROC at the end.

\begin{table}[h]
    \centering
    \caption{OOD detection performance (AUROC) with single sample batches. Small initialization buffer with various numbers of ID/OOD samples are used. BS denotes batch size. ImageNet as ID and RegNetY-16GF are used for evaluation.}
    \label{tab:single_sample}
    \resizebox{0.9\linewidth}{!}{
    \begin{tabular}{c||c|c|c|c|c}
        \toprule
        \makecell[c]{Initialization \\ buffer size}
        & ImageNet-O
        & Places
        & SUN
        & iNaturalist
        & Textures \\
        \midrule
        10  & 92.48 & 51.76 & 64.88 & 91.64 & 81.60 \\
        20  & 92.12 & 68.68 & 84.60 & 93.68 & 90.40 \\
        200 & 98.24 & 76.12 & 86.76 & 93.92 & 92.20 \\
        \midrule
        \rowcolor{gray!15}
        (BS = 200) & 99.82 & 96.31 & 90.86 & 95.28 & 98.36 \\
        \bottomrule
    \end{tabular}
    }
\end{table}

\cref{tab:single_sample} reports results with ImageNet as ID and RegNetY-16GF as the backbone. 
For most OOD datasets, a warm-up buffer of 20 samples or more is sufficient to reach performance close to the batched setting, indicating that the axis, once initialized, can be tracked from single-sample updates alone. 
Nevertheless, the cold start itself remains undefined under the current formulation, and we regard this as a limitation to be addressed by a more principled single-sample initialization.

\begin{figure*}[t]
    \centering
    \begin{minipage}{0.8\linewidth}
        \centering
        \includegraphics[width=\linewidth]{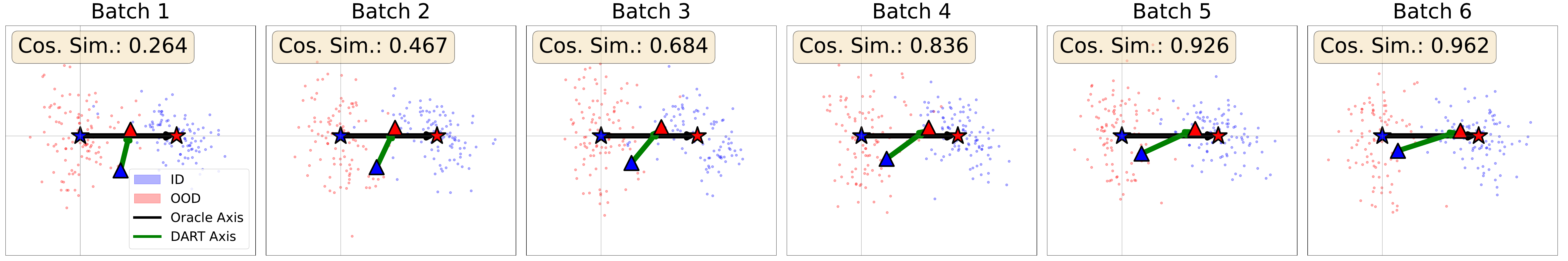}
    \end{minipage}
    \caption{Progression of axis alignment at CIFAR-100-C vs. LSUN-C under impulse noise}
    \label{fig:axis_align}
    \vspace{-1em}
\end{figure*}

\section{Asynchronous Covariate Shifts}

In~\cref{tab:asymmetric_type}, under unequal corruption-types, \ours~maintains over 90\% AUROC across all pairs.
In~\cref{tab:asymmetric_severity}, under unequal severities, it achieves nearly 100\% in most pairs, but drops to 0.29/0.36\% when ID has severity 5 and OOD has lower severity. 
We suspect that severely corrupted ID receives lower MSP than mildly corrupted OOD, so MSP-based flip correction is not triggered. 
However, such an ID/OOD-conditioned severity split is unlikely within a shared observation boundary.
We clarify that our real-world claims pertain to the shared-shift setting.

\definecolor{RdBuLow}{HTML}{D04747}    
\definecolor{RdBu90}{HTML}{5D8FC2}     
\definecolor{RdBu98}{HTML}{407AB7}     
\definecolor{RdBu99}{HTML}{3773B4}     
\definecolor{RdBu100}{HTML}{306CB1}    

\newcommand{\heatcell}[2]{%
    \cellcolor{#1}%
    \textcolor{white}{\strut #2}%
}

\newcommand{\diagheatcell}[2]{%
    \cellcolor{#1}%
    \textcolor[HTML]{000000}{\bfseries\strut #2}%
}

\begin{table}[h]
    \centering
    \caption{
        AUROC (\%) of \ours~under asymmetric ID/OOD covariate shifts.
        ImageNet-C vs. iNaturalist-C with RegNetY.
    }
    \label{tab:asymmetric_id_ood_shifts}
    \hfill
    \begin{minipage}[h]{0.49\linewidth}
        \centering
        \subcaption{Corruption-type}
        \label{tab:asymmetric_type}
        \setlength{\tabcolsep}{6pt}
        \resizebox{\linewidth}{!}{
            \begin{tabular}{c|ccc}
                \toprule
                \diagbox[width=2.5cm, height=1.3\line]{\textbf{ID}}{\textbf{OOD}}
                & Gauss. & Defoc. & Snow \\
                \midrule
                Gauss. & \diagheatcell{RdBu90}{90.37}  & \heatcell{RdBu90}{90.29} & \heatcell{RdBu100}{100} \\
                Defoc. & \heatcell{RdBu99}{99.53} & \diagheatcell{RdBu98}{97.85} & \heatcell{RdBu99}{99.24} \\
                Snow   & \heatcell{RdBu99}{99.37} & \heatcell{RdBu100}{99.79} & \diagheatcell{RdBu100}{99.96} \\
                \bottomrule
            \end{tabular}
        }
    \end{minipage}
    \hfill
    \begin{minipage}[h]{0.49\linewidth}
        \centering
        \subcaption{Severity}
        \label{tab:asymmetric_severity}
        \setlength{\tabcolsep}{6pt}
        \resizebox{\linewidth}{!}{
            \begin{tabular}{c|ccc}
                \toprule
                \diagbox[width=2.5cm, height=1.3\line]{\textbf{ID}}{\textbf{OOD}}
                & Sev. 1 & Sev. 3 & Sev. 5 \\
                \midrule
                Sev. 1 & \diagheatcell{RdBu100}{100} & \heatcell{RdBu100}{100} & \heatcell{RdBu100}{100} \\
                Sev. 3 & \heatcell{RdBu99}{99.51} & \diagheatcell{RdBu100}{99.62} & \heatcell{RdBu100}{99.65} \\
                Sev. 5 & \heatcell{RdBuLow}{0.29} & \heatcell{RdBuLow}{0.36} & \diagheatcell{RdBu90}{90.37} \\
                \bottomrule
            \end{tabular}
        }
    \end{minipage}
    \hfill
\end{table}


\section{Progression of Axis-Alignment} 
\label{app:axis_alignment}

\Cref{fig:axis_align} demonstrates the online convergence capability of \ours~in discovering the oracle discriminative axis. 
The stars denote the global ID/OOD centroids, whose connecting line is the oracle axis, while the triangles denote batch-wise ID/OOD prototypes estimated by \ours, whose connecting line is the online axis.
The cosine similarity between two axes increases across batches, demonstrating \ours's ability to navigate the high-dimensional feature space and progressively align with the true discriminative direction.

\begin{figure}[h]
    \centering
    \includegraphics[width=0.9\linewidth]{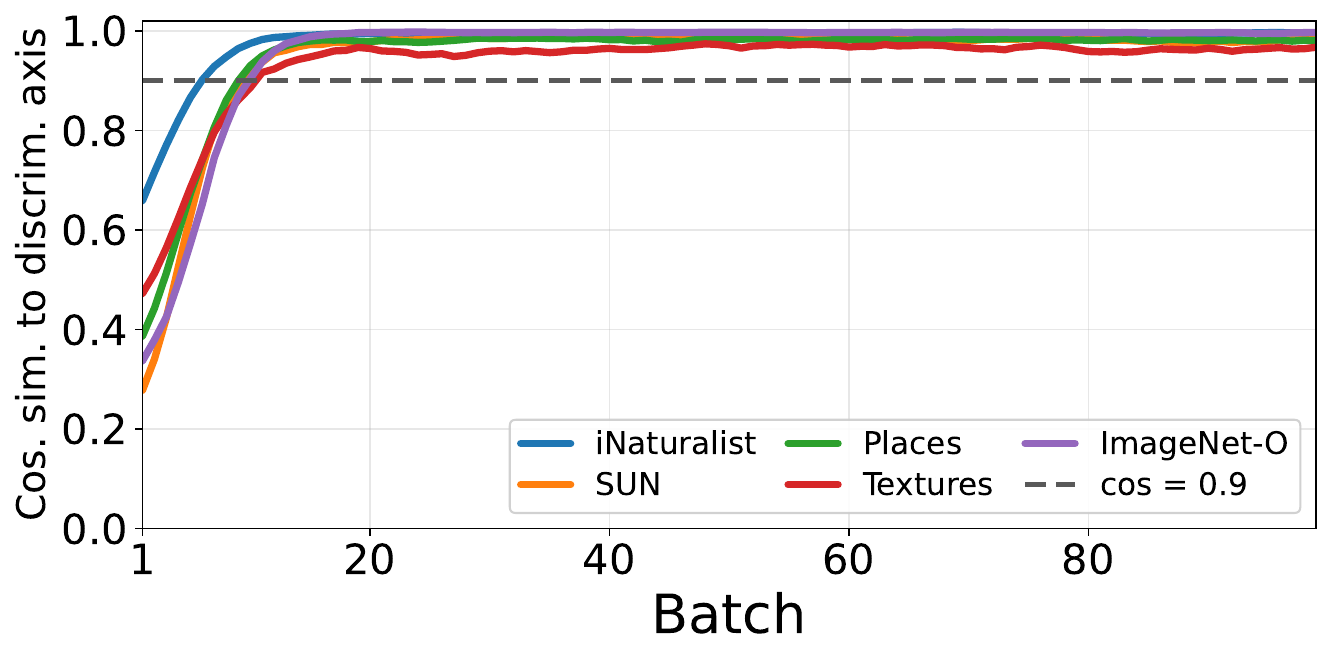}
    \caption{Cosine similarity between the tracked and oracle axes over test batches. Tracked axis progressively converges toward the oracle direction.}
    \label{fig:axis_alignment_cos}
\end{figure}

Prototype-alignment progress can also be considered as the recovery from the errors in the pseudo-partitions and prototype estimates. 
To directly evaluate this behavior and the claimed axis recovery,~\Cref{fig:axis_alignment_cos} plots the cosine similarity between the tracked and oracle axes over the test stream. 
Since the tracked axis is defined by the estimated ID/OOD prototypes, its alignment with the oracle axis provides a measure of the directional impact of pseudo-partitioning and prototype-estimation errors. 
Although the tracked axis is not yet well aligned in early batches, its cosine similarity rapidly increases and remains above 0.9 in later batches. 
This indicates that partial errors do not accumulate and that \ours~progressively recovers the oracle discriminative direction throughout the stream.

\section{System overhead}

\begin{figure}[h]
    \centering
    \includegraphics[width=\linewidth]{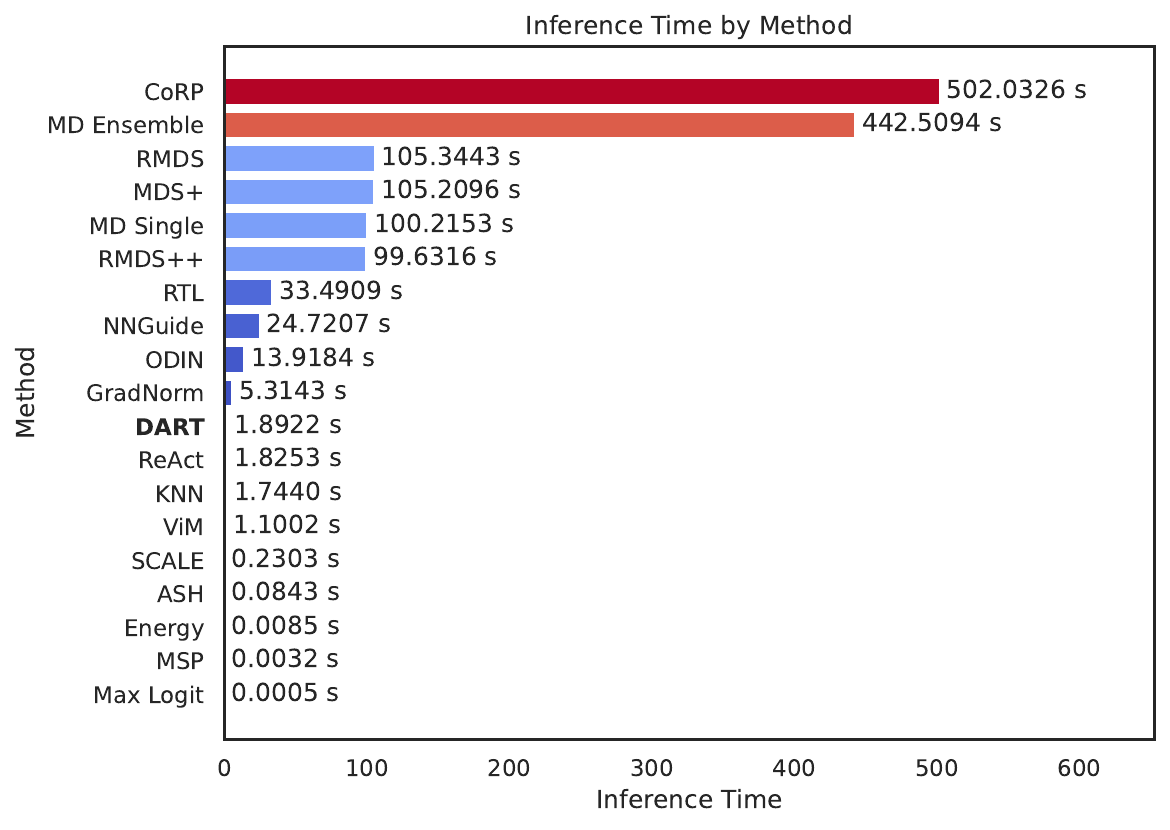}
    \caption{Inference time}
    \label{fig-inference_time}
\end{figure}

We measure wall-clock inference time for all methods on the same device, after the backbone forward pass, and only for the OOD-score computation (Figure~\ref{fig-inference_time}). Concretely, we report the total time required to process 100 mini-batches of size 200 (20k test samples in total), using RegNetY-16GF. Under this protocol, \ours~falls into the group of fast methods: it is markedly faster than recent baselines such as RTL, NNGuide, and MDS-based variants, which require regression fitting, KNN-style searches, or repeated Mahalanobis evaluations.

In \cref{tab:additional_computation_and_state}, we also compare additional computation and persistent state. 
\ours~requires only 0.29 GFLOPs per batch and 0.04 MB of persistent state.
In contrast, other methods, which show powerful performance, often require substantial computation and memory overhead.
For example, ODIN requires an additional forward pass and a full backward pass for input perturbation, while MDS++ performs class-wise Mahalanobis evaluation over all 1,000 ImageNet classes.
In terms of persistent state, KNN stores the full ImageNet training feature bank, whereas OODD retains a 50\%-sampled ID dictionary together with a dynamically updated OOD dictionary, resulting in multi-gigabyte storage.
These comparisons highlight the lightweight deployment footprint of \ours.

\begin{table}[h]
    \centering
    \caption{
        Additional computation and persistent state size of representative
        post-hoc and online OOD detection methods on ImageNet with a
        ResNet-50 backbone.
    }
    \label{tab:additional_computation_and_state}
    \setlength{\tabcolsep}{7pt}
    \renewcommand{\arraystretch}{1.12}
    \resizebox{\linewidth}{!}{
    \begin{tabular}{l|rr|rr}
        \toprule
        &
        \multicolumn{2}{c|}{Additional computation}
        &
        \multicolumn{2}{c}{Additional persistent state}
        \\
        \cmidrule(lr){2-3}
        \cmidrule(lr){4-5}

        Method
        &
        \shortstack{Arithmetic\\(GFLOPs / batch)}
        &
        \shortstack{Overhead\\(\% of one forward)}
        &
        \shortstack{Number of\\FP32 values}
        &
        \shortstack{State size\\(MB)}
        \\
        \midrule
        \midrule

        MSP
        &
        $0.0006$
        &
        $<0.0001\%$
        &
        $0$
        &
        $0$
        \\

        ODIN
        &
        $4{,}907.02$
        &
        $300.00\%$
        &
        $2$
        &
        $8\times10^{-6}$
        \\

        RTL
        &
        $58.96$
        &
        $3.60\%$
        &
        $4{,}196{,}352$
        &
        $16.79$
        \\

        ViM
        &
        $0.86$
        &
        $0.06\%$
        &
        $2{,}148{,}353$
        &
        $8.60$
        \\

        KNN
        &
        $1{,}049.53$
        &
        $64.17\%$
        &
        $2{,}623{,}830{,}016$
        &
        $10{,}495.32$
        \\

        CoRP
        &
        $10.07$
        &
        $0.62\%$
        &
        $12{,}193{,}792$
        &
        $48.78$
        \\

        MDS++
        &
        $1{,}841.97$
        &
        $112.61\%$
        &
        $6{,}242{,}304$
        &
        $24.97$
        \\
        
        OODD
        &
        $526.55$
        &
        $32.19\%$
        &
        $1{,}316{,}374{,}528$
        &
        $5{,}265.50$
        \\

        \midrule
        
        \rowcolor{blue!5}
        \ours
        &
        $0.29$
        &
        $0.02\%$
        &
        $9{,}680$
        &
        $0.04$
        \\

        \bottomrule
    \end{tabular}
    }

\end{table}


\newpage

\section{Visualization of ROC curves}

\begin{figure}[h]
    \centering
    \begin{minipage}{\linewidth}
        \centering
        \includegraphics[width=\linewidth]{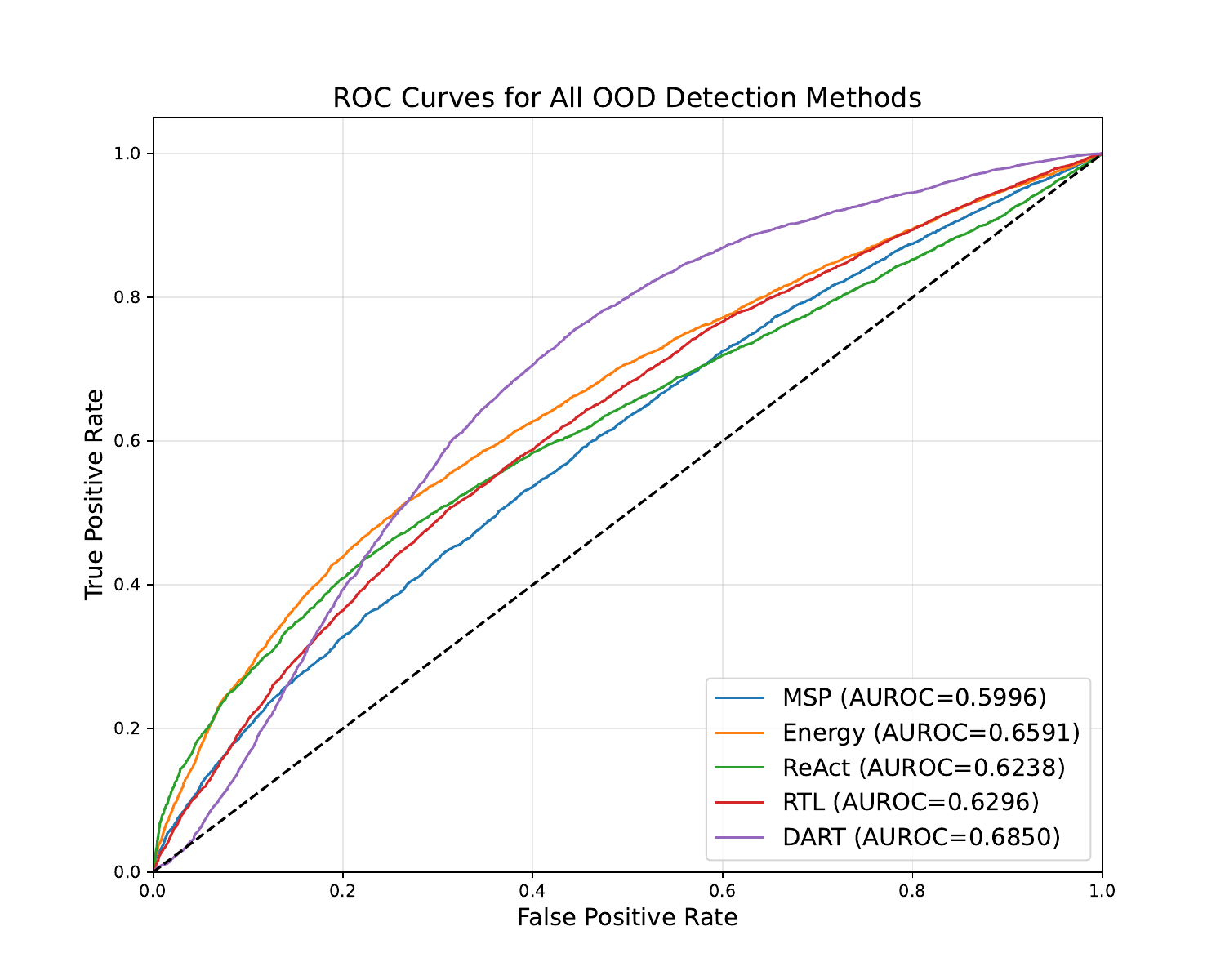}
        \subcaption{Glass blur}
        \label{fig:glassblur}
    \end{minipage}
    \hfill
    \begin{minipage}{\linewidth}
        \centering
        \includegraphics[width=\linewidth]{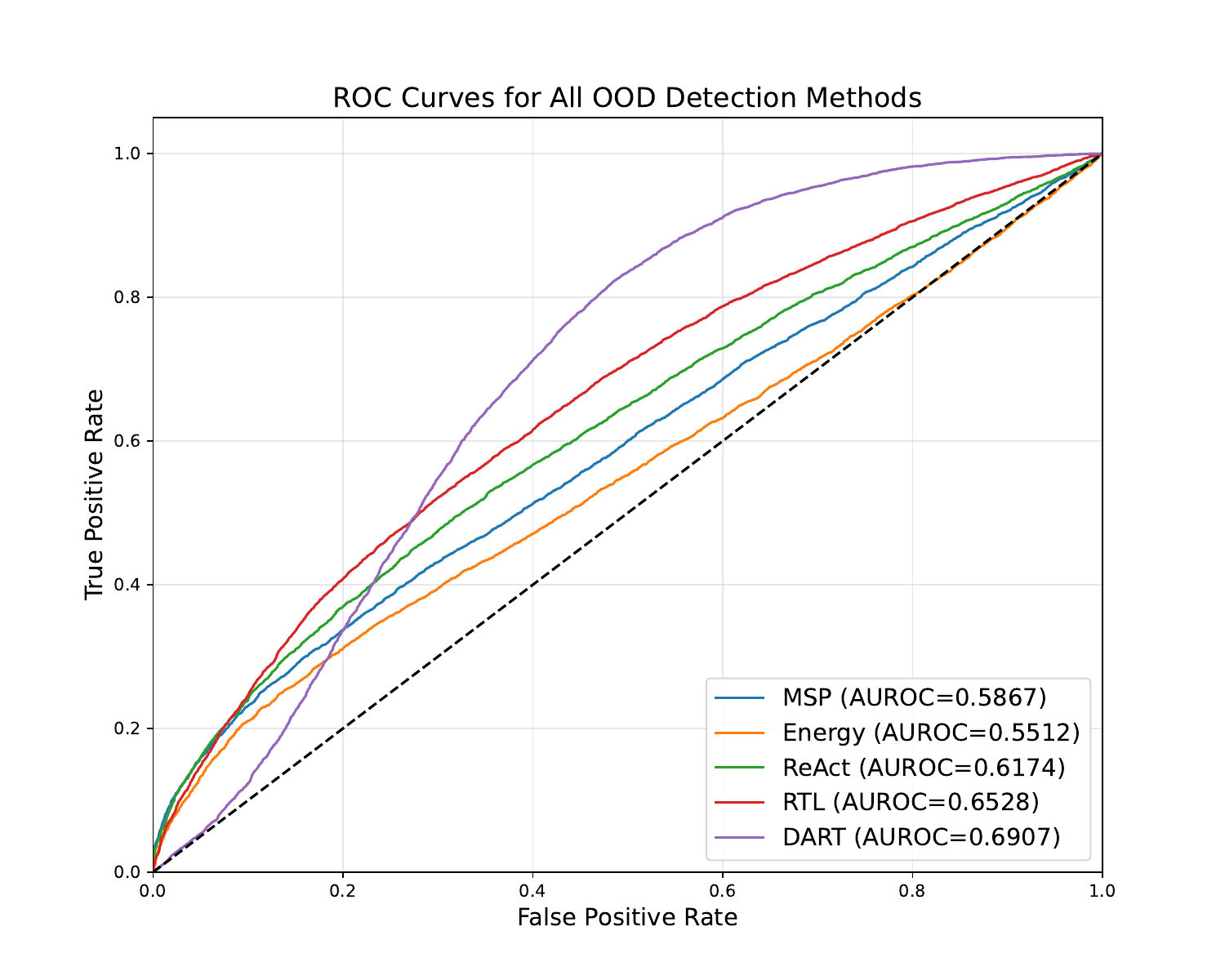}
        \subcaption{Pixelate}
        \label{fig:pixelate}
    \end{minipage}
    \caption{ROC curves for CIFAR-100 vs. LSUN under different corruptions}
    \label{fig:roc_curves}
\end{figure}

We visualize the ROC curves of our method and the baselines across several evaluation settings. As shown in Figure~\ref{fig:roc_curves}, \ours~achieves lower FPR in the high-TPR region even when overall AUROC is comparable to baselines. Since real-world deployment requires maintaining high ID acceptance while minimizing false alarms, FPR@95TPR better captures the performance that matters in practice.

\newpage

\section{Extended RDS distribution visualizations}

In the main paper, we present representative visualizations of RDS distributions for a subset of corruption types, to illustrate how ID and OOD samples are separated across different feature levels. These visualizations support our observation that the most discriminative layer can vary depending on the type of corruption. For completeness, we provide the full set of visualizations covering all 15 corruption types in this appendix.

\subsection{RDS distribution visualizations}

\begin{figure}[h]
    \centering
    \begin{subfigure}[b]{0.85\linewidth}
        \centering
        \includegraphics[width=\linewidth]{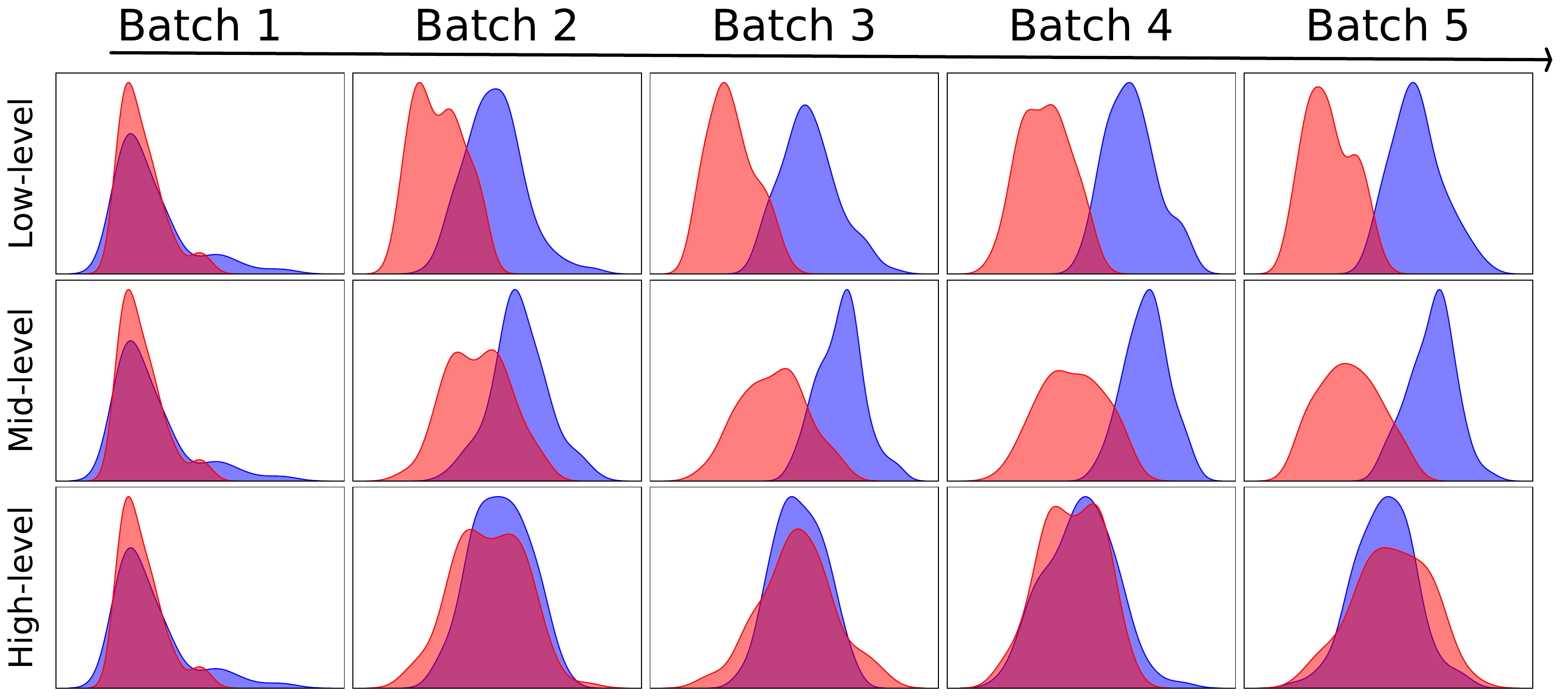}
        \caption{RDS under gaussian noise}
        \label{fig:appendix_rds_distribution_gaus}
    \end{subfigure}
    \hfill
    \begin{subfigure}[b]{0.85\linewidth}
        \centering
        \includegraphics[width=\linewidth]{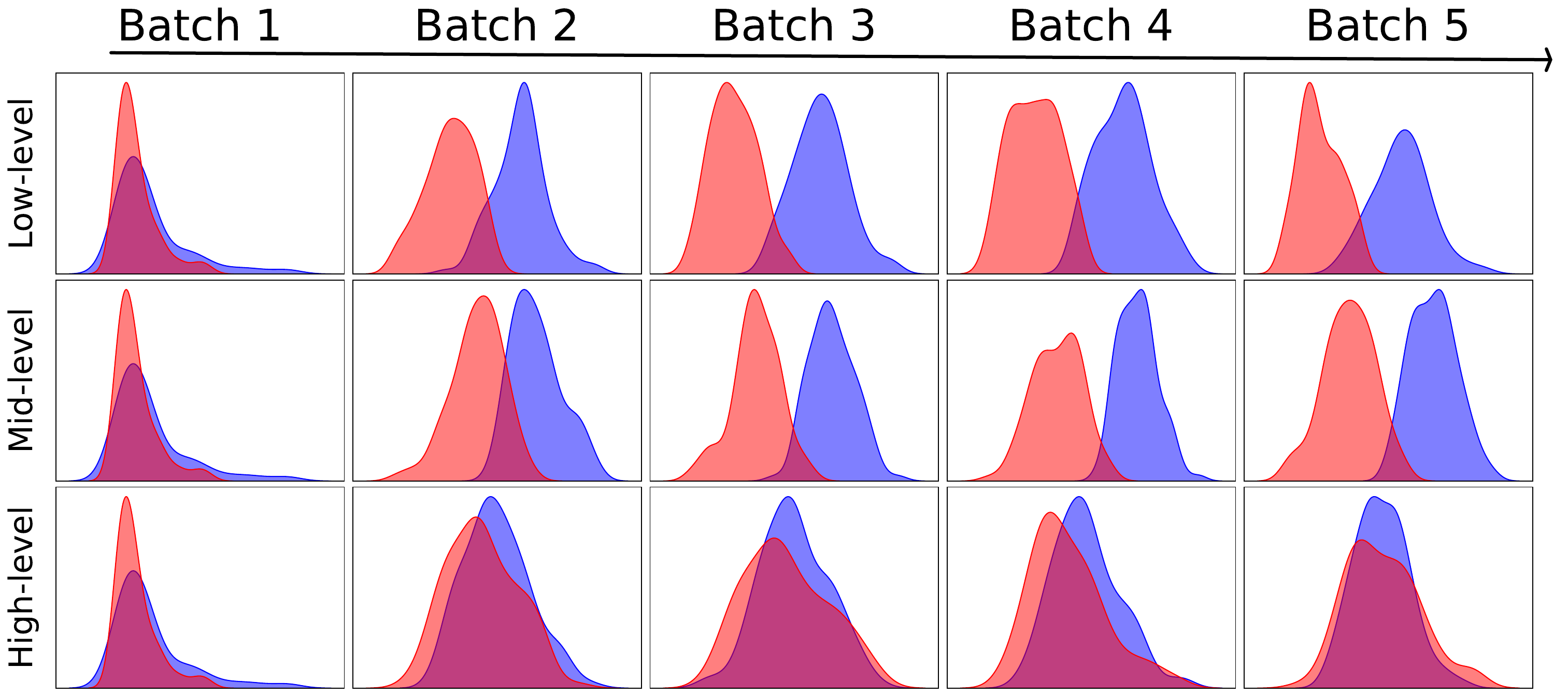}
        \caption{RDS under shot noise}
        \label{fig:appendix_rds_distribution_shot}
    \end{subfigure}
    \hfill
    \begin{subfigure}[b]{0.85\linewidth}
        \centering
        \includegraphics[width=\linewidth]{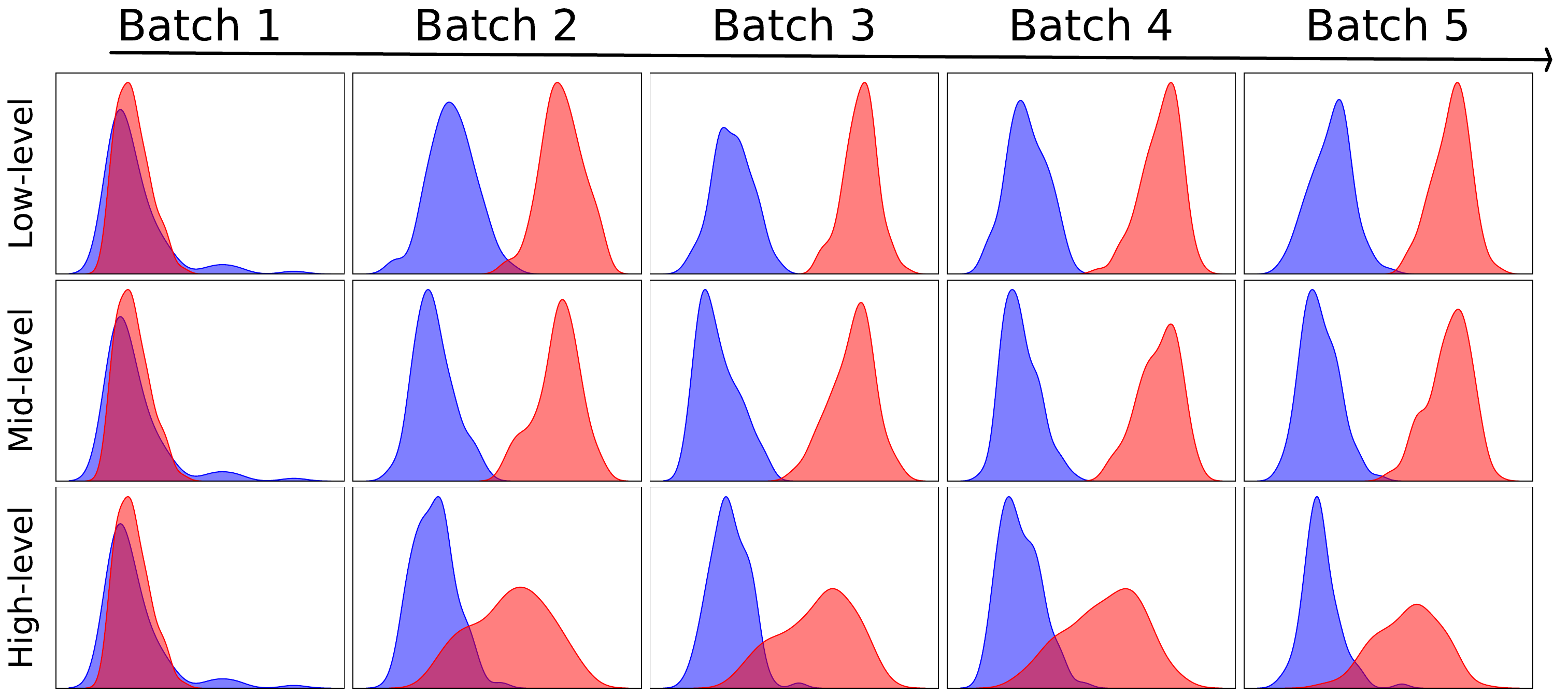}
        \caption{RDS under impulse noise}
        \label{fig:appendix_rds_distribution_impul}
    \end{subfigure}
    \begin{subfigure}[b]{0.85\linewidth}
        \centering
        \includegraphics[width=\linewidth]{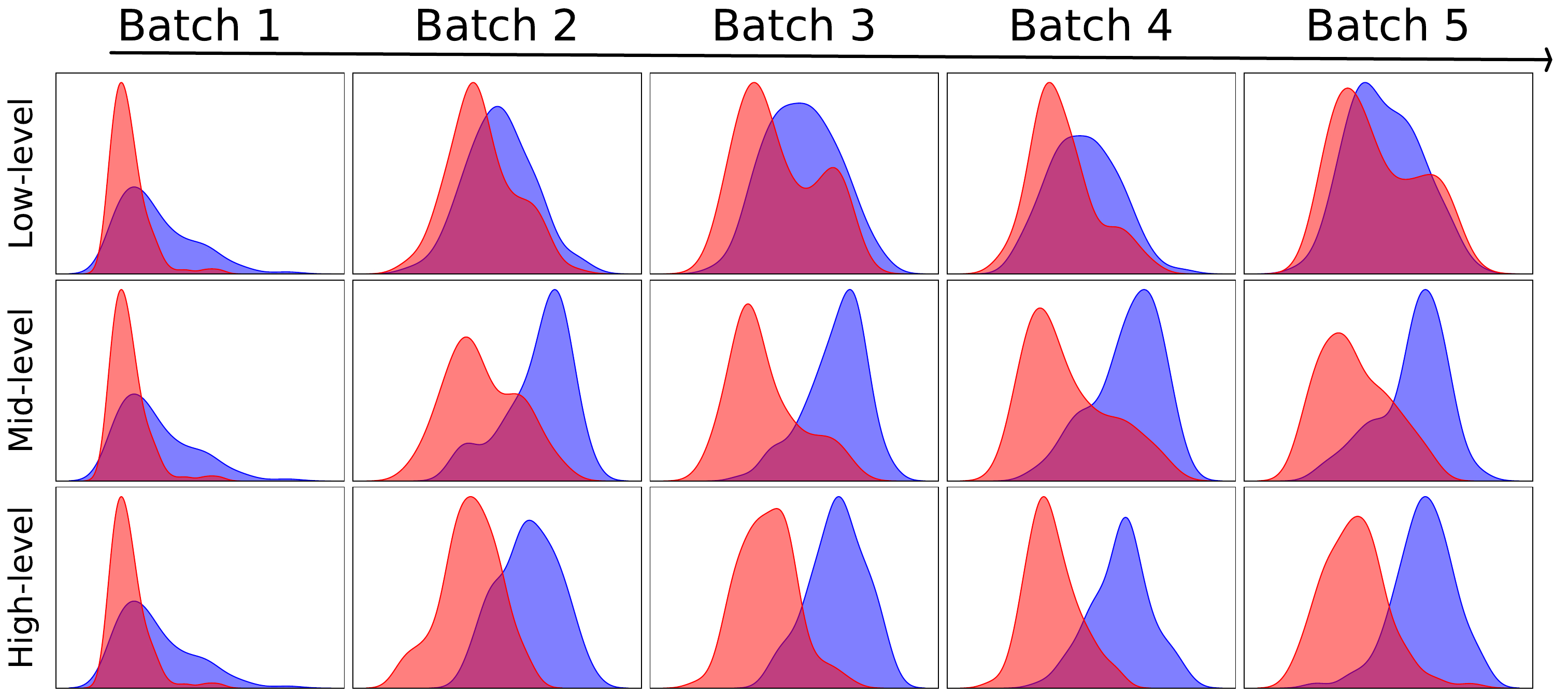}
        \caption{RDS under defocus blur}
        \label{fig:appendix_rds_distribution_defocus}
    \end{subfigure}
\end{figure}

\newpage

\begin{figure}[h]\ContinuedFloat
    \centering
    \begin{subfigure}[b]{0.85\linewidth}
        \centering
        \includegraphics[width=\linewidth]{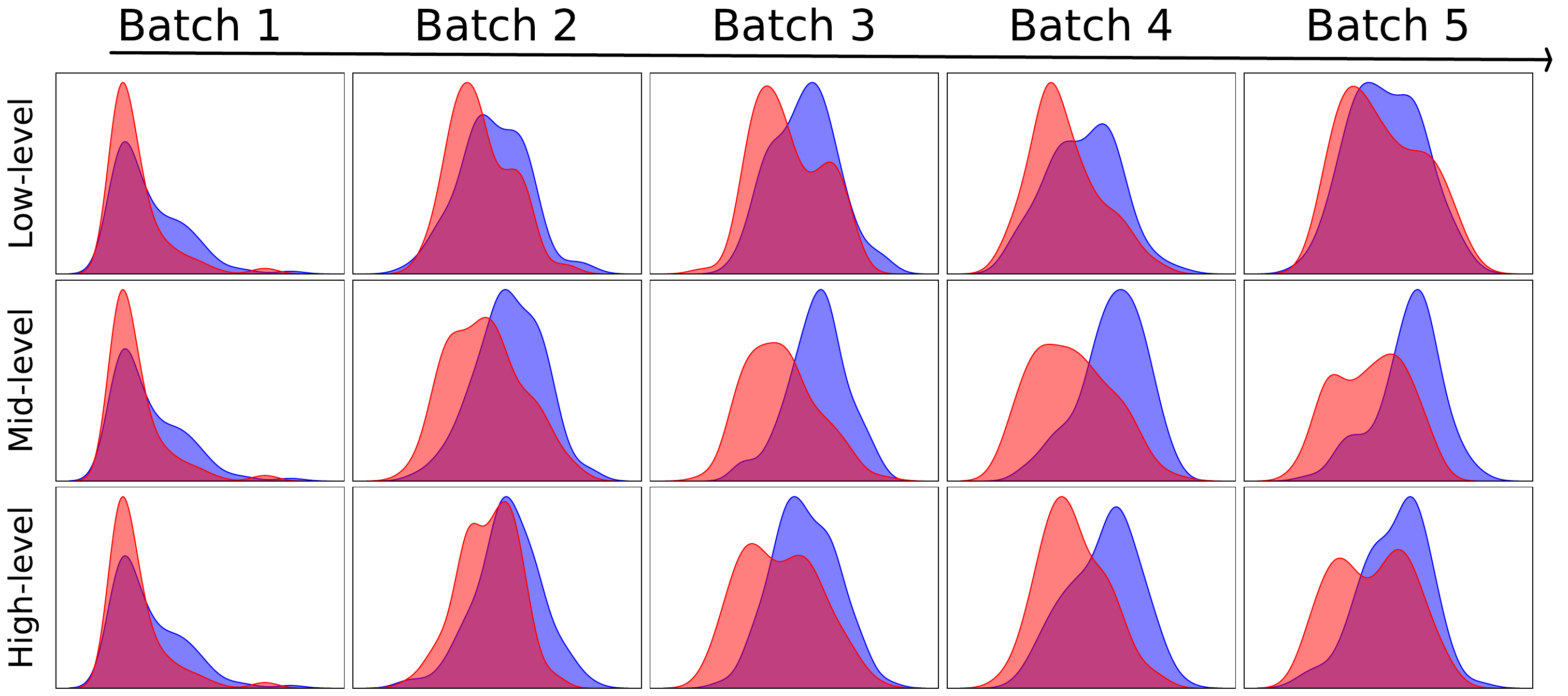}
        \caption{RDS under glass blur}
        \label{fig:appendix_rds_distribution_glass}
    \end{subfigure}
    \hfill
    \begin{subfigure}[b]{0.85\linewidth}
        \centering
        \includegraphics[width=\linewidth]{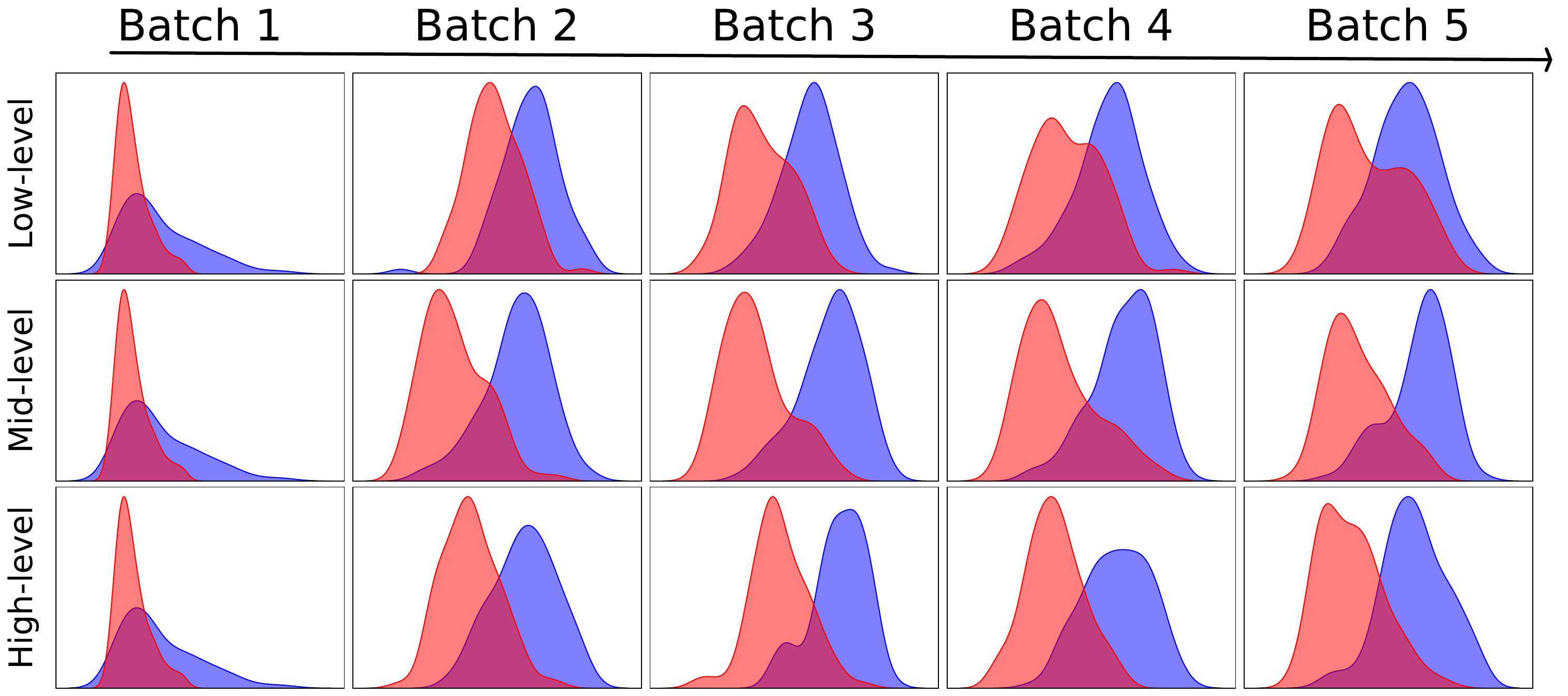}
        \caption{RDS under motion blur}
        \label{fig:appendix_rds_distribution_motion}
    \end{subfigure}
    \hfill
    \begin{subfigure}[b]{0.85\linewidth}
        \centering
        \includegraphics[width=\linewidth]{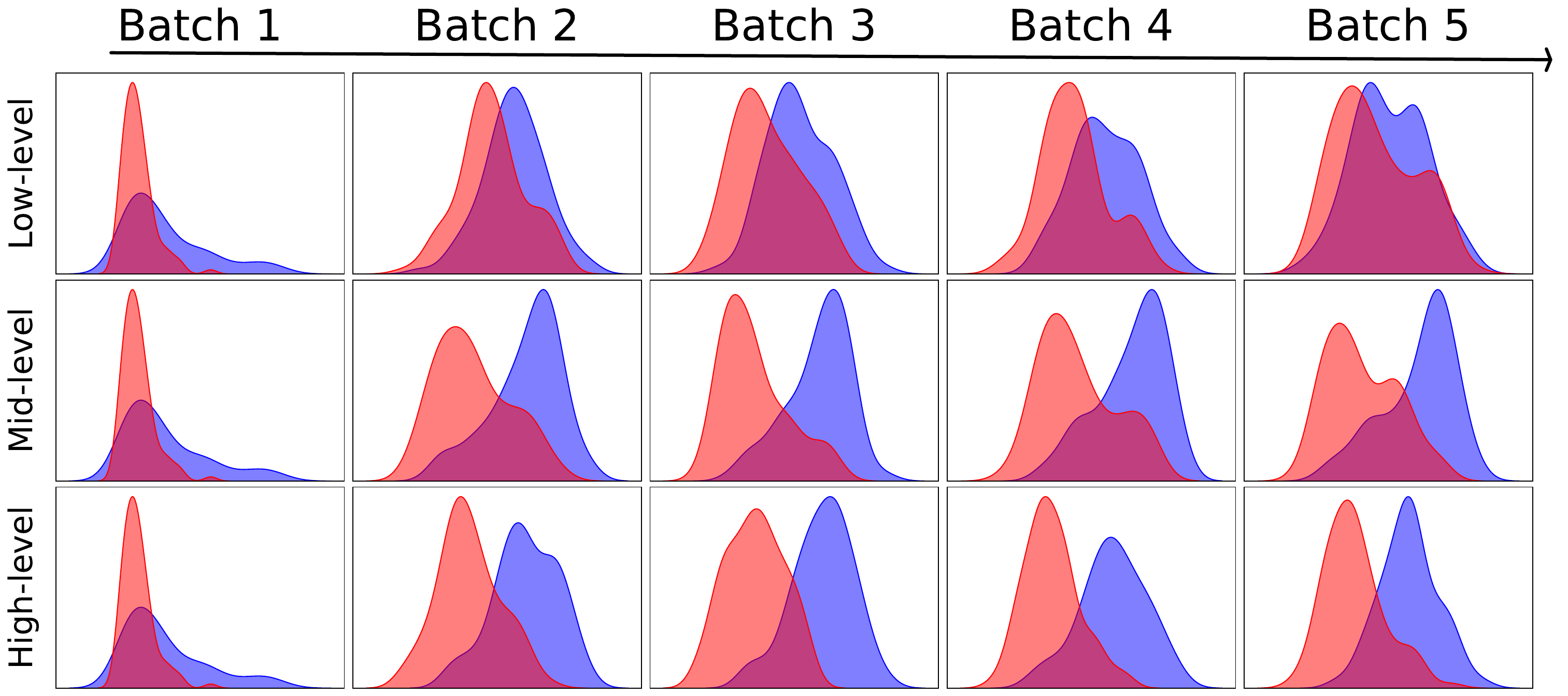}
        \caption{RDS under zoom blur}
        \label{fig:appendix_rds_distribution_zoom}
    \end{subfigure}
    \hfill
    \begin{subfigure}[b]{0.85\linewidth}
        \centering
        \includegraphics[width=\linewidth]{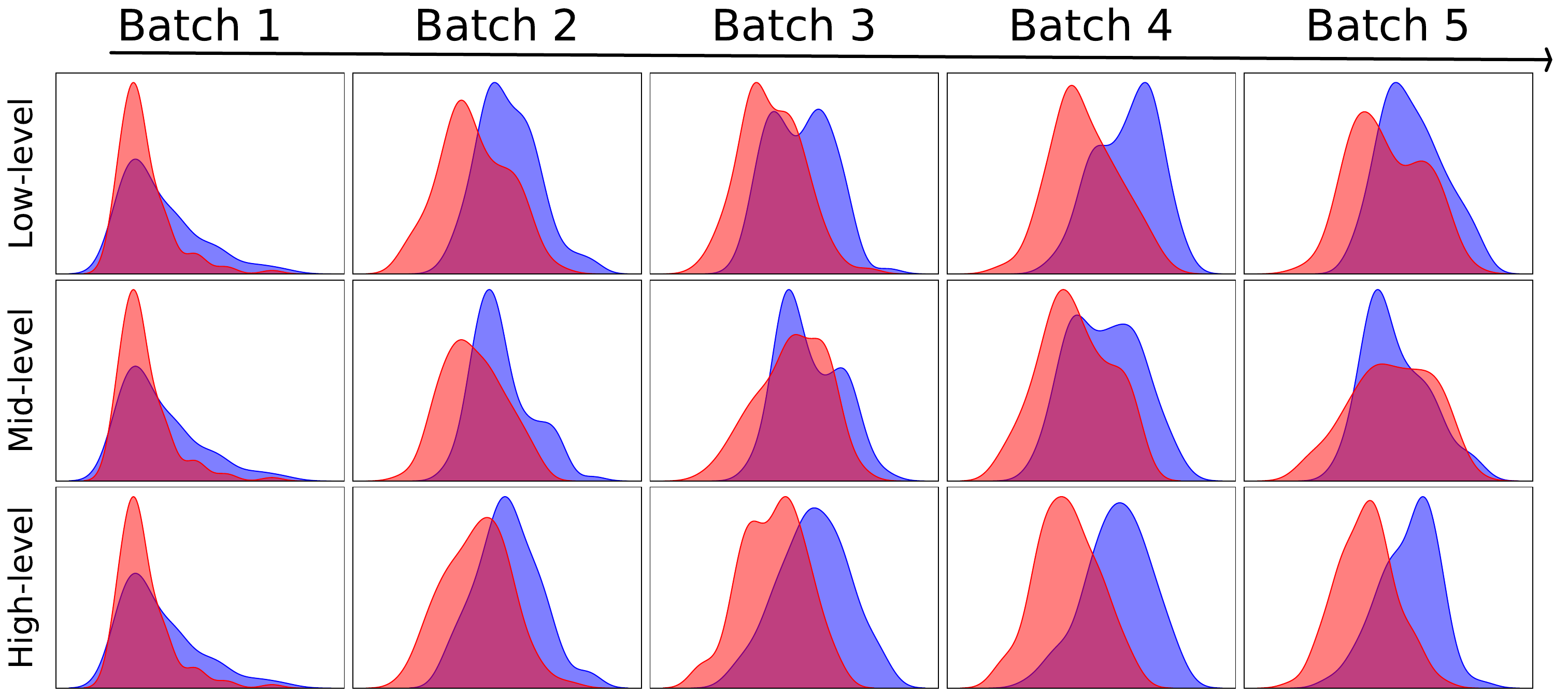}
        \caption{RDS under snow}
        \label{fig:appendix_rds_distribution_snow}
    \end{subfigure}
    \hfill
    \begin{subfigure}[b]{0.85\linewidth}
        \centering
        \includegraphics[width=\linewidth]{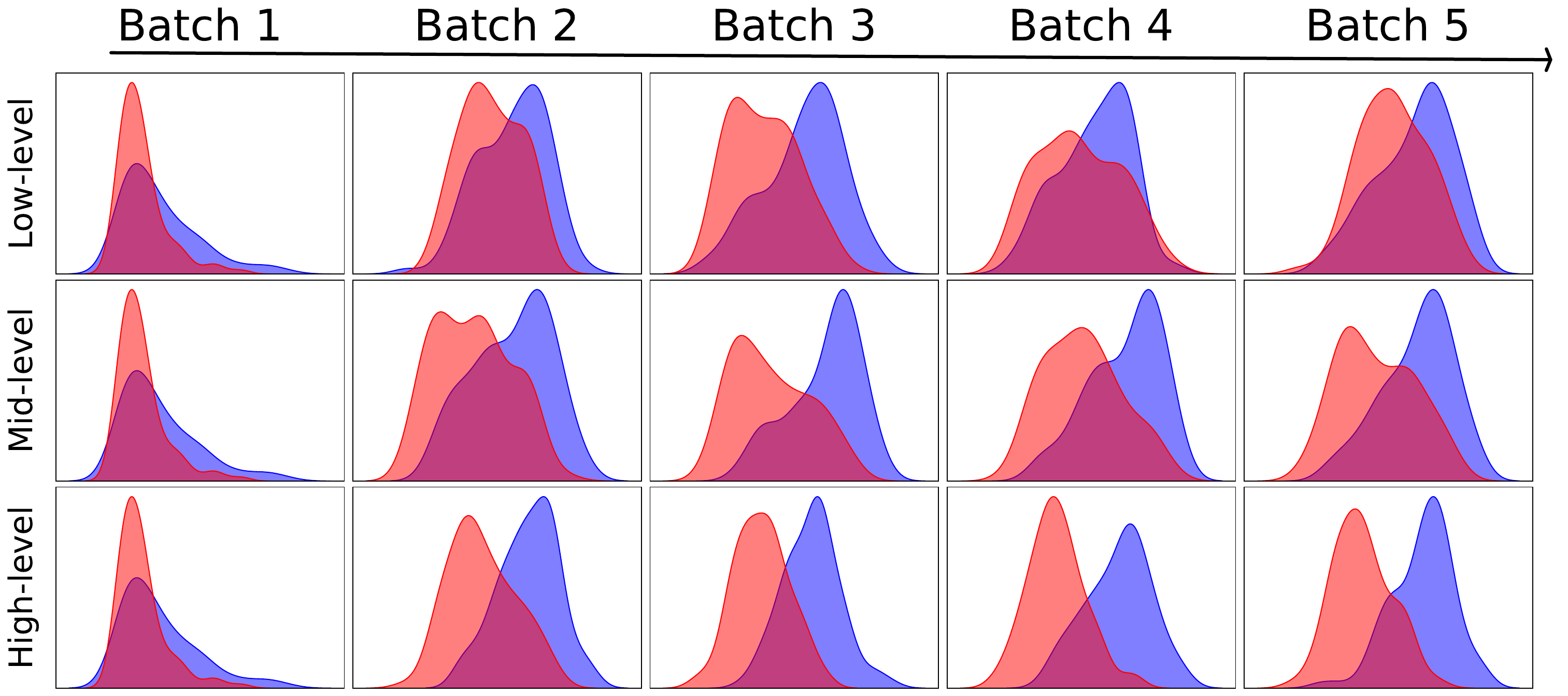}
        \caption{RDS under frost}
        \label{fig:appendix_rds_distribution_frost}
    \end{subfigure}
    \hfill
    \begin{subfigure}[b]{0.85\linewidth}
        \centering
        \includegraphics[width=\linewidth]{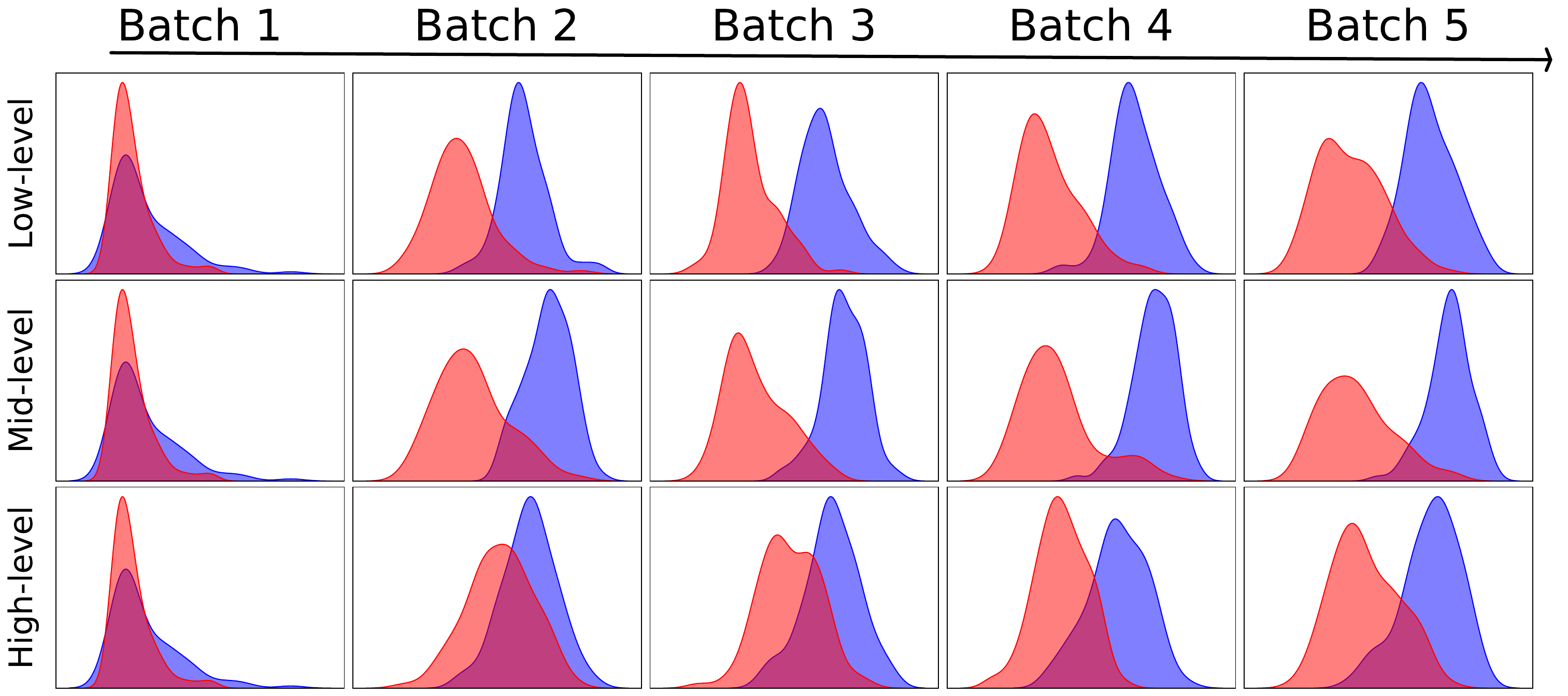}
        \caption{RDS under fog}
        \label{fig:appendix_rds_distribution_fog}
    \end{subfigure}
\end{figure}

\newpage

\begin{figure}[h]\ContinuedFloat
    \centering
    \begin{subfigure}[b]{0.85\linewidth}
        \centering
        \includegraphics[width=\linewidth]{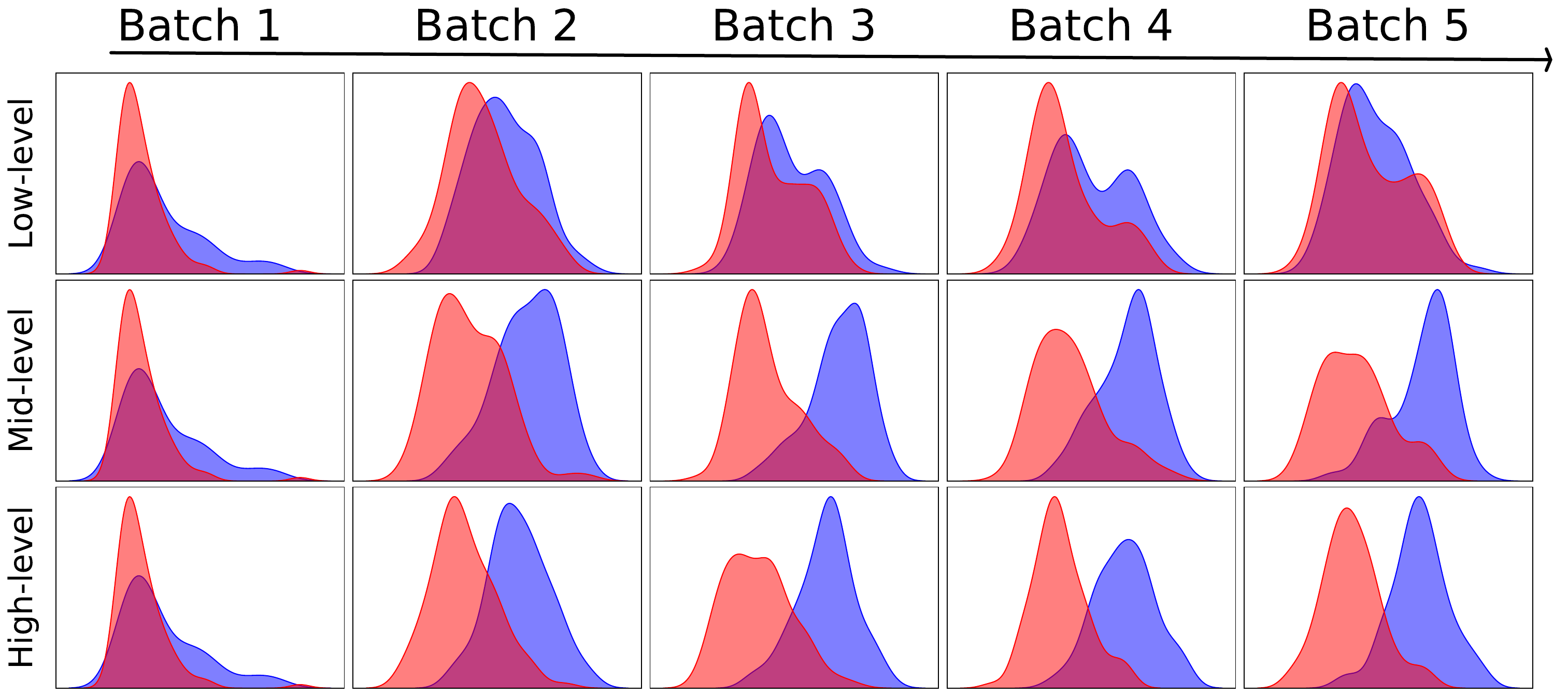}
        \caption{RDS under brightness}
        \label{fig:appendix_rds_distribution_bright}
    \end{subfigure}
    \hfill
    \begin{subfigure}[b]{0.85\linewidth}
        \centering
        \includegraphics[width=\linewidth]{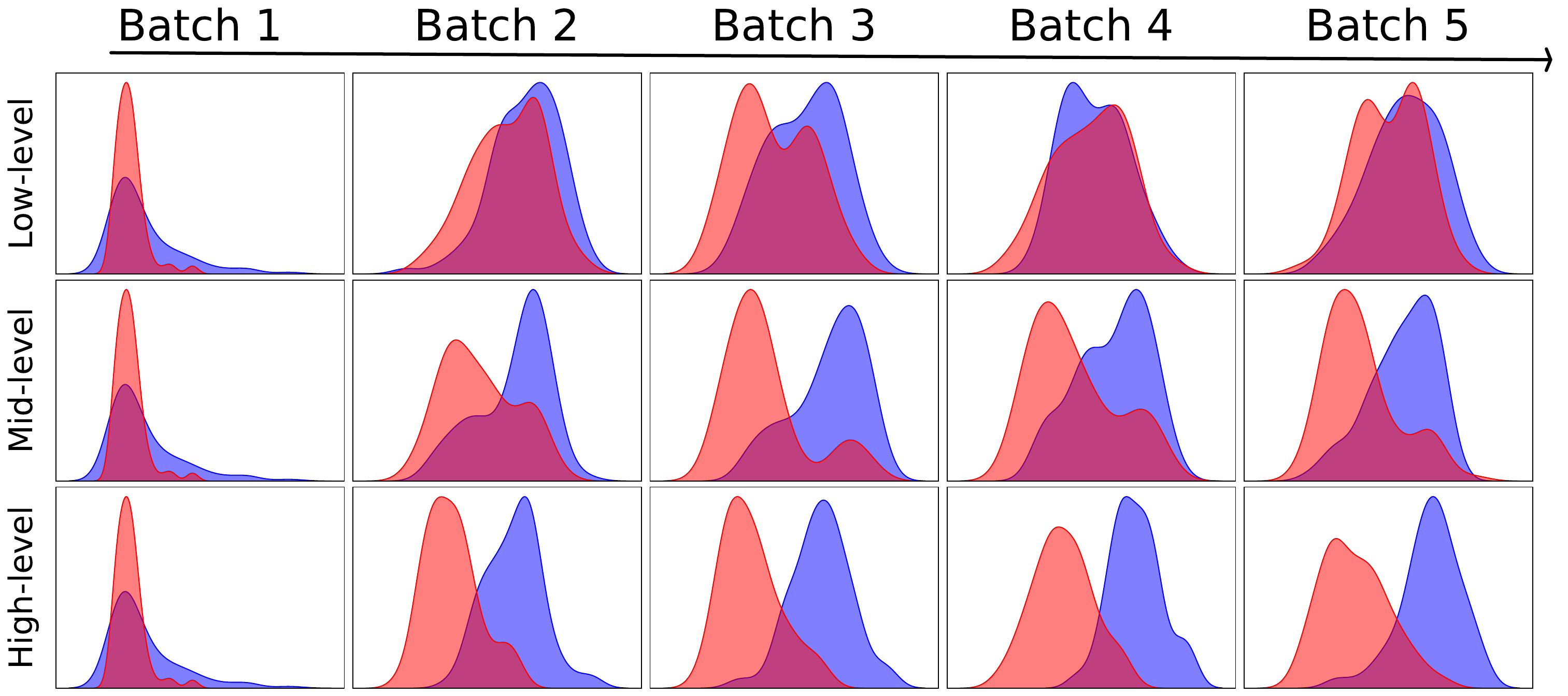}
        \caption{RDS under contrast}
        \label{fig:appendix_rds_distribution_contrast}
    \end{subfigure}
    \label{fig:appendix_rds_distribution_1}
    \hfill
    \begin{subfigure}[b]{0.85\linewidth}
        \centering
        \includegraphics[width=\linewidth]{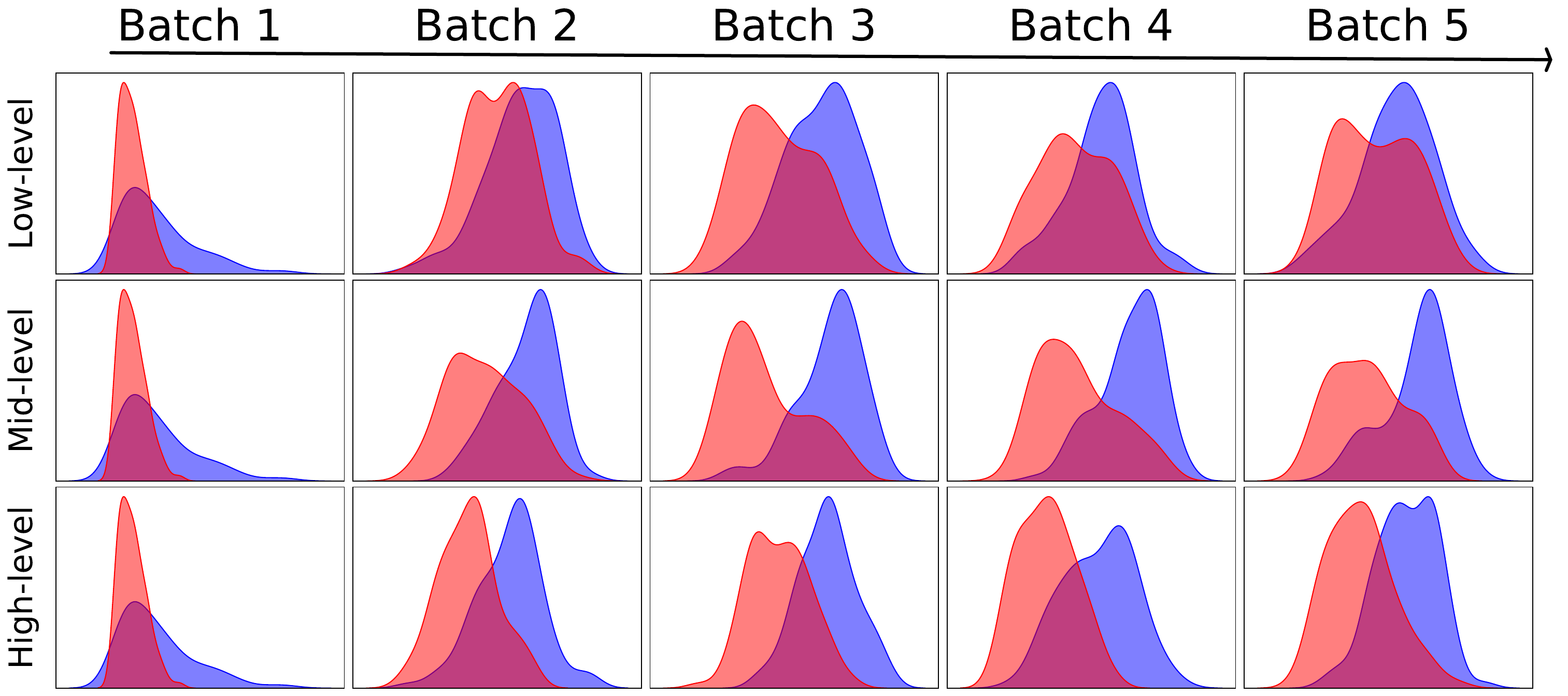}
        \caption{RDS under elastic transform}
        \label{fig:appendix_rds_distribution_elastic}
    \end{subfigure}
    \hfill
    \begin{subfigure}[b]{0.85\linewidth}
        \centering
        \includegraphics[width=\linewidth]{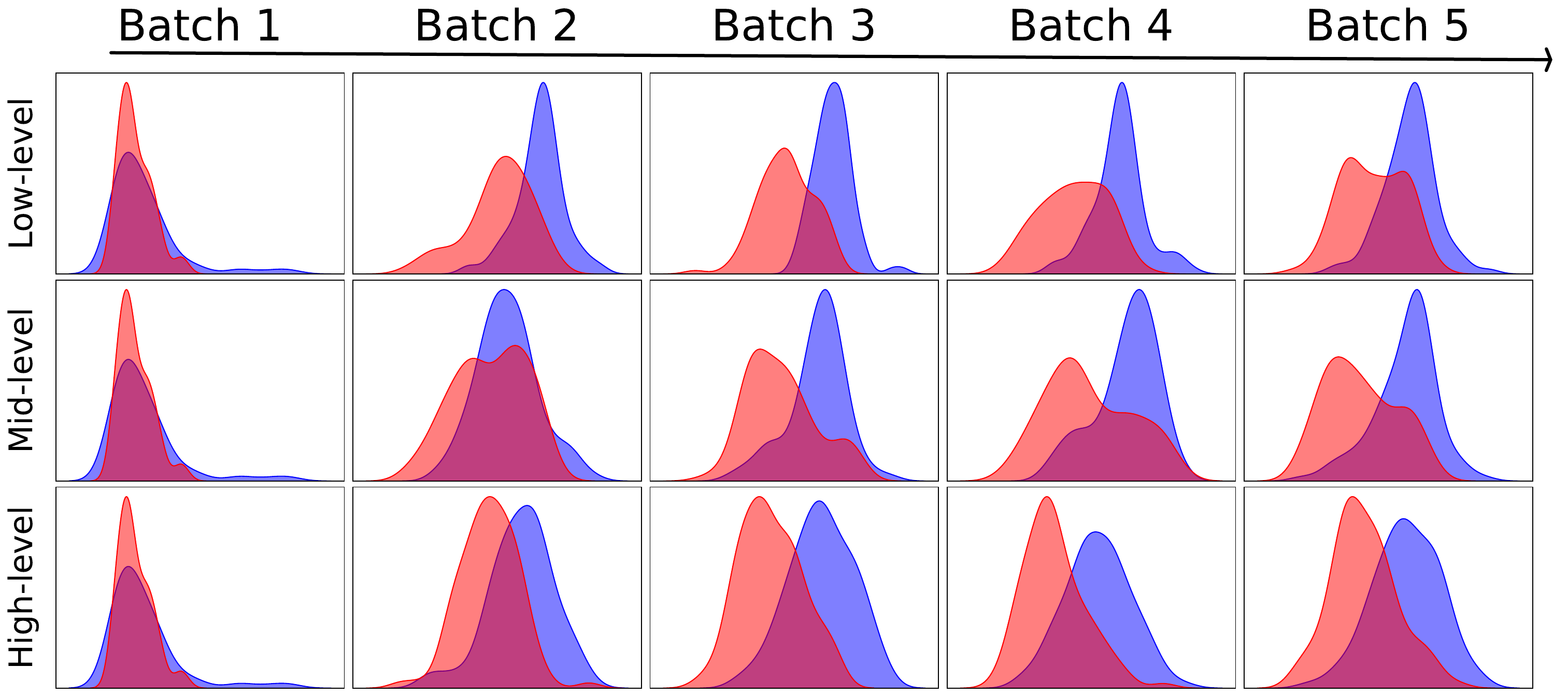}
        \caption{RDS under pixelate}
        \label{fig:appendix_feat_distribution_pixel}
    \end{subfigure}
    \hfill
    \begin{subfigure}[b]{0.85\linewidth}
        \centering
        \includegraphics[width=\linewidth]{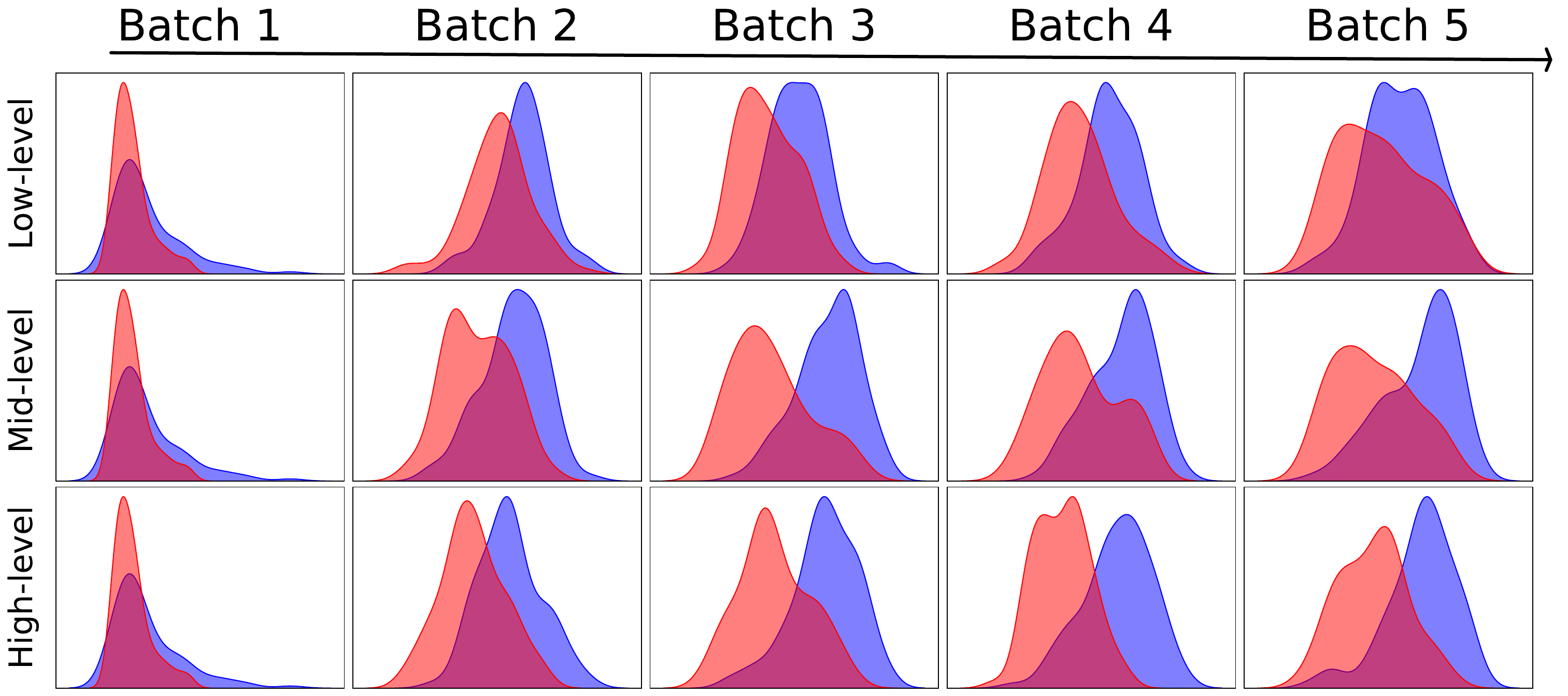}
        \caption{RDS under jpeg compression}
        \label{fig:appendix_rds_distribution_jpeg}
    \end{subfigure}
    
    \caption{Full corruptions visualizations for RDS distributions. Each plot shows RDS distributions at different network depths (low, mid, high-level) through five sequential batches. 
    The visualizations reveal how different corruption types affect RDS separability at specific network layers.
    }
    \label{fig:appendix_rds_distribution_2}
\end{figure}

\onecolumn

\newpage

\section{Full Results}
Here we show the full results for all OOD datasets which was abbreviated as average in the main manuscript due to space limits.

\begin{table*}[h]
    \centering
    \caption{OOD detection performance comparison with CIFAR-100 ID and the corresponding OODs. FPR@95TPR (\%) is lower the better and AUROC (\%) is higher the better. (Best: \textbf{bolded}, Second-best: \underline{underlined})}
    \resizebox{\linewidth}{!}{
    \begin{tabular}{l|c|c@{\hspace{2pt}}c|c@{\hspace{2pt}}c|c@{\hspace{2pt}}c|c@{\hspace{2pt}}c|c@{\hspace{2pt}}c||c@{\hspace{2pt}}c}
        \toprule
        \rowcolor{gray!15}
        \multicolumn{14}{c}{\textbf{Clean}} \\
        \midrule
        \multirow{2}{*}{Method}
        & \multirow{2}{*}{\makecell[c]{Training set \\ prior}}
        & \multicolumn{2}{c|}{SVHN}
        & \multicolumn{2}{c|}{Places365}
        & \multicolumn{2}{c|}{LSUN}
        & \multicolumn{2}{c|}{iSUN}
        & \multicolumn{2}{c||}{Textures}
        & \multicolumn{2}{c}{Average} \\
        \cmidrule(lr){3-14}
        & & FPR95 \(\downarrow\) & AUROC \(\uparrow\)
        & FPR95 \(\downarrow\) & AUROC \(\uparrow\)
        & FPR95 \(\downarrow\) & AUROC \(\uparrow\)
        & FPR95 \(\downarrow\) & AUROC \(\uparrow\)
        & FPR95 \(\downarrow\) & AUROC \(\uparrow\)
        & FPR95 \(\downarrow\) & AUROC \(\uparrow\) \\
        \midrule
        MSP & \cellcolor{blue!10} NO & 79.32 & 77.27 & 80.37 & 74.99 & 78.39 & 76.91 & 81.23 & 74.56 & 82.55 & 72.70 & 80.37 & 75.29 \\
        Energy & \cellcolor{blue!10} NO & 80.32 & 78.74 & 78.47 & 75.69 & 78.39 & 78.68 & 83.42 & 74.53 & 79.17 & 76.31 & 79.95 & 76.79 \\
        Max logit & \cellcolor{blue!10} NO & 79.85 & 79.11 & 78.63 & 76.01 & 77.87 & 78.92 & 82.49 & 75.00 & 79.69 & 76.25 & 79.71 & 77.06 \\
        GradNorm & \cellcolor{blue!10} NO & 96.80 & 44.47 & 94.81 & 51.36 & 98.22 & 32.12 & 98.48 & 30.97 & 86.31 & 59.18 & 94.92 & 43.62 \\
        ODIN & \cellcolor{blue!10} NO & 64.15 & 80.96 & 84.30 & 69.48 & 90.32 & 63.24 & 91.03 & 61.21 & 75.75 & 72.60 & 81.11 & 69.50 \\
        SCALE & \cellcolor{blue!10} NO & 74.38 & 81.36 & 78.47 & 75.55 & 80.91 & 75.14 & 83.88 & 72.60 & 65.34 & 82.98 & 76.60 & 77.53 \\
        ASH & \cellcolor{blue!10} NO & 79.20 & 79.65 & 79.96 & 75.11 & 81.64 & 77.12 & 85.45 & 73.20 & 75.96 & 78.72 & 80.44 & 76.76 \\
        RTL & \cellcolor{blue!10} NO & 50.15 & 87.42 & 73.23 & 75.59 & 58.46 & 85.01 & 68.38 & 80.67 & 70.33 & 75.12 & 64.11 & 80.76 \\
        ViM & \cellcolor{red!10} YES & 54.18 & 85.44 & 86.74 & 64.33 & 66.48 & 81.52 & 66.82 & 80.95 & 85.47 & 65.83 & 71.94 & 75.61 \\
        KNN & \cellcolor{red!10} YES & 63.34 & 86.06 & 80.09 & 73.38 & 66.57 & 84.85 & 72.86 & 80.60 & 74.54 & 80.89 & 71.48 & 81.16 \\
        MDS\textsubscript{single} & \cellcolor{red!10} YES & 77.75 & 75.58 & 90.94 & 59.26 & 69.56 & 80.02 & 70.64 & 78.54 & 91.15 & 56.50 & 80.01 & 69.98 \\
        MDS\textsubscript{ensemble} & \cellcolor{red!10} YES & 62.92 & 88.78 & 93.41 & 61.12 & \textbf{13.12} & \textbf{97.34} & \underline{16.44} & \underline{96.55} & 43.79 & 87.99 & 45.94 & 86.36 \\
        ReAct & \cellcolor{red!10} YES & 87.25 & 70.36 & 79.97 & 75.71 & 73.91 & 80.58 & 75.75 & 78.48 & 80.39 & 75.32 & 79.45 & 76.09 \\
        NNGuide & \cellcolor{red!10} YES & 86.85 & 71.34 & 87.71 & 67.73 & 94.66 & 59.74 & 95.50 & 57.33 & 73.01 & 75.32 & 87.55 & 66.29 \\
        CoRP & \cellcolor{red!10} YES & 43.43 & 91.59 & 86.30 & 68.31 & 82.53 & 77.18 & 80.56 & 77.13 & \underline{34.89} & \underline{91.04} & 65.54 & 81.05 \\
        MDS++ & \cellcolor{red!10} YES & 83.59 & 81.25 & 85.42 & 70.98 & 84.16 & 77.33 & 86.64 & 73.88 & 90.00 & 74.15 & 85.96 & 75.52 \\
        RMDS & \cellcolor{red!10} YES & 68.73 & 85.23 & 77.53 & 77.11 & 52.64 & 88.28 & 57.46 & 85.64 & 83.13 & 76.29 & 67.90 & 82.51 \\
        RMDS++ & \cellcolor{red!10} YES & 74.20 & 84.57 & 76.79 & \underline{77.67} & 65.24 & 85.70 & 71.02 & 82.49 & 82.29 & 77.44 & 73.91 & 81.57 \\
        OODD & \cellcolor{red!10} YES & \underline{22.11} & \underline{95.79} & \textbf{65.14} & \textbf{81.34} & \underline{23.48} & \underline{95.16} & 29.27 & 93.38 & \textbf{25.94} & \textbf{94.00} & \underline{33.19} & \textbf{91.93} \\
\rowcolor{blue!5}
        \ours & \cellcolor{blue!10} NO & \textbf{9.64} & \textbf{97.67} & \underline{70.00} & 75.12 & 30.24 & 91.49 & \textbf{14.79} & \textbf{96.62} & 40.91 & 88.02 & \textbf{33.12} & \underline{89.78} \\
        \bottomrule
    \end{tabular}
    }
\end{table*}

\begin{table*}[h]
    \centering
    \caption{OOD detection performance comparison with ImageNet ID and the corresponding OODs. FPR@95TPR (\%) is lower the better and AUROC (\%) is higher the better. (Best: \textbf{bolded}, Second-best: \underline{underlined})}
    \resizebox{\linewidth}{!}{
    \begin{tabular}{l|c|c@{\hspace{2pt}}c|c@{\hspace{2pt}}c|c@{\hspace{2pt}}c|c@{\hspace{2pt}}c|c@{\hspace{2pt}}c||c@{\hspace{2pt}}c}
        \toprule
        \rowcolor{gray!15}
        \multicolumn{14}{c}{\textbf{Clean}} \\
        \midrule
        \multirow{2}{*}{Method}
        & \multirow{2}{*}{\makecell[c]{Training set \\ prior}}
        & \multicolumn{2}{c|}{ImageNet-O}
        & \multicolumn{2}{c|}{Places}
        & \multicolumn{2}{c|}{SUN}
        & \multicolumn{2}{c|}{iNaturalist}
        & \multicolumn{2}{c||}{Textures}
        & \multicolumn{2}{c}{Average} \\
        \cmidrule(lr){3-14}
        & & FPR95 \(\downarrow\) & AUROC \(\uparrow\)
        & FPR95 \(\downarrow\) & AUROC \(\uparrow\)
        & FPR95 \(\downarrow\) & AUROC \(\uparrow\)
        & FPR95 \(\downarrow\) & AUROC \(\uparrow\)
        & FPR95 \(\downarrow\) & AUROC \(\uparrow\)
        & FPR95 \(\downarrow\) & AUROC \(\uparrow\) \\
        \midrule
        MSP & \cellcolor{blue!10} NO & 52.87 & 83.02 & 54.76 & 84.30 & 52.16 & 84.92 & 20.33 & 95.25 & 43.08 & 87.23 & 44.64 & 86.94 \\
        Energy & \cellcolor{blue!10} NO & 57.59 & 76.17 & 48.11 & 81.05 & 41.17 & 85.04 & \underline{8.68} & 97.49 & 35.69 & 87.87 & 38.25 & 85.52 \\
        Max logit & \cellcolor{blue!10} NO & 52.30 & 78.42 & 47.22 & 82.29 & 41.65 & 85.60 & 9.50 & 97.43 & 34.57 & 88.39 & 37.05 & 86.43 \\
        GradNorm & \cellcolor{blue!10} NO & 93.95 & 34.15 & 87.71 & 52.32 & 79.27 & 62.51 & 74.90 & 63.44 & 66.67 & 73.87 & 80.50 & 57.26 \\
        ODIN & \cellcolor{blue!10} NO & 60.40 & 82.05 & 71.28 & 76.69 & 68.97 & 77.47 & 47.64 & 88.78 & 55.90 & 84.33 & 60.84 & 81.86 \\
        SCALE & \cellcolor{blue!10} NO & 52.97 & 77.70 & \underline{45.51} & 82.83 & \underline{37.89} & 86.79 & 8.72 & 97.46 & 29.96 & 90.30 & \underline{35.01} & 87.02 \\
        ASH & \cellcolor{blue!10} NO & 57.04 & 76.29 & 47.82 & 81.17 & 40.78 & 85.21 & \textbf{8.43} & 98.01 & 34.80 & 88.15 & 37.77 & 85.77 \\
        RTL & \cellcolor{blue!10} NO & 52.60 & 79.13 & 49.71 & 83.58 & 48.01 & 82.53 & 21.53 & 90.75 & 39.67 & 85.22 & 42.30 & 84.24 \\
        ViM & \cellcolor{red!10} YES & 30.21 & 93.82 & 71.37 & 82.93 & 72.21 & 83.42 & 50.70 & 90.12 & 58.42 & 87.70 & 56.58 & 87.60 \\
        KNN & \cellcolor{red!10} YES & 69.58 & 85.71 & 97.60 & 69.55 & 96.05 & 71.65 & 99.40 & 66.69 & 61.32 & 85.77 & 84.79 & 75.87 \\
        MDS\textsubscript{single} & \cellcolor{red!10} YES & 48.14 & 90.43 & 87.04 & 76.70 & 90.16 & 76.11 & 88.65 & 78.79 & 81.98 & 80.81 & 79.19 & 80.57 \\
        MDS\textsubscript{ensemble} & \cellcolor{red!10} YES & \underline{8.69} & \underline{97.82} & 78.62 & 81.37 & 77.80 & 82.59 & 88.75 & 77.80 & \underline{22.62} & \underline{94.71} & 55.30 & 86.86 \\
        ReAct & \cellcolor{red!10} YES & 97.32 & 53.19 & 90.93 & 65.47 & 85.13 & 72.83 & 74.11 & 85.74 & 84.53 & 76.10 & 86.40 & 70.67 \\
        NNGuide & \cellcolor{red!10} YES & 87.68 & 49.90 & 85.03 & 61.03 & 76.69 & 68.44 & 49.44 & 87.67 & 54.18 & 83.23 & 70.60 & 70.05 \\
        CoRP & \cellcolor{red!10} YES & 42.37 & 93.23 & 68.17 & 83.92 & 62.14 & 86.35 & 33.19 & 94.68 & 39.27 & 92.67 & 49.03 & 90.17 \\
        MDS++ & \cellcolor{red!10} YES & 34.49 & 94.26 & 73.19 & 81.07 & 65.30 & 84.69 & 9.86 & \textbf{98.01} & 26.49 & 94.37 & 41.87 & 90.48 \\
        RMDS & \cellcolor{red!10} YES & 41.02 & 92.19 & 63.46 & \underline{87.29} & 61.54 & \underline{88.84} & 9.41 & \underline{97.54} & 55.96 & 89.27 & 46.28 & \underline{91.03} \\
        RMDS++ & \cellcolor{red!10} YES & 57.33 & 90.42 & 73.07 & 85.76 & 70.16 & 87.53 & 17.70 & 96.44 & 61.79 & 88.50 & 56.01 & 89.73 \\
        OODD & \cellcolor{red!10} YES & 94.09 & 83.80 & 95.52 & 77.85 & 81.25 & 85.78 & 43.03 & 94.68 & 57.68 & 92.52 & 74.31 & 86.93 \\
\rowcolor{blue!5}
        \ours & \cellcolor{blue!10} NO & \textbf{0.59} & \textbf{99.82} & \textbf{14.41} & \textbf{96.31} & \textbf{36.14} & \textbf{90.86} & 20.71 & 95.28 & \textbf{7.90} & \textbf{98.36} & \textbf{15.95} & \textbf{96.13} \\
        \bottomrule
    \end{tabular}
    }
\end{table*}

\begin{table*}[h]
    \vspace*{\fill}
    \centering
    \caption{OOD detection performance comparison with ViT-Tiny. We evaluate with CIFAR-100 ID, CIFAR-100-C csID and the corresponding OODs. (Best: \textbf{bolded}, Second-best: \underline{underlined})}
    \resizebox{\linewidth}{!}{
    \begin{tabular}{l|c|c@{\hspace{2pt}}c|c@{\hspace{2pt}}c|c@{\hspace{2pt}}c|c@{\hspace{2pt}}c|c@{\hspace{2pt}}c||c@{\hspace{2pt}}c}
        \toprule
        \rowcolor{gray!15}
        \multicolumn{14}{c}{\textbf{Covariate Shifted}} \\
        \midrule
        \multirow{2}{*}{Method}
        & \multirow{2}{*}{\makecell[c]{Training set \\ prior}}
        & \multicolumn{2}{c|}{SVHN-C}
        & \multicolumn{2}{c|}{Places365-C}
        & \multicolumn{2}{c|}{LSUN-C}
        & \multicolumn{2}{c|}{iSUN-C}
        & \multicolumn{2}{c||}{Textures-C}
        & \multicolumn{2}{c}{Average} \\
        \cmidrule(lr){3-14}
        & & FPR95 \(\downarrow\) & AUROC \(\uparrow\)
        & FPR95 \(\downarrow\) & AUROC \(\uparrow\)
        & FPR95 \(\downarrow\) & AUROC \(\uparrow\)
        & FPR95 \(\downarrow\) & AUROC \(\uparrow\)
        & FPR95 \(\downarrow\) & AUROC \(\uparrow\)
        & FPR95 \(\downarrow\) & AUROC \(\uparrow\) \\
        \midrule
        MSP & \cellcolor{blue!10} NO & 88.32 & 54.65 & 89.98 & 57.41 & 89.13 & 60.20 & 90.96 & 57.75 & 89.39 & 55.15 & 89.56 & 57.03 \\
        Energy & \cellcolor{blue!10} NO & 83.57 & 60.42 & 88.13 & 60.19 & 87.47 & 63.71 & 89.33 & 60.96 & 86.57 & 60.73 & 87.01 & 61.20 \\
        Max logit & \cellcolor{blue!10} NO & 84.66 & 59.74 & 88.59 & 59.76 & 87.74 & 63.31 & 89.76 & 60.53 & 87.17 & 59.93 & 87.58 & 60.65 \\
        GradNorm & \cellcolor{blue!10} NO & 89.95 & 62.39 & 89.89 & 56.74 & 92.07 & 54.39 & 92.37 & 53.47 & 83.40 & 65.13 & 89.54 & 58.42 \\
        ODIN & \cellcolor{blue!10} NO & 85.29 & 61.87 & 88.93 & 58.27 & 89.56 & 58.57 & 89.54 & 59.06 & 84.33 & 64.82 & 87.53 & 60.52 \\
        SCALE & \cellcolor{blue!10} NO & 84.84 & 59.26 & 89.66 & 58.25 & 90.05 & 60.75 & 91.51 & 57.95 & 88.16 & 58.76 & 88.84 & 58.99 \\
        ASH & \cellcolor{blue!10} NO & 92.41 & 58.90 & 91.56 & 55.26 & 93.51 & 52.40 & 93.88 & 51.33 & 88.62 & 62.12 & 92.00 & 56.00 \\
        RTL & \cellcolor{blue!10} NO & 80.17 & 59.38 & 86.96 & 57.54 & 83.75 & 62.21 & 88.03 & 56.60 & 83.18 & 57.72 & 84.42 & 58.69 \\
        ViM & \cellcolor{red!10} YES & 86.98 & 58.01 & 91.24 & 57.30 & 88.56 & 62.26 & 90.90 & 58.85 & 89.78 & 57.33 & 89.49 & 58.75 \\
        KNN & \cellcolor{red!10} YES & 90.16 & 56.65 & 90.34 & 54.45 & 90.01 & 60.53 & 92.23 & 55.79 & 87.99 & 56.64 & 90.15 & 56.81 \\
        MDS\textsubscript{single} & \cellcolor{red!10} YES & 97.65 & 32.82 & 96.87 & 34.13 & 96.77 & 33.42 & 97.27 & 31.31 & 97.54 & 30.30 & 97.22 & 32.40 \\
        MDS\textsubscript{ensemble} & \cellcolor{red!10} YES & \underline{37.91} & \textbf{90.30} & 94.09 & 45.50 & 61.56 & 61.16 & \underline{61.78} & 61.13 & \textbf{53.29} & \textbf{71.68} & \underline{61.73} & 65.95 \\
        ReAct & \cellcolor{red!10} YES & 82.37 & 60.08 & 87.19 & 61.31 & 85.39 & 65.80 & 87.74 & 62.86 & 86.99 & 59.98 & 85.94 & 62.01 \\
        NNGuide & \cellcolor{red!10} YES & 86.45 & 59.03 & 87.05 & 59.69 & 86.04 & 64.58 & 88.36 & 61.35 & 84.53 & 61.21 & 86.49 & 61.17 \\
        CoRP & \cellcolor{red!10} YES & 88.45 & 59.62 & 89.92 & 56.64 & 87.28 & 63.92 & 89.22 & 59.97 & 86.69 & 59.10 & 88.31 & 59.85 \\
        MDS++ & \cellcolor{red!10} YES & 77.22 & 67.31 & 83.05 & \underline{61.79} & 78.16 & 68.41 & 80.03 & 65.75 & 72.31 & 68.01 & 78.15 & 66.25 \\
        RMDS & \cellcolor{red!10} YES & 76.29 & 65.91 & 87.26 & 60.66 & 84.90 & 65.54 & 87.27 & 61.98 & 77.44 & 63.73 & 82.63 & 63.56 \\
        RMDS++ & \cellcolor{red!10} YES & 75.35 & 65.59 & 86.15 & 61.22 & 83.89 & 66.16 & 86.10 & 62.95 & 75.02 & 64.35 & 81.30 & 64.05 \\
        OODD & \cellcolor{red!10} YES & 58.09 & \underline{73.40} & \underline{77.70} & \textbf{66.19} & \underline{59.98} & \underline{78.36} & 67.62 & \underline{73.08} & 64.52 & \underline{71.14} & 65.58 & \textbf{72.43} \\
\rowcolor{blue!5}
        \ours & \cellcolor{blue!10} NO & \textbf{36.44} & 71.92 & \textbf{67.29} & 57.33 & \textbf{25.08} & \textbf{80.93} & \textbf{31.67} & \textbf{75.50} & \underline{60.90} & 57.30 & \textbf{44.28} & \underline{68.60} \\
        \midrule
        \rowcolor{gray!15}
        \multicolumn{14}{c}{\textbf{Clean}} \\
        \midrule
        \multirow{2}{*}{Method}
        & \multirow{2}{*}{\makecell[c]{Training set \\ prior}}
        & \multicolumn{2}{c|}{SVHN}
        & \multicolumn{2}{c|}{Places365}
        & \multicolumn{2}{c|}{LSUN}
        & \multicolumn{2}{c|}{iSUN}
        & \multicolumn{2}{c||}{Textures}
        & \multicolumn{2}{c}{Average} \\
        \cmidrule(lr){3-14}
        &
        & FPR95 \(\downarrow\) & AUROC \(\uparrow\)
        & FPR95 \(\downarrow\) & AUROC \(\uparrow\)
        & FPR95 \(\downarrow\) & AUROC \(\uparrow\)
        & FPR95 \(\downarrow\) & AUROC \(\uparrow\)
        & FPR95 \(\downarrow\) & AUROC \(\uparrow\)
        & FPR95 \(\downarrow\) & AUROC \(\uparrow\) \\
        \midrule
        MSP & \cellcolor{blue!10} NO & 62.06 & 82.50 & 78.47 & 74.69 & 70.48 & 81.33 & 74.85 & 78.79 & 65.92 & 81.63 & 70.36 & 79.79 \\
        Energy & \cellcolor{blue!10} NO & 41.79 & 89.13 & 73.81 & 77.71 & 61.09 & 86.02 & 66.13 & 83.60 & 51.04 & 87.68 & 58.77 & 84.83 \\
        Max logit & \cellcolor{blue!10} NO & 44.96 & 88.65 & 74.40 & 77.59 & 61.78 & 85.80 & 67.05 & 83.33 & 53.15 & 87.33 & 60.27 & 84.54 \\
        GradNorm & \cellcolor{blue!10} NO & 70.09 & 81.30 & 83.25 & 69.27 & 85.99 & 72.71 & 87.16 & 70.48 & 65.14 & 82.44 & 78.33 & 75.24 \\
        ODIN & \cellcolor{blue!10} NO & 75.65 & 75.47 & 87.48 & 64.34 & 84.41 & 68.70 & 83.64 & 69.79 & 70.07 & 77.78 & 80.25 & 71.22 \\
        SCALE & \cellcolor{blue!10} NO & 44.17 & 88.43 & 73.71 & 77.68 & 67.15 & 83.85 & 71.51 & 81.28 & 52.19 & 87.07 & 61.75 & 83.66 \\
        ASH & \cellcolor{blue!10} NO & 77.25 & 79.20 & 83.45 & 69.11 & 86.72 & 71.52 & 88.41 & 69.02 & 70.01 & 81.14 & 81.17 & 74.00 \\
        RTL & \cellcolor{blue!10} NO & 26.70 & 89.94 & 67.55 & 77.48 & 27.65 & 93.41 & 39.20 & 89.62 & 39.88 & 88.30 & 40.20 & 87.75 \\
        ViM & \cellcolor{red!10} YES & 47.34 & 88.18 & 71.80 & 78.03 & 57.30 & 87.18 & 63.52 & 84.49 & 51.37 & 87.41 & 58.27 & 85.06 \\
        KNN & \cellcolor{red!10} YES & 60.23 & 83.70 & 79.73 & 69.78 & 65.70 & 83.28 & 72.82 & 77.93 & 58.29 & 83.19 & 67.35 & 79.58 \\
        MDS\textsubscript{single} & \cellcolor{red!10} YES & 97.62 & 31.26 & 94.24 & 47.56 & 96.96 & 31.19 & 97.67 & 27.57 & 95.40 & 40.26 & 96.38 & 35.57 \\
        MDS\textsubscript{ensemble} & \cellcolor{red!10} YES & \underline{13.93} & \underline{97.14} & 94.58 & 55.26 & \underline{3.03} & \underline{99.06} & \underline{3.93} & \underline{98.93} & \underline{30.44} & \underline{91.63} & \underline{29.18} & \underline{88.40} \\
        ReAct & \cellcolor{red!10} YES & 40.18 & 89.58 & 72.80 & 78.53 & 57.23 & 68.84 & 62.69 & 84.64 & 51.40 & 87.38 & 56.86 & 81.79 \\
        NNGuide & \cellcolor{red!10} YES & 47.09 & 87.22 & 76.40 & 74.95 & 60.59 & 85.51 & 66.31 & 82.58 & 51.10 & 86.91 & 60.30 & 83.43 \\
        CoRP & \cellcolor{red!10} YES & 61.85 & 85.00 & 80.76 & 73.19 & 64.51 & 85.81 & 68.88 & 82.48 & 60.42 & 85.68 & 67.28 & 82.43 \\
        MDS++ & \cellcolor{red!10} YES & 39.14 & 88.84 & 66.54 & 78.57 & 42.39 & 90.18 & 48.78 & 87.41 & 36.66 & 90.59 & 46.70 & 87.12 \\
        RMDS & \cellcolor{red!10} YES & 36.21 & 89.32 & \textbf{62.15} & \underline{80.43} & 47.61 & 88.92 & 55.13 & 85.97 & 38.81 & 90.00 & 47.98 & 86.93 \\
        RMDS++ & \cellcolor{red!10} YES & 36.30 & 89.09 & \underline{62.64} & 80.08 & 48.16 & 88.67 & 55.21 & 85.72 & 38.26 & 89.98 & 48.11 & 86.71 \\
        OODD & \cellcolor{red!10} YES & 16.35 & 95.80 & 67.08 & 78.32 & 18.51 & 95.64 & 27.48 & 92.79 & \textbf{28.09} & \textbf{92.92} & 31.50 & 91.09 \\
\rowcolor{blue!5}
        \ours & \cellcolor{blue!10} NO & \textbf{3.24} & \textbf{99.23} & 62.77 & \textbf{81.73} & \textbf{0.85} & \textbf{99.80} & \textbf{1.04} & \textbf{99.74} & 35.00 & 91.04 & \textbf{20.58} & \textbf{94.31} \\
        \bottomrule
    \end{tabular}
    }
    \vspace*{\fill}
\end{table*}

\newpage
\begin{table*}[h]
\vspace*{\fill}
    \centering
    \caption{OOD detection performance comparison with Swin-Tiny. We evaluate with CIFAR-100 ID, CIFAR-100-C csID and the corresponding OODs. (Best: \textbf{bolded}, Second-best: \underline{underlined})}
    \resizebox{\linewidth}{!}{
    \begin{tabular}{l|c|c@{\hspace{2pt}}c|c@{\hspace{2pt}}c|c@{\hspace{2pt}}c|c@{\hspace{2pt}}c|c@{\hspace{2pt}}c||c@{\hspace{2pt}}c}
        \toprule
        \rowcolor{gray!15}
        \multicolumn{14}{c}{\textbf{Covariate Shifted}} \\
        \midrule
        \multirow{2}{*}{Method}
        & \multirow{2}{*}{\makecell[c]{Training set \\ prior}}
        & \multicolumn{2}{c|}{SVHN-C}
        & \multicolumn{2}{c|}{Places365-C}
        & \multicolumn{2}{c|}{LSUN-C}
        & \multicolumn{2}{c|}{iSUN-C}
        & \multicolumn{2}{c||}{Textures-C}
        & \multicolumn{2}{c}{Average} \\
        \cmidrule(lr){3-14}
        & & FPR95 \(\downarrow\) & AUROC \(\uparrow\)
        & FPR95 \(\downarrow\) & AUROC \(\uparrow\)
        & FPR95 \(\downarrow\) & AUROC \(\uparrow\)
        & FPR95 \(\downarrow\) & AUROC \(\uparrow\)
        & FPR95 \(\downarrow\) & AUROC \(\uparrow\)
        & FPR95 \(\downarrow\) & AUROC \(\uparrow\) \\
        \midrule
        MSP & \cellcolor{blue!10} NO & 87.48 & 58.92 & 87.20 & 63.29 & 82.95 & 67.03 & 85.14 & 64.75 & 84.33 & 62.97 & 85.42 & 63.39 \\
        Energy & \cellcolor{blue!10} NO & 81.88 & 67.29 & 86.20 & 64.20 & 78.30 & 71.63 & 81.65 & 68.08 & 73.47 & 70.53 & 80.30 & 68.35 \\
        Max logit & \cellcolor{blue!10} NO & 83.72 & 66.10 & 86.30 & 64.38 & 79.40 & 71.16 & 82.44 & 67.79 & 77.63 & 69.22 & 81.90 & 67.73 \\
        GradNorm & \cellcolor{blue!10} NO & 82.38 & 68.38 & 89.90 & 62.00 & 80.76 & 72.92 & 79.68 & \underline{73.52} & \underline{61.12} & \textbf{80.40} & 78.77 & 71.44 \\
        ODIN & \cellcolor{blue!10} NO & 80.78 & 66.20 & 89.74 & 59.86 & 91.86 & 58.65 & 91.08 & 59.38 & 70.96 & 74.06 & 84.88 & 63.63 \\
        SCALE & \cellcolor{blue!10} NO & 79.13 & 69.57 & 86.21 & 65.09 & 76.71 & 73.93 & 78.22 & 72.28 & 63.15 & \underline{76.96} & 76.68 & 71.57 \\
        ASH & \cellcolor{blue!10} NO & 78.65 & 70.33 & 87.57 & 63.95 & 77.48 & 73.87 & 78.64 & 72.06 & 62.42 & 76.76 & 76.95 & 71.39 \\
        RTL & \cellcolor{blue!10} NO & 86.72 & 57.38 & 86.72 & 62.68 & 79.41 & 69.61 & 83.11 & 65.84 & 82.61 & 62.58 & 83.71 & 63.62 \\
        ViM & \cellcolor{red!10} YES & 95.84 & 51.26 & 94.98 & 50.17 & 97.60 & 46.77 & 98.04 & 43.00 & 98.22 & 37.61 & 96.94 & 45.76 \\
        KNN & \cellcolor{red!10} YES & 82.69 & 68.02 & 87.62 & 60.94 & 82.50 & 67.52 & 85.40 & 62.77 & 83.06 & 65.31 & 84.25 & 64.91 \\
        MDS\textsubscript{single} & \cellcolor{red!10} YES & 98.23 & 38.79 & 97.74 & 41.80 & 99.32 & 32.79 & 99.36 & 30.22 & 99.13 & 21.86 & 98.76 & 33.09 \\
        MDS\textsubscript{ensemble} & \cellcolor{red!10} YES & 97.17 & 44.72 & 97.94 & 39.09 & 99.35 & 31.32 & 99.38 & 29.27 & 99.14 & 24.34 & 98.60 & 33.75 \\
        ReAct & \cellcolor{red!10} YES & 78.42 & 69.44 & 85.84 & 65.20 & 76.79 & 73.35 & 78.67 & 71.44 & 64.09 & 76.25 & 76.76 & 71.14 \\
        NNGuide & \cellcolor{red!10} YES & 78.56 & \underline{72.74} & \underline{84.64} & \underline{67.62} & \underline{73.65} & \underline{76.11} & \underline{76.30} & 73.44 & 69.52 & 76.99 & 76.53 & \underline{73.38} \\
        CoRP & \cellcolor{red!10} YES & 84.83 & 67.05 & 89.32 & 59.20 & 85.16 & 65.79 & 86.98 & 61.99 & 84.56 & 64.22 & 86.17 & 63.65 \\
        MDS++ & \cellcolor{red!10} YES & 72.36 & \textbf{75.29} & 88.15 & 60.55 & 78.41 & 69.66 & 79.70 & 66.70 & 61.44 & 76.79 & \underline{76.01} & 69.80 \\
        RMDS & \cellcolor{red!10} YES & 91.82 & 59.47 & 91.80 & 56.76 & 89.66 & 61.30 & 91.74 & 57.40 & 91.61 & 55.75 & 91.33 & 58.14 \\
        RMDS++ & \cellcolor{red!10} YES & 89.90 & 63.59 & 90.80 & 58.82 & 85.84 & 64.50 & 88.39 & 60.72 & 85.74 & 61.41 & 88.13 & 61.81 \\
        OODD & \cellcolor{red!10} YES & \underline{64.59} & 64.49 & 94.47 & 49.89 & 92.05 & 55.47 & 92.54 & 51.21 & 89.78 & 47.89 & 86.69 & 53.79 \\
\rowcolor{blue!5}
        \ours & \cellcolor{blue!10} NO & \textbf{44.90} & 70.99 & \textbf{57.62} & \textbf{75.48} & \textbf{14.28} & \textbf{95.53} & \textbf{24.97} & \textbf{87.74} & \textbf{52.74} & 76.66 & \textbf{38.90} & \textbf{81.28} \\
        \midrule
        \rowcolor{gray!15}
        \multicolumn{14}{c}{\textbf{Clean}} \\
        \midrule
        \multirow{2}{*}{Method}
        & \multirow{2}{*}{\makecell[c]{Training set \\ prior}}
        & \multicolumn{2}{c|}{SVHN}
        & \multicolumn{2}{c|}{Places365}
        & \multicolumn{2}{c|}{LSUN}
        & \multicolumn{2}{c|}{iSUN}
        & \multicolumn{2}{c||}{Textures}
        & \multicolumn{2}{c}{Average} \\
        \cmidrule(lr){3-14}
        &
        & FPR95 \(\downarrow\) & AUROC \(\uparrow\)
        & FPR95 \(\downarrow\) & AUROC \(\uparrow\)
        & FPR95 \(\downarrow\) & AUROC \(\uparrow\)
        & FPR95 \(\downarrow\) & AUROC \(\uparrow\)
        & FPR95 \(\downarrow\) & AUROC \(\uparrow\)
        & FPR95 \(\downarrow\) & AUROC \(\uparrow\) \\
        \midrule
        MSP & \cellcolor{blue!10} NO & 59.06 & 86.09 & 69.83 & 79.17 & 58.09 & 85.76 & 62.11 & 83.42 & 51.72 & 87.13 & 60.16 & 84.31 \\
        Energy & \cellcolor{blue!10} NO & 35.88 & 91.83 & 58.64 & 83.16 & 38.61 & 91.40 & 44.35 & 88.95 & 27.80 & 93.67 & 41.06 & 89.80 \\
        Max logit & \cellcolor{blue!10} NO & 37.54 & 91.54 & 58.73 & 83.04 & 39.41 & 91.10 & 45.02 & 88.66 & 29.86 & 93.28 & 42.11 & 89.52 \\
        GradNorm & \cellcolor{blue!10} NO & 85.31 & 61.74 & 83.37 & 67.78 & 70.06 & 80.46 & 68.41 & 80.33 & 51.02 & 82.32 & 71.63 & 74.53 \\
        ODIN & \cellcolor{blue!10} NO & 89.89 & 67.25 & 95.52 & 59.39 & 97.62 & 57.25 & 97.04 & 57.87 & 75.93 & 76.50 & 91.20 & 63.65 \\
        SCALE & \cellcolor{blue!10} NO & 49.51 & 87.94 & 62.13 & 81.02 & 40.86 & 91.00 & 44.03 & 89.32 & 27.67 & 93.21 & 44.84 & 88.50 \\
        ASH & \cellcolor{blue!10} NO & 54.51 & 85.90 & 67.96 & 79.09 & 46.42 & 89.69 & 48.93 & 88.02 & 32.31 & 92.14 & 50.03 & 86.97 \\
        RTL & \cellcolor{blue!10} NO & 43.25 & 87.35 & 70.99 & 74.76 & 44.41 & 87.35 & 51.83 & 84.40 & 44.03 & 87.30 & 50.90 & 84.23 \\
        ViM & \cellcolor{red!10} YES & 83.71 & 80.08 & 93.84 & 62.22 & 96.72 & 63.50 & 97.19 & 59.52 & 93.54 & 65.96 & 93.00 & 66.26 \\
        KNN & \cellcolor{red!10} YES & 31.60 & 93.26 & 66.75 & 80.31 & 46.95 & 89.67 & 53.14 & 86.22 & 31.23 & 93.09 & 45.93 & 88.51 \\
        MDS\textsubscript{single} & \cellcolor{red!10} YES & 97.07 & 59.75 & 98.46 & 43.08 & 99.67 & 35.81 & 99.72 & 33.62 & 98.43 & 33.27 & 98.67 & 41.11 \\
        MDS\textsubscript{ensemble} & \cellcolor{red!10} YES & 96.51 & 64.83 & 98.40 & 43.33 & 99.57 & 38.31 & 99.65 & 36.58 & 97.99 & 44.61 & 98.42 & 45.53 \\
        ReAct & \cellcolor{red!10} YES & 42.84 & 90.56 & 58.78 & 82.17 & 38.80 & 91.42 & 42.60 & 89.57 & 25.89 & 93.83 & 41.78 & 89.51 \\
        NNGuide & \cellcolor{red!10} YES & 37.51 & 91.82 & \textbf{56.58} & \textbf{84.28} & 33.61 & \underline{93.04} & \underline{38.34} & \underline{91.40} & \underline{24.15} & \underline{94.75} & 38.04 & \underline{91.06} \\
        CoRP & \cellcolor{red!10} YES & 41.14 & 92.32 & 68.66 & 80.55 & 50.94 & 89.81 & 55.48 & 87.22 & 35.63 & 92.77 & 50.37 & 88.53 \\
        MDS++ & \cellcolor{red!10} YES & \underline{30.97} & \underline{93.67} & \underline{56.68} & \underline{83.95} & \underline{33.31} & 92.98 & 39.00 & 90.82 & \textbf{16.43} & \textbf{96.28} & \underline{35.28} & \textbf{91.54} \\
        RMDS & \cellcolor{red!10} YES & 52.84 & 89.57 & 64.26 & 82.37 & 52.70 & 88.97 & 59.42 & 86.09 & 40.78 & 91.23 & 54.00 & 87.65 \\
        RMDS++ & \cellcolor{red!10} YES & 47.14 & 90.76 & 58.64 & 83.56 & 44.55 & 90.16 & 50.92 & 87.55 & 33.78 & 92.48 & 47.01 & 88.90 \\
        OODD & \cellcolor{red!10} YES & 51.68 & 91.28 & 85.69 & 74.44 & 84.03 & 79.98 & 87.90 & 74.33 & 82.31 & 75.88 & 78.32 & 79.18 \\
\rowcolor{blue!5}
        \ours & \cellcolor{blue!10} NO & \textbf{7.27} & \textbf{97.88} & 77.97 & 64.13 & \textbf{6.31} & \textbf{98.52} & \textbf{6.68} & \textbf{98.49} & \underline{61.06} & 81.14 & \textbf{31.86} & 88.03 \\
        \bottomrule
    \end{tabular}
    }
    \vspace*{\fill}
\end{table*}

\newpage
\begin{table*}[h]
    \centering
    \caption{OOD detection performance comparison with ResNet-50 on ImageNet-based benchmark. 
    Results on covariated-shifted dataset are the average of all 15 corruptions with severity level 5. (Best: \textbf{bolded}, Second-best: \underline{underlined})}
    \label{tab:results_resnet50}
    \resizebox{\linewidth}{!}{
    \begin{tabular}{l|c|c@{\hspace{2pt}}c|c@{\hspace{2pt}}c|c@{\hspace{2pt}}c|c@{\hspace{2pt}}c|c@{\hspace{2pt}}c||c@{\hspace{2pt}}c}
        \toprule
        \rowcolor{gray!15}
        \multicolumn{14}{c}{\textbf{Covariate Shifted}} \\
        \midrule
        \multirow{2}{*}{Method} 
        & \multirow{2}{*}{\makecell[c]{Training set \\ prior}} 
        & \multicolumn{2}{c|}{ImageNet-O-C} 
        & \multicolumn{2}{c|}{Places-C} 
        & \multicolumn{2}{c|}{SUN-C} 
        & \multicolumn{2}{c|}{iNaturalist-C} 
        & \multicolumn{2}{c||}{Textures-C} 
        & \multicolumn{2}{c}{Average} \\
        \cmidrule(lr){3-14} 
        & & FPR95 \(\downarrow\) & AUROC \(\uparrow\) 
        & FPR95 \(\downarrow\) & AUROC \(\uparrow\) 
        & FPR95 \(\downarrow\) & AUROC \(\uparrow\) 
        & FPR95 \(\downarrow\) & AUROC \(\uparrow\) 
        & FPR95 \(\downarrow\) & AUROC \(\uparrow\) 
        & FPR95 \(\downarrow\) & AUROC \(\uparrow\) \\
        \midrule
        MSP                         & \cellcolor{blue!10} NO 
        & 86.14 & 55.60 & 86.79 & 63.17 & 84.84 & 65.27 & 77.89 & 69.62 & 88.52 & 54.85 & 84.84 & 61.70 \\
        Energy                      & \cellcolor{blue!10} NO 
        & 82.16 & 58.92 & 89.03 & 61.29 & 87.91 & 63.58 & 86.44 & 64.52 & 88.05 & 55.85 & 86.72 & 60.83 \\
        Max logit                   & \cellcolor{blue!10} NO  
        & 83.88 & 57.81 & 87.57 & 63.00 & 85.79 & 65.42 & 81.49 & 68.07 & 88.02 & 55.87 & 85.35 & 62.03 \\
        GradNorm                    & \cellcolor{blue!10} NO 
        & 76.86 & 64.58 & 70.84 & 77.06 & 65.23 & 80.51 & 51.36 & 85.62 & 67.11 & 74.14 & 66.28 & 76.38 \\
        ODIN                        & \cellcolor{blue!10} NO 
        & 76.69 & 68.04 & \underline{25.14} & \underline{92.96} & \underline{22.70} & \underline{93.69} & 27.84 & 92.22 & 38.11 & 85.81 & \underline{38.10} & \underline{86.54} \\
        SCALE                       & \cellcolor{blue!10} NO 
        & 79.10 & 63.07 & 79.87 & 70.76 & 76.10 & 74.05 & 63.83 & 79.72 & 75.88 & 68.39 & 74.96 & 71.20 \\
        ASH                         & \cellcolor{blue!10} NO 
        & 80.51 & 61.30 & 87.83 & 63.20 & 86.48 & 65.88 & 82.64 & 68.22 & 84.77 & 60.05 & 84.45 & 63.73 \\
        RTL                         & \cellcolor{blue!10} NO 
        & 82.27 & 57.85 & 76.94 & 71.71 & 73.29 & 74.47 & 64.06 & 78.47 & 81.14 & 59.14 & 75.54 & 68.33 \\
        ViM                         & \cellcolor{red!10} YES 
        & 95.64 & 38.92 & 99.50 & 17.64 & 99.65 & 14.90 & 99.94 & 8.60  & 96.94 & 25.14 & 98.33 & 21.04 \\
        KNN                         & \cellcolor{red!10} YES 
        & 83.81 & 63.63 & 90.93 & 55.23 & 91.12 & 56.45 & 95.20 & 45.36 & 61.31 & 73.33 & 84.47 & 58.80 \\
        MDS\textsubscript{single}   & \cellcolor{red!10} YES 
        & 95.14 & 38.80 & 99.52 & 18.75 & 99.67 & 15.95 & 99.93 & 9.60  & 96.20 & 27.69 & 98.09 & 22.16 \\
        MDS\textsubscript{ensemble} & \cellcolor{red!10} YES 
        & 79.71 & 62.76 & 92.44 & 42.67 & 92.17 & 41.57 & 94.26 & 33.49 & 64.44 & 62.83 & 84.60 & 48.66 \\
        ReAct                       & \cellcolor{red!10} YES 
        & 82.24 & 59.09 & 84.47 & 67.56 & 82.92 & 69.76 & 80.76 & 70.72 & 84.85 & 59.33 & 83.05 & 65.29 \\
        NNGuide                     & \cellcolor{red!10} YES 
        & 73.91 & 67.80 & 68.11 & 77.74 & 62.29 & 81.42 & 50.54 & 85.43 & 57.13 & 78.16 & 62.40 & 78.11 \\
        CoRP                        & \cellcolor{red!10} YES
        & 75.89 & \underline{73.38} & 75.07 & 73.16 & 71.42 & 76.37 & 67.32 & 78.15 & 44.54 & 86.27 & 66.85 & 77.47 \\
        MDS++                       & \cellcolor{red!10} YES
        & \textbf{63.76} & \textbf{77.47} & 81.50 & 66.32 & 79.47 & 69.03 & 66.54 & 77.67 & \underline{36.03} & \textbf{88.92} & 65.46 & 75.88 \\
        RMDS                        & \cellcolor{red!10} YES
        & 90.38 & 50.24 & 95.39 & 44.28 & 95.88 & 42.31 & 96.52 & 45.83 & 85.80 & 50.46 & 92.79 & 46.62 \\
        RMDS++                      & \cellcolor{red!10} YES
        & 77.40 & 64.63 & 86.31 & 58.47 & 85.77 & 58.86 & 78.16 & 68.26 & 61.45 & 73.34 & 77.82 & 64.71 \\
        OODD                        & \cellcolor{red!10} YES
        & 69.63 & 71.60 & 53.83 & 82.98 & 45.64 & 86.60 & \underline{25.13} & \underline{93.69} & 38.52 & \underline{86.52} & 46.55 & 84.28 \\
        \midrule
        \rowcolor{blue!5}
        \ours                       & \cellcolor{blue!10} NO  
        & \underline{73.52} & 58.78 & \textbf{2.46}  & \textbf{99.34} & \textbf{1.06}  & \textbf{99.59} & \textbf{1.18}  & \textbf{99.61} & \textbf{23.75} & 79.75 & \textbf{20.39} & \textbf{87.41} \\
        \midrule
        \rowcolor{gray!15}
        \multicolumn{14}{c}{\textbf{Clean}} \\
        \midrule
        \multirow{2}{*}{Method} 
        & \multirow{2}{*}{\makecell[c]{Training set \\ prior}} 
        & \multicolumn{2}{c|}{ImageNet-O} 
        & \multicolumn{2}{c|}{Places} 
        & \multicolumn{2}{c|}{SUN} 
        & \multicolumn{2}{c|}{iNaturalist} 
        & \multicolumn{2}{c||}{Textures} 
        & \multicolumn{2}{c}{Average} \\
        \cmidrule(lr){3-14} 
        &
        & FPR95 \(\downarrow\) & AUROC \(\uparrow\) 
        & FPR95 \(\downarrow\) & AUROC \(\uparrow\) 
        & FPR95 \(\downarrow\) & AUROC \(\uparrow\) 
        & FPR95 \(\downarrow\) & AUROC \(\uparrow\) 
        & FPR95 \(\downarrow\) & AUROC \(\uparrow\) 
        & FPR95 \(\downarrow\) & AUROC \(\uparrow\) \\
        \midrule
        MSP                         & \cellcolor{blue!10} NO 
        & 64.81 & 75.42 & 56.33 & 85.09 & 53.18 & 85.97 & 36.95 & 91.68 & 53.73 & 83.95 & 53.00 & 84.42 \\
        Energy                      & \cellcolor{blue!10} NO 
        & 55.18 & 82.01 & 42.09 & 89.64 & 34.38 & 91.54 & 25.30 & 94.38 & 33.96 & 90.49 & 38.18 & 89.61 \\
        Max logit                   & \cellcolor{blue!10} NO  
        & 55.09 & 81.85 & 42.79 & 78.62 & 35.50 & 91.39 & 23.88 & 95.05 & 35.07 & 90.28 & 38.47 & 87.44 \\
        GradNorm                    & \cellcolor{blue!10} NO 
        & 55.97 & 78.38 & 47.85 & 87.45 & 42.19 & 88.60 & 24.01 & 94.34 & 42.17 & 86.85 & 42.44 & 87.12 \\
        ODIN                        & \cellcolor{blue!10} NO 
        & 18.70 & 94.49 & 94.09 & 64.38 & 93.09 & 64.67 & 96.72 & 51.10 & 29.60 & 91.41 & 66.44 & 73.21 \\
        SCALE                       & \cellcolor{blue!10} NO 
        & 50.55 & 84.83 & 37.29 & 91.44 & 30.29 & 93.06 & 17.02 & 96.40 & 31.67 & 92.22 & 33.36 & 91.59 \\
        ASH                         & \cellcolor{blue!10} NO 
        & 42.77 & 87.82 & \underline{32.37} & \underline{92.47} & 24.55 & 94.22 & \underline{10.87} & \underline{97.60} & 19.80 & 94.77 & 26.07 & 93.38 \\
        RTL                         & \cellcolor{blue!10} NO 
        & 56.98 & 75.34 & 42.41 & 86.56 & 36.48 & 87.81 & 23.85 & 91.29 & 38.71 & 85.14 & 39.69 & 85.23 \\
        ViM                         & \cellcolor{red!10} YES 
        & 56.20 & 77.62 & 42.62 & 86.49 & 31.60 & 90.55 & 21.81 & 94.30 & 33.31 & 89.08 & 37.11 & 87.61 \\
        KNN                         & \cellcolor{red!10} YES 
        & 11.09 & 97.01 & 72.62 & 83.16 & 72.49 & 84.40 & 82.54 & 79.21 & 16.31 & 96.53 & 51.01 & 88.06 \\
        MDS\textsubscript{single}    & \cellcolor{red!10} YES 
        & 11.10 & 95.67 & 54.71 & 87.07 & 45.46 & 90.53 & 78.67 & 76.90 & 19.89 & 94.97 & 41.97 & 89.03 \\
        MDS\textsubscript{ensemble}  & \cellcolor{red!10} YES 
        & 30.76 & 88.27 & 95.87 & 62.81 & 95.47 & 62.19 & 97.37 & 50.94 & 49.44 & 86.03 & 73.78 & 70.05 \\
        ReAct                       & \cellcolor{red!10} YES 
        & 16.84 & 95.44 & 36.62 & 89.54 & 34.89 & 89.80 & 38.99 & 88.92 & 35.50 & 88.30 & 32.57 & 90.40 \\
        NNGuide                     & \cellcolor{red!10} YES 
        & 38.59 & 88.32 & \textbf{28.59} & \textbf{93.10} & 20.02 & \underline{95.18} & 17.82 & 96.20 & 21.11 & 94.17 & 25.23 & 93.39 \\
        CoRP                        & \cellcolor{red!10} YES  
        & 30.01 & 89.25 & 64.00 & 84.28 & 55.98 & 87.82 & 83.16 & 73.97 & 23.66 & 94.53 & 51.36 & 85.97 \\
        MDS++                       & \cellcolor{red!10} YES  
        & 14.92 & 97.16 & 61.42 & 84.92 & 49.94 & 88.97 & 33.84 & 93.72 & \textbf{1.69}  & \textbf{99.49} & 32.36 & 92.85 \\
        RMDS                        & \cellcolor{red!10} YES  
        & 61.88 & 88.56 & 87.73 & 79.15 & 88.16 & 80.24 & 65.98 & 90.30 & 51.95 & 88.58 & 71.14 & 84.97 \\
        RMDS++                      & \cellcolor{red!10} YES  
        & 56.42 & 86.76 & 75.65 & 82.07 & 73.61 & 83.96 & 36.83 & 93.50 & 29.65 & 91.97 & 54.43 & 87.65 \\
        OODD                        & \cellcolor{red!10} YES
        & \underline{9.77}  & \underline{97.79} & 37.01 & 90.23 & \underline{18.97} & 94.94 & \textbf{4.50}  & \textbf{98.74} & \underline{5.83}  & \underline{98.53} & \textbf{15.22} & \textbf{96.05} \\
        \midrule
        \rowcolor{blue!5}
        \ours                       & \cellcolor{blue!10} NO  
        & \textbf{0.60}  & \textbf{99.83} & 33.90 & 89.31 & \textbf{15.90} & \textbf{95.77} & 14.29 & 96.46 & 23.68 & 94.78 & \underline{17.67} & \underline{95.23} \\
        \bottomrule
    \end{tabular}
    }
\end{table*}


\end{document}